\RequirePackage[svgnames,table,dvipsnames]{xcolor}
\documentclass[11pt,letterpaper,logo]{merged_style}

\usepackage[comma,authoryear,compress]{natbib}
\usepackage{subcaption}
\usepackage{etoolbox}
\usepackage{xspace}
\usepackage[capitalize]{cleveref}
\usepackage{multirow}
\usepackage{array}
\usepackage{algorithm}
\usepackage{algpseudocode}
\usepackage{placeins}
\usepackage{etoc}
\AtBeginDocument{\etocdepthtag.toc{main}}
\newcommand{\clip}{\operatorname{clip}}
\newcommand{\psnr}{\operatorname{PSNR}}
\newcommand{\ssim}{\operatorname{SSIM}}
\newcommand{\lpips}{\operatorname{LPIPS}}

\newcommand{\methodtype}[1]{{\rmfamily\upshape\bfseries #1}}

\newcommand{\tworowhead}[1]{%
  \multirow[c]{2}{*}[-\dimexpr(\aboverulesep+\belowrulesep+\cmidrulewidth)/2\relax]{#1}}

\makeatletter
\DeclareRobustCommand\onedot{\futurelet\@let@token\@onedot}
\def\@onedot{\ifx\@let@token.\else.\null\fi\xspace}
\newcommand{\etal}{\emph{et al}\onedot}
\renewcommand{\@pnumwidth}{1.9em}
\makeatother

\newcommand{\darkzerotitle}{Dark-Zero: Learning to Illuminate via Evolving Degradations\\for Zero-Shot Low-Light Enhancement}
\title{\darkzerotitle}
\runningtitle{Dark-Zero: Learning to Illuminate via Evolving Degradations for Zero-Shot Low-Light Enhancement}
\newcommand{\authoremail}[1]{\href{mailto:#1}{\textcolor{black}{\texttt{#1}}}}

\author{\begin{minipage}{\textwidth}
  \centering
  Hanzhen Wang\textsuperscript{\normalfont *,1}, \quad
  Yuanyang Qi\textsuperscript{\normalfont *,2}, \quad
  Jie Hong\textsuperscript{\normalfont 2}, \quad
  Shuang Qiu\textsuperscript{\normalfont \textdagger,1}\par
  \vspace{5pt}
  {\normalfont\fontsize{11}{13}\selectfont
    \textsuperscript{1}City University of Hong Kong\qquad
    \textsuperscript{2}The University of Hong Kong\par
    \vspace{5pt}
    \authoremail{hanzhwang9-c@my.cityu.edu.hk}\quad
    \authoremail{yuanyang.qi@connect.hku.hk}\par
    \authoremail{jiehong@hku.hk}\quad
    \authoremail{shuanqiu@cityu.edu.hk}\par}
\end{minipage}}

\bannerlogos{%
  \hfill
  \includegraphics[height=10.5mm]{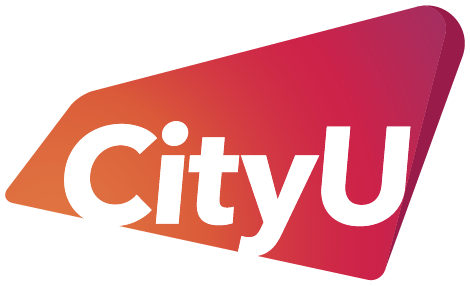}\hspace{12pt}%
  \includegraphics[height=11mm]{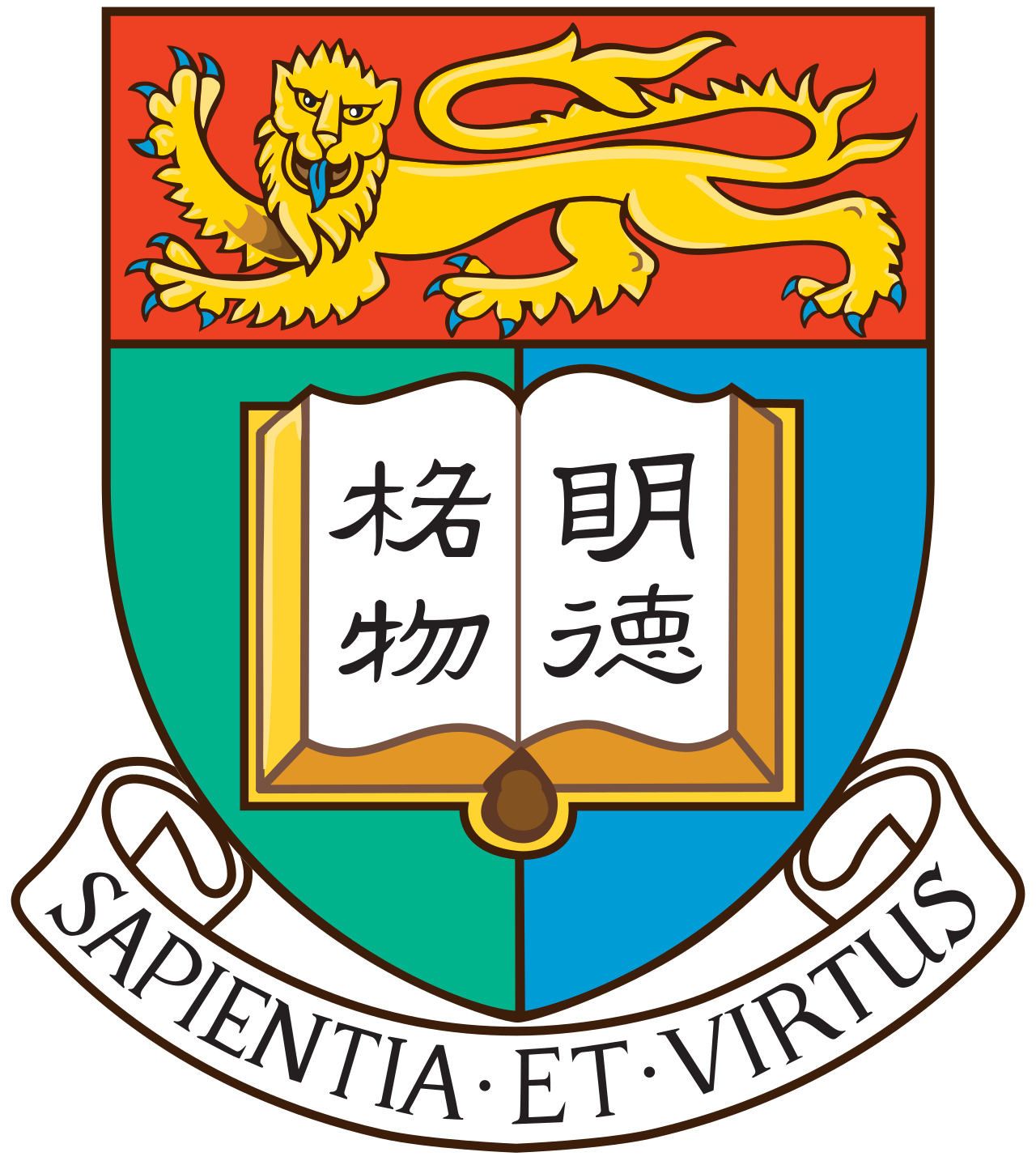}%
  \vspace{-1.1cm}
}
\hypersetup{pdfauthor={Hanzhen Wang, Yuanyang Qi, Jie Hong, Shuang Qiu}}

\crefname{section}{Sec.}{Secs.}
\Crefname{section}{Section}{Sections}
\crefname{table}{Tab.}{Tabs.}
\Crefname{table}{Table}{Tables}
\crefname{appendix}{Appendix}{Appendices}
\Crefname{appendix}{Appendix}{Appendices}

\begin{abstract}
Low-light image enhancement aims to improve visibility and perceptual quality in dark scenes. To avoid collecting low-light training data, normal-light-only training relies on transferable synthetic supervision for real low-light enhancement. However, existing approaches synthesize image pairs from fixed degradation distributions throughout training, failing to adapt as the restoration model improves. We therefore propose Dark-Zero, a Challenger-Solver framework that dynamically adapts synthetic supervision according to the Solver's measured learning progress. Specifically, the Challenger network generates synthetic low-light inputs through a learned physical degradation policy, while the Solver network learns to reconstruct the normal-light references. The Solver's relative PSNR improvement on each generated pair after its updates serves as a learning-progress reward for training the Challenger via proximal policy optimization. Both networks are trained from random initialization using only normal-light images from COCO, without any real low-light data. Evaluated across five Solver architectures, Dark-Zero with DarkIR achieves a mean PSNR of 22.84\,dB on LOL, a state-of-the-art score among zero-shot LLIE methods. On the Unpaired benchmark, it also outperforms MR.~Illuminate, the existing state-of-the-art zero-shot method, across most no-reference metrics. At inference, Dark-Zero requires only a single forward pass through the fixed Solver, using approximately $1/1{,}100$ of the inference FLOPs reported for MR.~Illuminate.
\end{abstract}

\begin{document}
\maketitle
\begingroup
\renewcommand{\thefootnote}{}\NoHyper
\footnotetext{\begin{tabular}[t]{@{}l@{}}
\textsuperscript{*}Equal contribution.\\
\textsuperscript{\textdagger}Corresponding author.
\end{tabular}}\endNoHyper
\endgroup

\suppressfloats[t]
\begin{figure}[t]
  \centering
  \includegraphics[width=\linewidth]{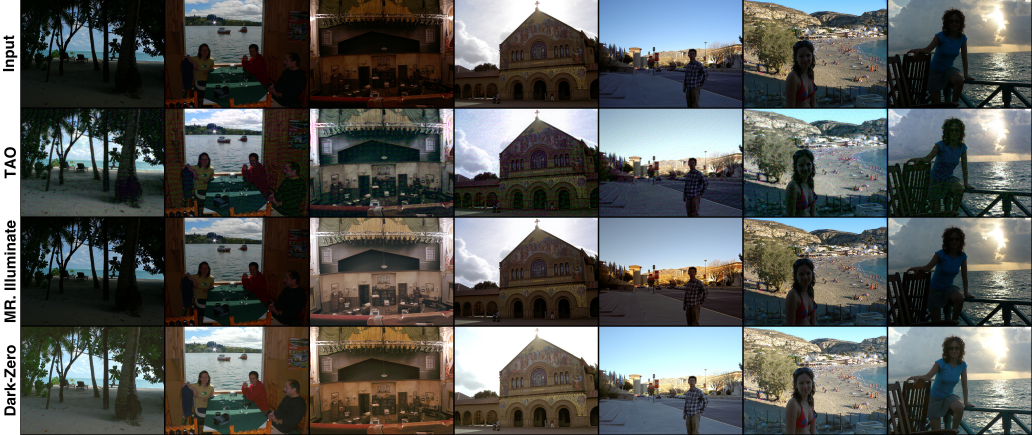}
  \caption{Qualitative comparison on unpaired VV and DICM images spanning dim interiors and backlit outdoor scenes. Below the low-light inputs, the three rows show results from TAO, MR.~Illuminate, and Dark-Zero, respectively.}
  \label{fig:teaser}
\end{figure}

\vspace{+6pt}
\section{Introduction}
\label{sec:introduction}
\vspace{+6pt}
Low-light image enhancement (LLIE) improves visibility in poorly illuminated scenes by recovering structural details, natural color, and balanced contrast, thereby supporting both human perception and downstream vision tasks. However, sensor noise, color distortions, and nonlinear camera processing complicate restoration from weak image signals~\citep{brooks2019unprocessing,wei2020physics}. Furthermore, acquiring aligned normal/low-light pairs across diverse illumination conditions is prohibitively difficult, limiting diversity of paired supervision.

\begin{figure}[t]
\centering
\includegraphics[width=\linewidth]{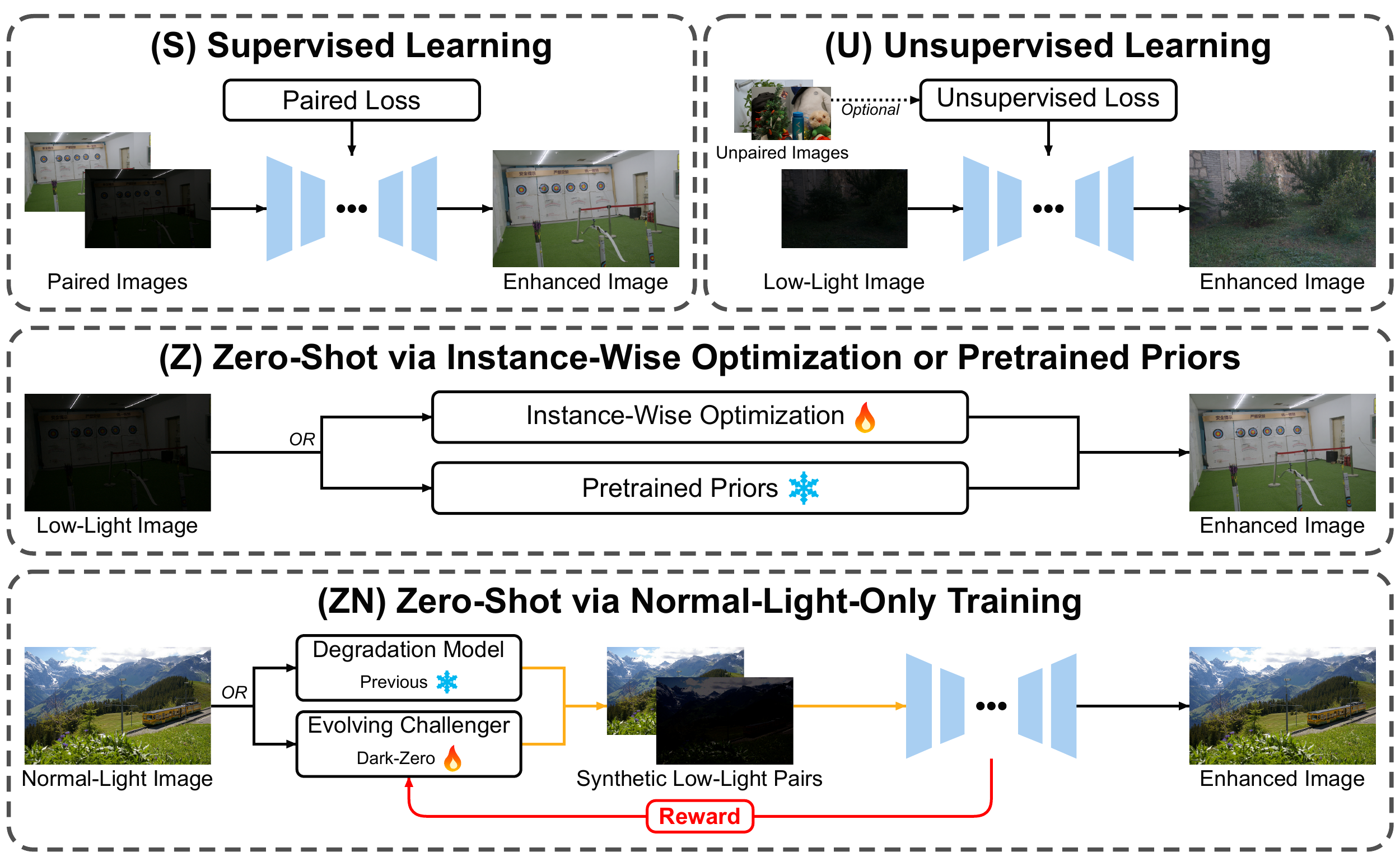}
\caption{Comparison of LLIE pipelines by sources of restoration guidance and learning stages. The diagram traces how supervision and prior knowledge guide the enhancement process, clarifying each pipeline's data dependencies and adaptation requirements during training and inference.}
\label{fig:training_categories}
\end{figure}

Existing LLIE methods vary in their reliance on paired supervision and in how they restore test images. Building upon the classification in MR.~Illuminate~\citep{cho2026mr}, we group LLIE methods into four categories according to data and inference requirements, as illustrated in \cref{fig:training_categories}. Supervised methods (\methodtype{S}), including Retinexformer~\citep{cai2023retinexformer} and DarkIR~\citep{Feijoo_2025_CVPR}, learn from aligned real low-light and normal-light pairs on benchmarks such as LOL~\citep{wei2018deep}. Unsupervised methods (\methodtype{U}) use unpaired or zero-reference supervision yet typically still require real low-light training images. Both \methodtype{Z} and \methodtype{ZN} address zero-shot LLIE, enhancing real low-light images without prior task-specific training on real low-light data. Specifically, \methodtype{Z} methods rely on pretrained image priors or instance-wise optimization. In contrast, we focus on the more challenging \methodtype{ZN} setting of normal-light-only training, where synthetic degradation provides supervision and the trained model generalizes to unseen real low-light images in a single forward pass without further parameter updates.

In the \methodtype{ZN} setting, the degradation process determines which low-light conditions the model encounters during training and thus shapes its generalization to real images. However, a fixed degradation process keeps the training distribution unchanged even as the restoration model improves. Overly severe degradation can obscure recoverable structure, while easy examples that the model already restores well offer little further learning signal. Learning-progress curricula~\citep{4141061,graves2017automated,portelas2020teacher} suggest using changes in learner performance to dynamically guide training input generation. This motivates our central question: 
How can we adapt synthetic supervision to the restoration model's learning progress using only normal-light images?

To address this question, we propose \textbf{Dark-Zero}, short for \textbf{D}egradation \textbf{A}daptation through \textbf{R}estoration \textbf{K}nowledge \textbf{Zero}-Shot Learning, a Challenger-Solver framework. Under normal-light-only training, the Challenger generates synthetic low-light inputs from reference images, and the Solver learns to reconstruct those references. For physical low-light synthesis, we adapt the sequential image signal processing (ISP) controller of AdaptiveISP~\citep{NEURIPS2024_cc596a80} as the Challenger. Notably, both networks are trained from random initialization using normal-light-only Microsoft Common Objects in Context (COCO)~\citep{10.1007/978-3-319-10602-1_48} images. To assess the immediate training value of each generated pair, we introduce a learning progress reward based on the Solver's relative improvement in peak signal-to-noise ratio (PSNR). PSNR is measured before and after a short sequence of supervised updates on that same pair, with the resulting reward guiding Challenger updates via proximal policy optimization (PPO)~\citep{schulman2017proximal}. 

To assess the generalization of this adaptive approach, we evaluate five Solver architectures on unseen real low-light images. Using DarkIR as the main Solver, Dark-Zero surpasses MR.~Illuminate's published LOL PSNR by an average of $1.10$,dB across three random seeds. On the Unpaired benchmark, Dark-Zero's mean integrated local natural image quality evaluator (ILNIQE)~\citep{zhang2015feature} and blind/referenceless image spatial quality evaluator (BRISQUE)~\citep{mittal2012no} scores are lower than MR.~Illuminate's published values by $1.86$ and $0.42$, respectively. Detailed evaluation and aggregation protocols are specified in \cref{sec:datasets,sec:implementation}. \Cref{fig:teaser} complements these quantitative metrics with a qualitative comparison against TAO~\citep{gou2023test} and MR.~Illuminate on unpaired VV~\citep{vonikakis2018evaluation} and DICM~\citep{lee2013contrast} images. Dark-Zero brightens shadowed regions in these examples while retaining visible scene structure. With the Challenger confined to training, the trained Solver produces these restorations in a single forward pass after input-only mean-intensity rescaling. For the main configuration, the estimated inference cost at $400\times600$ resolution is about $1/1{,}100$ of MR.~Illuminate's reported floating-point operations (FLOPs). Our main contributions are fourfold:
\begin{itemize}
\item We introduce Dark-Zero, a Challenger-Solver framework that couples low-light synthesis with restoration learning to adapt synthetic supervision for zero-shot LLIE under the normal-light-only training constraint, \textit{i.e.}, ZN setting.
\item We develop an evolving Challenger that optimizes its physical low-light degradation policy via PPO, leveraging the Solver's relative restoration gains as a learning-progress reward.
\item Our experiments on paired and unpaired benchmarks demonstrate generalization from normal-light-only training to unseen real low-light images. In these evaluations, Dark-Zero achieves state-of-the-art zero-shot results.
\item Dark-Zero supports various Solver architectures through a common training interface. Its main configuration restores images in one Solver forward pass at substantially lower inference cost than MR.~Illuminate.
\end{itemize}

\begin{figure*}[!t]
\centering
\includegraphics[width=\linewidth]{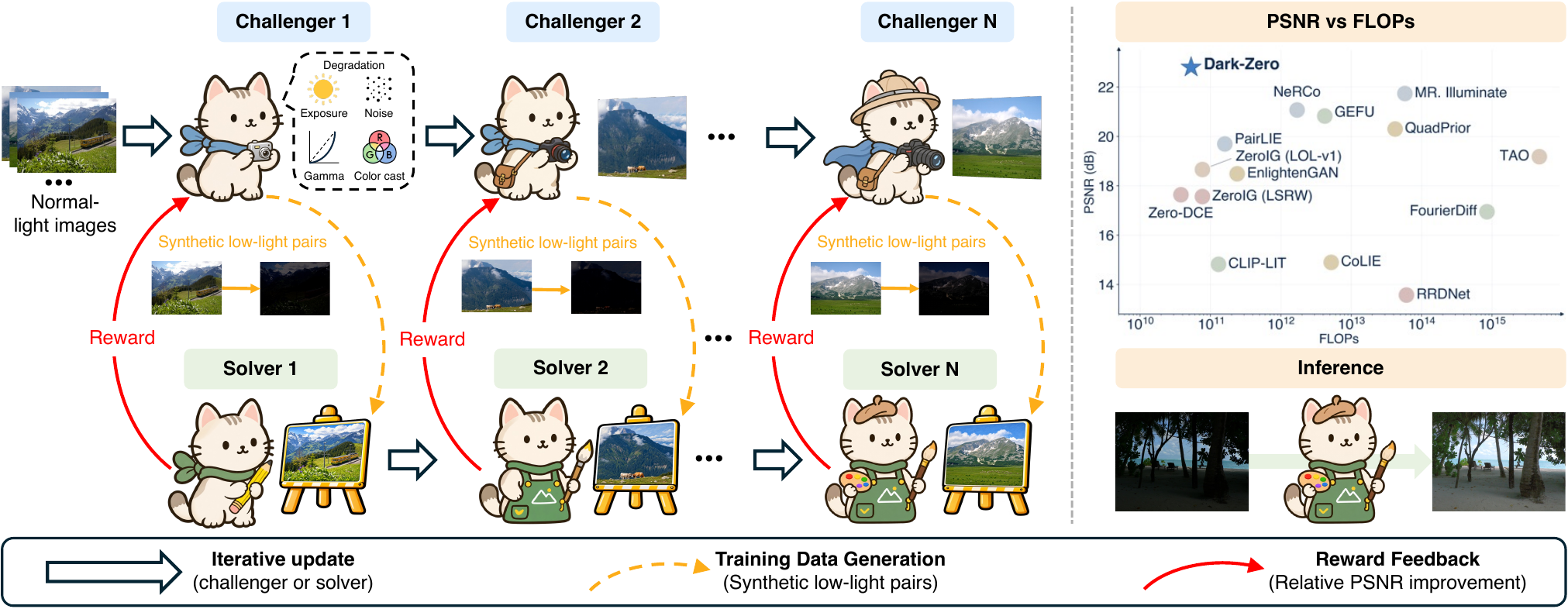}
\caption{Dark-Zero training and Solver-only inference. The numbered Challenger-Solver pairs depict successive training states, with synthetic low-light pairs and relative PSNR feedback linking the two components. The upper-right panel compares LOL PSNR and inference FLOPs across a subset of representative methods from \cref{tab:main}, and the lower-right panel shows inference with the trained Solver.}
\label{fig:darkzero_framework}
\end{figure*}

\section{Related Work}
\label{sec:related}

\paragraph{Low-light enhancement and restoration.} Paired LLIE research has explored noise handling~\citep{xu2022snr}, color representations~\citep{11093284}, and brightness supervision~\citep{liao2025gt}, alongside  diffusion~\citep{hou2023global,liu2024diff} and efficient Retinex models~\citep{brateanu2026multinex}. To relax the requirement for aligned normal-light targets, unsupervised approaches leverage adversarial learning~\citep{jiang2021enlightengan}, curve estimation~\citep{guo2020zero,9369102}, or paired low-light observations~\citep{fu2023learning}, as well as vision-language guidance~\citep{liang2023iterative,yang2023implicit} and continuous exposure modeling~\citep{ICLR2025_5c89b403}. Latent-Retinex diffusion~\citep{jiang2024lightendiffusion} and joint denoising~\citep{ICLR2025_6264f71a} also support unsupervised enhancement. For zero-shot LLIE, methods adapt to individual inputs~\citep{zhang2019zero,zhu2020zero,chobola2024fast} or use pretrained diffusion priors~\citep{fei2023generative,lv2024fourier,huang2025zero,Lin_Ye_Chen_Fu_Wang_Chai_Xing_Li_Zhu_Ding_2025}. Illumination guidance is used by ZeroIG~\citep{shi2024zero}, while QuadPrior~\citep{wang2024zero} and HVR-SSLE~\citep{11396663} learn from normal-light images. Within this normal-light-only setting, Dark-Zero adapts synthesis to Solver learning progress and keeps the trained Solver fixed at inference.

\paragraph{Physical degradation synthesis and control.} For normal-light-only synthetic supervision, illumination estimation~\citep{7782813} and sensor-noise models~\citep{foi2008practical} motivate physical degradation pipelines, while domain randomization~\citep{tobin2017domain} supports synthetic-to-real transfer. Image transformations can also adapt through learned control, with \mbox{ReLLIE}~\citep{10.1145/3474085.3475410} optimizing enhancement curves and \mbox{AdaptiveISP} selecting ISP configurations. To guide such control, \mbox{LightDiff}~\citep{li2024light} defines its reinforcement learning reward using downstream perception. In contrast, Dark-Zero formulates its reward from the Solver's learning progress to dynamically control the physical low-light degradations.

\paragraph{Curriculum learning and self-evolving systems.} Curriculum learning~\citep{bengio2009curriculum} organizes training examples by difficulty, whereas learning-progress methods use changes in learner performance as feedback. For generated training data, Learning to Simulate~\citep{ruiz2019learningsimulate} adapts synthesis based on held-out validation performance, while Protagonist Antagonist Induced Regret Environment Design (PAIRED)~\citep{dennis2020emergent} generates environments via inter-agent regret. At the task level, Absolute Zero~\citep{NEURIPS2025_9837dc00}, R-Zero~\citep{ICLR2026_d49b9aac}, and Agent0~\citep{xia2025agent0} couple task generation with learning. This coupling also appears in Dr.~Zero~\citep{yue2026dr} for search and Tool-R0~\citep{acikgoz2026tool} for tool use. Within such self-evolving systems, R-Few~\citep{yu2025guided} introduces limited human supervision as guidance. For LLIE, Dark-Zero uses relative PSNR gains after Solver updates on each synthetic pair to guide degradation synthesis. Each pair provides a known normal-light reference, so reward computation relies entirely on synthetic supervision.

\section{Method}
\label{sec:method}
This section presents Dark-Zero under normal-light-only training. We begin with the problem setup and Challenger-Solver framework. Then, we describe degradation synthesis and supervised restoration. Finally, we describe how Solver learning progress guides PPO updates to the Challenger.

\subsection{Problem Setup}
\label{sec:setup}
For RGB images of height $H$ and width $W$, LLIE aims to map a low-light observation in $[0,1]^{3\times H\times W}$ to an enhanced image in the same space. Here, zero-shot LLIE excludes prior task-specific training on real low-light images under either paired or unpaired supervision.

In this work, we focus on the normal-light-only \methodtype{ZN} workflow introduced in \cref{sec:introduction}. Let $\mathcal D_N$ denote the collection of normal-light training images, and let $x\in[0,1]^{3\times H\times W}$ be a reference image sampled from $\mathcal D_N$. Each training pair consists of a low-light input synthesized from $x$ and the image $x$ itself as the normal-light restoration target. Real low-light evaluation images and any corresponding references are strictly excluded from both training and reward computation. Starting from random initialization, the restoration network learns solely from these synthetic pairs and keeps its parameters fixed throughout evaluation.

\subsection{The Dark-Zero Framework}
\label{sec:overview}
As shown in \cref{fig:darkzero_framework},
Dark-Zero couples a Challenger policy $\pi_\theta$ and a Solver network $f_\phi$, parameterized by $\theta$ and $\phi$, respectively. During training, the Challenger synthesizes a low-light input $y$ from $x$, and the Solver prediction $f_\phi(y)$ is supervised by the reference image $x$. The Solver's measured learning progress on these synthetic pairs supplies the Challenger's reward. At test time, enhancement requires only a single Solver forward pass.

The Challenger operates in the linear RGB domain, motivated by how nonlinear camera ISP tone mapping alters sensor noise statistics. Let $T:[0,1]^{3\times H\times W}\to[0,1]^{3\times H\times W}$ be the invertible, channelwise standard RGB (sRGB) transfer function, and initialize the image state as $z_0=T^{-1}(x)$. Let $T_d=4$ denote the number of degradation stages. Applying the selected transformations to $z_0$ produces $z_{T_d}$, and $y=T(z_{T_d})$ is the corresponding sRGB input to the restoration backbone. A trajectory includes the selected modules, sampled continuous actions, and intermediate image states.

\Cref{alg:training} in \cref{app:schedule} specifies the training loop. Each outer iteration processes $M=4$ normal-light references sequentially, with $K=2$ supervised Solver updates per generated pair. The Challenger policy remains fixed throughout trajectory collection. For each reference, it samples the order and parameters of all $T_d$ degradation operators to synthesize a low-light input. Solver PSNR is measured on the same generated pair before and after its $K$ updates, and the relative improvement defines the trajectory reward. Solver parameters and Adam~\citep{kingma2014adam} optimizer state persist between pairs, so later pairs are evaluated with the updated Solver. After collection, two PPO optimizer steps jointly update the Challenger policy and value head.

\subsection{Degradation Challenger}
\label{sec:challenger}
\label{sec:operators}
We modify AdaptiveISP's sequential controller, originally designed to improve downstream detection, to synthesize low-light inputs from normal-light references. The adapted controller combines masked module selection and Gaussian parameter sampling in a generation policy trained with the Solver's learning-progress reward.

At stage $t\in\{0,\ldots,T_d-1\}$, the policy observes $o_t=(z_t,u_t,t/(T_d-1))$. Here $z_t$ is the image after $t$ degradations, and $u_t\in\{0,1\}^{4}$ records which operators have already been used, with $u_0=0$. For policy encoding, $z_t$ is resized to $64\times64$, while degradation operators are applied at the full training patch resolution.

An action consists of a module identity $m_t\in\{\mathrm{exp},\mathrm{noise},\mathrm{gamma},\mathrm{color}\}$ and a raw continuous vector $a_t\in\mathbb R^{d_{m_t}}$, where the active dimension $d_m$ is one for exposure and gamma and two for noise and color. Given $o_t$, the policy first samples a module from the categorical distribution $m_t\sim\pi^{\rm mod}_\theta(\cdot\mid o_t)$ and then a raw action from the diagonal Gaussian $a_t\sim\pi^{\rm par}_\theta(\cdot\mid o_t,m_t)$. A usage mask excludes previously selected operators, so the four stages apply each operator once in an image-dependent order. A shared convolutional encoder extracts features used to select degradation operators, sample their parameters, and estimate expected rewards, as detailed in \cref{app:architecture}.

To map raw actions to physical controls, we define $\clip(v,l,u)=\min\{\max\{v,l\},u\}$ for $l\leq u$ and apply it elementwise to tensors. A raw action coordinate $a$ is clipped and rescaled to a bounded physical parameter by
\begin{equation}
h(a;a_{\min},a_{\max})
=a_{\min}+\frac{a_{\max}-a_{\min}}{2}\bigl(\clip(a,-1,1)+1\bigr),
\label{eq:mapping}
\end{equation}
where $[a_{\min},a_{\max}]$ is the operator's allowed interval. Policy likelihoods use the cached raw Gaussian action, while rendering uses the physical parameter obtained after clipping and rescaling.

The controls are exposure scale $e$, noise amplitudes $s,r$, gamma exponent $g$, and unnormalized red and blue gains $q_R,q_B$, with the green gain fixed to one. The normalized color-gain vector $\tilde q=(q_R,1,q_B)/\operatorname{mean}(q_R,1,q_B)$ is broadcast across spatial dimensions, where the mean averages the three gains. \Cref{tab:appranges} lists the bounds of all six physical controls before color-gain normalization. To apply these controls, let $\mathcal C(z)=\clip(z,0,1)$ and $\eta(z;s,r)=s\sqrt{\max(z,0)}+r$. The noise tensor $n$ has the same shape as the operator input, with standard Gaussian entries conditional on that input and the sampled noise amplitudes. With gamma floor $\epsilon_\gamma=10^{-8}$ and $\odot$ denoting elementwise multiplication, the four pointwise transformations are
\begin{align}
&D_{\rm exp}(z;e)=\mathcal C(ez), \notag                                      \\
&D_{\rm noise}(z;s,r,n)=\mathcal C\bigl(z+\eta(z;s,r)\odot n\bigr), \label{eq:noise}       \\
&D_{\rm gamma}(z;g)=\mathcal C\bigl(\max(z,\epsilon_\gamma)^g\bigr), \notag                \\
&D_{\rm color}(z;q_R,q_B)=\mathcal C(z\odot\tilde q). \notag
\end{align}

At each stage, the selected transformation maps $z_t$ to $z_{t+1}$, with $\mathcal C$ constraining the output to $[0,1]$. The shot and read terms in \cref{eq:noise} share a Gaussian draw at each pixel and channel, producing a cross term in the conditional variance before clipping, as derived in \cref{app:physical}.

\subsection{Restoration Solver}
\label{sec:solver}
We train each Solver backbone with the same reconstruction objective. For a generated pair $(y,x)$, we clip the Solver's sRGB prediction to $[0,1]$, giving $\hat x=\mathcal C(f_\phi(y))$ for supervision. The objective combines mean absolute error, a structural similarity (SSIM) loss~\citep{wang2004image}, and learned perceptual image patch similarity (LPIPS)~\citep{zhang2018unreasonable}. With weights $\lambda_{\rm L1}=1$, $\lambda_{\rm SSIM}=0.5$, and $\lambda_{\rm LPIPS}=2$, the Solver minimizes
\begin{equation}
\mathcal L_S(\phi;y,x)=\frac{\lambda_{\rm L1}}{3HW}\|\hat x-x\|_1
+\lambda_{\rm SSIM}\bigl(1-\ssim(\hat x,x)\bigr)
+\lambda_{\rm LPIPS}\lpips(\hat x,x),
\label{eq:solverloss}
\end{equation}
where $\|\cdot\|_1$ denotes the entrywise L1 norm, so the first term averages absolute errors over pixels and channels. For SSIM, we use an $11\times11$ Gaussian window with standard deviation 1.5. LPIPS uses a pretrained perceptual network based on AlexNet~\citep{NIPS2012_c399862d}, with sRGB inputs mapped to $[-1,1]$. The perceptual network's weights remain fixed, while gradients propagate through it to train the Solver.

Each Solver update depends on both the network parameters and Adam's optimizer state. Within one outer iteration, let $i\in\{1,\ldots,M\}$ index image pairs in processing order. The full state after $k\in\{0,\ldots,K\}$ updates on pair $i$ is $w_i^{(k)}=(\phi_i^{(k)},\omega_i^{(k)})$, where $\omega_i^{(k)}$ contains Adam's moment estimates and update counter. The $K$ consecutive updates on the fixed pair $(y_i,x_i)$ satisfy $w_i^{(k+1)}=\mathcal U_S(w_i^{(k)};y_i,x_i)$ for $k\in\{0,\ldots,K-1\}$, where $\mathcal U_S$ applies an Adam update using the loss in \cref{eq:solverloss} and gradient norm clipping~\citep{pascanu2013difficulty}. The final parameters and optimizer state carry over to the next pair, including across outer iterations. \Cref{app:schedule} specifies these transitions with an explicit index for outer iterations.

\subsection{Learning-progress policy optimization}
\label{sec:reward}
\label{sec:ppo}
To guide the Challenger's policy updates, we measure the Solver's restoration progress on each generated pair. For pair $i$, the Solver's PSNR values before and after its $K$ supervised updates are given by
\begin{equation}
 P_{{\rm pre},i}=\psnr\bigl(f_{\phi_i^{(0)}}(y_i),x_i\bigr),\qquad
 P_{{\rm post},i}=\psnr\bigl(f_{\phi_i^{(K)}}(y_i),x_i\bigr).
 \label{eq:progressscores}
\end{equation}

Both measurements use the same synthetic pair, with any additional input noise held fixed between evaluations. PSNR is computed in sRGB from positive, finite mean squared error, as defined in \cref{app:schedule}. For finite pre- and post-update scores, the terminal Challenger reward uses a stabilizer $\epsilon_R=0.001$ and is defined as
\begin{equation}
 R_i=\frac{P_{{\rm post},i}-P_{{\rm pre},i}}
 {\max\{P_{{\rm pre},i},\epsilon_R\}}.
 \label{eq:reward}
\end{equation}

This reward measures immediate progress on the generated pair as a stabilized relative change on the logarithmic PSNR scale, with $\epsilon_R$ ensuring a positive denominator for finite scores. Its effect on the restoration of real images is assessed through the matched comparisons in \cref{sec:ablations}.

We adapt the PPO clipped objective by assigning one likelihood ratio and one terminal advantage to each complete degradation trajectory. Here $\theta$ denotes the shared encoder and policy head parameters, and $\psi$ denotes the value head parameters. For trajectory $\tau_i$ collected under frozen parameters $\theta_{\rm old},\psi_{\rm old}$, let $\ell_\theta(\tau_i)$ denote the summed policy log likelihood of its raw actions and module selections, conditioned on the cached observations. The ratio $r_i(\theta)=\exp(\ell_\theta(\tau_i)-\ell_{\theta_{\rm old}}(\tau_i))$ compares the current and frozen policies over all four stages, following the decomposition in \cref{app:ppocache}.

To estimate the advantage of each trajectory, the value predictor $V_{\theta,\psi}(o_{i,0})$ uses its initial observation $o_{i,0}$ to predict the terminal reward. The raw advantage is $A_i=R_i-V_{\theta_{\rm old},\psi_{\rm old}}(o_{i,0})$, where the value baseline remains fixed during the PPO updates. Normalizing these advantages across the $M$ trajectories yields $\widehat A_i$. With clipping threshold $c=0.2$, the policy loss is defined as
\begin{equation}
 \mathcal L_{\rm PG}=-\frac{1}{M}\sum_{i=1}^{M}
 \min\bigl(r_i\widehat A_i,
 \clip(r_i,1-c,1+c)\widehat A_i\bigr).
 \label{eq:policyloss}
\end{equation}

Through $\widehat A_i$, this loss links policy updates to direct measurements of learning progress. We add a value loss and an entropy bonus $\mathcal H$ averaged over cached stages and trajectories. With respective weights $\lambda_V$ and $\lambda_H$, the joint objective is given by
\begin{equation}
 \begin{split}
 \mathcal L_C={}\mathcal L_{\rm PG}-\lambda_H\mathcal H
 +\frac{\lambda_V}{M}\sum_{i=1}^{M}\bigl(V_{\theta,\psi}(o_{i,0})-R_i\bigr)^2.
 \end{split}
 \label{eq:combined}
\end{equation}

The value term trains $V_{\theta,\psi}$ to predict $R_i$, while the entropy term encourages exploration. \Cref{app:ppocache} specifies the trajectory likelihoods, advantage normalization, entropy computation, and remaining optimization details.

\section{Experiments}
\label{sec:experiments}
We evaluate the restoration quality of Dark-Zero on real low-light benchmarks and compare its inference cost with existing methods. To examine how the framework's components and the Solver architecture affect these results, we conduct ablations on Challenger adaptation, the relative PSNR reward, and the Solver loss, alongside evaluations across five Solver backbones.

\subsection{Datasets}
\label{sec:datasets}
We use the $118,287$ natural images in Microsoft COCO 2017 \texttt{train2017} as the reference pool for normal-light-only training. These reference images are center-cropped to $256\times256$ RGB patches and horizontally flipped with a probability of $0.5$. All low-light training inputs are synthesized online from these patches with no COCO annotations.

For paired evaluation, we use LOL-v1 and LOL-v2-real~\citep{yang2021sparse}, which contain $15$ and $100$ test pairs, respectively. The LSRW~\citep{HAI2023103712} test set contains $30$ Huawei and $20$ Nikon test pairs. For unpaired evaluation, we use Unpaired, a collection of $232$ low-light images from DICM, LIME, MEF~\citep{ma2015perceptual}, NPE~\citep{wang2013naturalness}, and VV. \Cref{tab:appup} gives the counts for the constituent Unpaired datasets. Real low-light inputs and any paired normal-light references from these evaluation datasets are strictly excluded from training, reward computation, and parameter updates. Evaluating the same trained Solver across all benchmarks without further fine-tuning tests generalization from synthetic training pairs to real low-light images.

We retain MR.~Illuminate's LOL label for the quoted baselines and use it for the combined LOL-v1 and LOL-v2-real results of Dark-Zero. Within each run, we weight the LOL-v1 and LOL-v2-real dataset means by $15/115$ and $100/115$, respectively. Similarly, the Huawei and Nikon means for LSRW receive weights of $30/50$ and $20/50$, respectively. For Unpaired, we weight each dataset mean by the fraction of collection images belonging to that dataset. These weights give each test image equal weight within its collection, up to rounding of the dataset means.

\subsection{Experimental Setup}
\label{sec:implementation}
We adopt DarkIR-m as the primary Solver. For the main experiments, ablations, and Solver comparisons, we train each Dark-Zero configuration for $5,000$ outer iterations with seeds $0$, $21$, and $42$. Under the update schedule in \cref{sec:overview}, each run comprises $40,000$ Solver optimizer updates. Adam optimizes the Solver and Challenger with constant learning rates of $10^{-4}$ and $3\times10^{-5}$, respectively. Training curves and qualitative illustrations are generated with seed $21$. Additional optimizer settings, PPO hyperparameters, and Solver architecture details are provided in \cref{app:hyperparameters}. Inference FLOPs are measured for a $400\times600$ RGB input, with estimation and counting conventions detailed in \cref{app:flops}.

Before Solver inference, we apply the mean-intensity preprocessing of MR.~Illuminate to evaluation inputs, as specified in \cref{app:evalpreprocess}. Paired evaluation reports PSNR, SSIM, and LPIPS, while Unpaired evaluation reports ILNIQE, BRISQUE, and estimated noise level (NL)~\citep{Chen_2015_ICCV}. Higher values indicate better performance for PSNR and SSIM, whereas lower values are preferable for LPIPS, ILNIQE, BRISQUE, and NL. For Dark-Zero, we first aggregate each metric within each run using the image-count weights defined in \cref{sec:datasets}. We then average the resulting scores across the three random seeds with equal weight and round the mean to four decimal places. Paired dataset scores and collection-level scores for individual runs are reported in \cref{tab:appsolverpairedruns} and \cref{tab:appsolverruns}, respectively. The main-experiment means and standard deviations are summarized in \cref{tab:appmainvariance}.

For the inherited baseline results, we follow the eligibility criteria of MR.~Illuminate, which require publicly available, runnable implementations and any pretrained checkpoints needed for evaluation. \Cref{app:extendedcomparison} documents the official implementations and checkpoint sources for the additional comparisons.

\subsection{Main Results and Analysis}
\label{sec:paired}
\label{sec:transfer}
\label{sec:analysis}
\Cref{tab:main} compares restoration quality and inference cost, with Dark-Zero represented by the DarkIR mean over the three seeds. MR.~Illuminate serves as our primary baseline, from which the 16 external rows are directly quoted, retaining their original training-data labels and reported evaluation protocols. The workflow categories follow \cref{sec:introduction}, while LOL-v1+ denotes LOL-v1 combined with other training data and n/a denotes the absence of a task-specific training dataset. Detailed dataset-level results and comparisons are provided in \cref{app:datasetrecords}, with additional published and reweighted results in \cref{app:extendedcomparison}. \Cref{fig:main_qualitative} shows paired-image examples from LOL and LSRW. Training curves and their analysis appear in \cref{app:trainingcurves,app:degradationcurves}.

\begin{table*}[t]
\begin{minipage}{\linewidth}
\caption{Reported results on LOL, LSRW, and Unpaired using external values and category labels from MR.~Illuminate and the three-run DarkIR mean for Dark-Zero. Within \methodtype{Z}, \textbf{(P)} denotes methods with \textbf{pretrained diffusion models}, and \textbf{(I)} denotes methods with \textbf{iterative optimization of model parameters for each test image}. Red boldface, black boldface, and underlining mark the best, second-best, and third-best quality values per column, respectively, including ties, while $\dagger$ denotes the area-based FLOPs estimate of DarkIR.}
\label{tab:main}
\centering
\scriptsize
\setlength{\tabcolsep}{2.4pt}
\resizebox{\linewidth}{!}{%
\begin{tabular}{@{}cllcccccccccc@{}}
\toprule
\tworowhead{Type} & \tworowhead{Method} & \tworowhead{Train data} & \tworowhead{FLOPs $\downarrow$} & \multicolumn{3}{c}{LOL} & \multicolumn{3}{c}{LSRW} & \multicolumn{3}{c}{Unpaired}\\
\cmidrule(lr){5-7}\cmidrule(lr){8-10}\cmidrule(lr){11-13}
& & & & PSNR $\uparrow$ & SSIM $\uparrow$ & LPIPS $\downarrow$ & PSNR $\uparrow$ & SSIM $\uparrow$ & LPIPS $\downarrow$ & ILNIQE $\downarrow$ & BRISQUE $\downarrow$ & NL $\downarrow$\\
\midrule
\methodtype{U} & CLIP-LIT~\citep{liang2023iterative} & Own data & $1.3\!\times\!10^{11}$ & 14.8179 & 0.5243 & 0.3706 & 13.4835 & 0.4051 & 0.3533 & 28.1403 & 26.0421 & 1.8638\\
\methodtype{U} & Zero-DCE~\citep{guo2020zero} & Own data & $3.8\!\times\!10^{10}$ & 17.6417 & 0.5717 & 0.3154 & 15.8680 & 0.4506 & 0.3152 & 26.8335 & 23.1635 & 1.5336\\
\methodtype{U} & Zero-DCE++~\citep{9369102} & Own data & $4.5\!\times\!10^{7}$ & 17.0175 & 0.4439 & 0.3145 & 16.2453 & 0.4568 & 0.3273 & 25.9503 & \underline{17.1177} & 1.2573\\
\methodtype{U} & NeRCo~\citep{yang2023implicit} & LOL-v1 & $1.7\!\times\!10^{12}$ & \underline{21.0728} & 0.7248 & 0.2591 & 17.3814 & 0.5303 & 0.5182 & 28.1309 & 30.0167 & 0.8358\\
\methodtype{U} & ZeroIG~\citep{shi2024zero} & LOL-v1 & $7.7\!\times\!10^{10}$ & 18.6589 & 0.7496 & 0.2415 & 16.4431 & 0.5087 & 0.3764 & 27.5778 & 26.2227 & 1.4861\\
\methodtype{U} & GEFU~\citep{11446150} & LOL-v1+ & $4.2\!\times\!10^{12}$ & 20.8308 & 0.7898 & 0.2078 & \textcolor{red}{\textbf{18.7498}} & \underline{0.5421} & \textcolor{red}{\textbf{0.2806}} & \textbf{24.1183} & 17.8064 & 0.6349\\
\methodtype{U} & EnlightenGAN~\citep{jiang2021enlightengan} & LOL-v1+ & $2.4\!\times\!10^{11}$ & 18.4888 & 0.6722 & 0.3105 & 17.0811 & 0.4705 & 0.3273 & 24.7138 & 18.9759 & 0.9667\\
\methodtype{U} & PairLIE~\citep{fu2023learning} & LOL-v1+ & $1.6\!\times\!10^{11}$ & 19.6999 & 0.7737 & 0.2353 & 17.6104 & 0.5190 & 0.3309 & 26.7961 & 31.0421 & 1.6407\\
\methodtype{U} & ZeroIG~\citep{shi2024zero} & LSRW & $7.7\!\times\!10^{10}$ & 17.5677 & 0.4778 & 0.3799 & 16.7516 & 0.5010 & 0.4000 & 34.2056 & 34.8248 & 2.4482\\
\methodtype{U} & QuadPrior~\citep{wang2024zero} & COCO & $4.2\!\times\!10^{13}$ & 20.3016 & \underline{0.8096} & \underline{0.2032} & 16.9469 & \textcolor{red}{\textbf{0.5601}} & 0.3824 & \underline{24.4373} & 18.1736 & \underline{0.4215}\\
\midrule
\methodtype{Z} & RRDNet~\citep{zhu2020zero} & n/a (I) & $6.1\!\times\!10^{13}$ & 13.5719 & 0.4791 & 0.3238 & 13.4272 & 0.3918 & 0.3358 & 26.6567 & 17.9583 & 1.1570\\
\methodtype{Z} & CoLIE~\citep{chobola2024fast} & n/a (I) & $5.2\!\times\!10^{12}$ & 14.8999 & 0.4985 & 0.3268 & 14.0013 & 0.4053 & 0.3424 & 26.8185 & 18.9680 & 0.9642\\
\methodtype{Z} & ExCNet~\citep{zhang2019zero} & n/a (I) & $1.7\!\times\!10^{7}$ & 16.2972 & 0.4589 & 0.3745 & 15.7021 & 0.4098 & 0.3375 & 27.3933 & 19.2416 & 1.8234\\
\methodtype{Z} & TAO~\citep{gou2023test} & n/a (P) & $4.7\!\times\!10^{15}$ & 19.1807 & 0.6065 & 0.3897 & 15.6891 & 0.4306 & 0.7023 & 42.0846 & 42.1383 & \textbf{0.3840}\\
\methodtype{Z} & FourierDiff~\citep{lv2024fourier} & n/a (P) & $8.5\!\times\!10^{14}$ & 16.9525 & 0.6039 & 0.2934 & 15.6251 & 0.4610 & 0.3207 & 25.9272 & 26.5694 & 1.2206\\
\methodtype{Z} & MR.~Illuminate~\citep{cho2026mr} & n/a (P) & $5.8\!\times\!10^{13}$ & \textbf{21.7393} & \textbf{0.8152} & \textbf{0.1771} & \underline{17.6634} & 0.5185 & \textbf{0.2829} & 25.4181 & \textbf{16.1843} & \textcolor{red}{\textbf{0.3794}}\\
\midrule
\methodtype{ZN} & Dark-Zero & COCO & $5.3\!\times\!10^{10}\,{}^{\dagger}$ & \textcolor{red}{\textbf{22.8392}} & \textcolor{red}{\textbf{0.8312}} & \textcolor{red}{\textbf{0.1602}} & \textbf{17.9566} & \textbf{0.5515} & \underline{0.3073} & \textcolor{red}{\textbf{23.5568}} & \textcolor{red}{\textbf{15.7642}} & 0.6824\\
\bottomrule
\end{tabular}}
\end{minipage}
\par\vspace{\floatsep}
\begin{minipage}{\linewidth}
\captionsetup{type=figure}
\centering
\includegraphics[width=\linewidth]{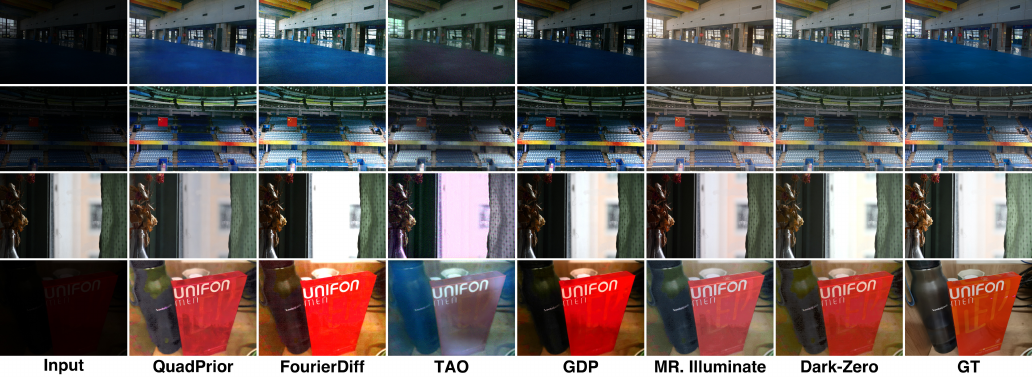}
\caption{Qualitative comparison of six enhancement methods on paired LOL and LSRW images, highlighting differences in brightness recovery and color fidelity. The first and last columns show the low-light inputs and ground-truth references, respectively.}
\label{fig:main_qualitative}
\end{minipage}
\end{table*}

On LOL, Dark-Zero reaches a mean PSNR of $22.8392$\,dB, exceeding the reported score of MR.~Illuminate by $1.0999$\,dB. Its mean SSIM is higher by $0.0160$, and its mean LPIPS is lower by $0.0169$. Dark-Zero thus achieves state-of-the-art reported scores on LOL among methods trained without real low-light images, outperforming QuadPrior and the \methodtype{Z} methods in \cref{tab:main}. On LSRW, the mean PSNR and SSIM of Dark-Zero exceed the reported values of MR.~Illuminate by $0.2932$\,dB and $0.0330$, respectively.

On Unpaired, Dark-Zero has the lowest mean ILNIQE and BRISQUE in \cref{tab:main}, at $23.5568$ and $15.7642$, respectively. These scores improve upon the reported values of MR.~Illuminate by $1.8613$ and $0.4201$. Its ILNIQE is also $0.5615$ below GEFU's $24.1183$, the best score among the \methodtype{U} methods in the table. The extended comparison in \cref{app:extendedcomparison} further reports a lower BRISQUE of $15.2528$ for CLODE, reconstructed using image-count weights. 

In the paired-image examples in \cref{fig:main_qualitative}, Dark-Zero recovers details in the dark hall and seating areas while retaining the visible structure of backlit objects. Together, these quantitative results and visual examples demonstrate the generalization of Dark-Zero from normal-light-only training to real low-light images. Dark-Zero not only achieves state-of-the-art performance in zero-shot LLIE but also competes with leading unsupervised methods under this training constraint.

At inference, the main DarkIR configuration has an estimated cost of $5.31\times10^{10}$ FLOPs, about $1/1{,}100$ of MR.~Illuminate's reported $5.8\times10^{13}$ FLOPs and $1/790$ of QuadPrior's $4.2\times10^{13}$. \Cref{app:flops} provides the sources, estimation procedure, and counting conventions for the FLOPs comparison.

\begin{table}[!t]
\begin{minipage}{\linewidth}
\caption{LOL ablations with DarkIR, reporting means over the three seeds. Red boldface marks the best mean for each metric.}
\label{tab:ablations}
\centering
\footnotesize
\setlength{\tabcolsep}{2.4pt}
\renewcommand{\arraystretch}{1.12}
\begin{tabular*}{1\linewidth}{@{\extracolsep{\fill}}lccc@{}}
\toprule
Variant & PSNR $\uparrow$ & SSIM $\uparrow$ & LPIPS $\downarrow$\\
\midrule
Full model & \textcolor{red}{\textbf{22.8392}} & \textcolor{red}{\textbf{0.8312}} & \textcolor{red}{\textbf{0.1602}}\\
\midrule
Absolute PSNR gain & 22.1020 & 0.8099 & 0.1891\\
L1-only Solver loss & 21.4208 & 0.7259 & 0.2674\\
Uniform random Challenger & 21.7617 & 0.8159 & 0.1726\\
\midrule
Frozen Challenger (1) & 21.4808 & 0.7403 & 0.2310\\
Frozen Challenger (1,000) & 21.2865 & 0.7956 & 0.1987\\
Frozen Challenger (3,000) & 22.1934 & 0.8266 & 0.1661\\
Frozen Challenger (pretrained) & 22.0461 & 0.8260 & 0.1604\\
\bottomrule
\end{tabular*}
\end{minipage}
\par\vspace{\floatsep}
\begin{minipage}{\linewidth}
\captionsetup{type=figure}
\centering
\includegraphics[width=\linewidth]{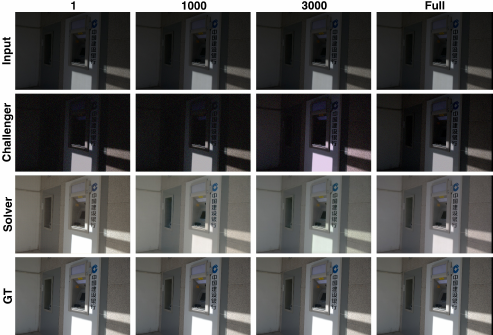}
\caption{Qualitative comparison of Frozen Challenger variants and the full model on LOL-v2-real. Columns show the three frozen variants and the full model. Rows show the low-light input, Challenger synthesis from the ground-truth reference, Solver restoration of the low-light input, and the reference, respectively.}
\label{fig:ablation_qualitative}
\end{minipage}
\end{table}

\subsection{Ablation Study}
\label{sec:ablations}
\Cref{tab:ablations} compares the full model and seven ablation variants using DarkIR over the three seeds under the same LOL evaluation protocol. Absolute PSNR gain replaces the relative reward with $R_i^{\mathrm{abs}}=P_{\mathrm{post},i}-P_{\mathrm{pre},i}$, and L1-only Solver loss retains only the L1 term in \cref{eq:solverloss}. Uniform random Challenger randomizes the degradation order and samples physical parameters uniformly within their allowed ranges.

To test continued Challenger adaptation, Frozen Challenger (1 / 1,000 / 3,000) stops Challenger updates after the indicated number of outer iterations and continues Solver training within the same run. Frozen Challenger (pretrained) instead uses the final Challenger from a completed training run with the corresponding seed and holds it fixed throughout a new Solver training run. Each frozen policy continues sampling degradation actions with its fixed network parameters. Per-seed results and sample standard deviations are reported in \cref{tab:appablationruns}. \Cref{fig:ablation_qualitative} compares the synthetic degradations and restorations from the three freezing schedules and the full model on a single scene.

The full model achieves the highest mean PSNR and SSIM and the lowest mean LPIPS among all variants in \cref{tab:ablations}. Compared with Uniform random Challenger, it gains $1.0775$\,dB in mean PSNR, supporting the use of learned challenge generation. It also gains $0.7372$\,dB over Absolute PSNR gain and $1.4184$\,dB over L1-only Solver loss in mean PSNR, supporting the contributions of relative feedback and the mixed reconstruction objective.

Across the three variants frozen during training, later freezing yields higher mean SSIM and lower mean LPIPS, while mean PSNR peaks for Frozen Challenger (3,000). The full model also improves mean PSNR by $0.7931$\,dB and mean SSIM by $0.0052$ over Frozen Challenger (pretrained), whereas mean LPIPS is similar, at $0.1602$ for the full model and $0.1604$ for the pretrained policy. Together, these comparisons support continued Challenger-Solver co-evolution over training a new Solver with a fixed pretrained Challenger in this setting. In the Solver-restoration row of \cref{fig:ablation_qualitative}, the full model retains the shadowed wall and sign colors of the ground-truth reference. The variants frozen earlier brighten these regions more strongly, reflecting the effect of the freezing schedule on restoration.

\subsection{Applicability across Solvers}
\label{sec:multisolver}
To evaluate applicability across restoration architectures, we train separate Dark-Zero models with DarkIR, CIDNet, Restormer~\citep{zamir2022restormer}, SNR-Aware, and LLFormer~\citep{wang2023ultra} as the Solver. All five follow the training protocol in \cref{sec:implementation}. The experiments use native image resolution in evaluation mode, with shared input preprocessing, dataset weights from \cref{sec:datasets}, and zero additional input noise. \Cref{tab:solvers} reports mean LOL scores over the three seeds and inference FLOPs, reusing the main-experiment runs for DarkIR. Individual run scores and sample standard deviations are shown in \cref{app:solverrunstats}, with FLOPs sources and counting conventions detailed in \cref{app:flops}.

\begin{table}[t]
\caption{LOL Solver comparison using means over the three seeds, ordered by inference FLOPs. The $\dagger$ and $\ddagger$ markers denote the DarkIR area estimate and official-code operator counts. Red boldface and underlining mark the best and second-best mean quality values, respectively, including ties.}
\label{tab:solvers}
\centering
\footnotesize
\setlength{\tabcolsep}{1.7pt}
\begin{tabular*}{1\linewidth}{@{\extracolsep{\fill}}lcccc@{}}
\toprule
Solver & FLOPs $\downarrow$ & PSNR $\uparrow$ & SSIM $\uparrow$ & LPIPS $\downarrow$\\
\midrule
DarkIR (main) & $5.3\!\times\!10^{10}\,{}^{\dagger}$ & \textcolor{red}{\textbf{22.8392}} & \textcolor{red}{\textbf{0.8312}} & \underline{0.1602}\\
CIDNet & $6.0\!\times\!10^{10}$ & \underline{22.7578} & 0.8047 & 0.1925\\
SNR-Aware & $1.9\!\times\!10^{11}$ & 22.3022 & 0.8071 & 0.2087\\
LLFormer & $4.1\!\times\!10^{11}\,{}^{\ddagger}$ & 22.3650 & 0.8260 & 0.1776\\
Restormer & $1.1\!\times\!10^{12}\,{}^{\ddagger}$ & 22.6557 & \underline{0.8294} & \textcolor{red}{\textbf{0.1564}}\\
\bottomrule
\end{tabular*}
\end{table}

Across the five tested Solvers, mean LOL PSNR ranges from $22.3022$ to $22.8392$\,dB. DarkIR achieves the highest mean PSNR and SSIM while also having the lowest reported inference FLOPs in \cref{tab:solvers}. Restormer attains the lowest mean LPIPS, reducing it from DarkIR's $0.1602$ to $0.1564$ at approximately $21$ times the reported FLOPs. Thus, Dark-Zero supports real low-light enhancement beyond the main DarkIR architecture, with all four alternative Solvers achieving mean LOL PSNR above $22$\,dB.

\section{Conclusion}
\label{sec:conclusion}
Dark-Zero adapts synthetic supervision for zero-shot LLIE using only normal-light training images. The Challenger learns physical degradations from the Solver's relative PSNR gains, linking synthesis to restoration progress. Dark-Zero achieves state-of-the-art reported LOL scores among the compared methods trained without real low-light images, averaged over three runs. Ablations support continued Challenger adaptation, relative feedback, and the mixed reconstruction loss. Five-Solver experiments demonstrate applicability across architectures, with backbone choice affecting quality and inference cost. Inference uses one forward pass via the fixed Solver, ensuring efficiency.

\bibliographystyle{plainnat-no-url}
\bibliography{gsref}
\clearpage
\appendix
\crefalias{section}{appendix}
\crefalias{subsection}{appendix}
\crefalias{subsubsection}{appendix}
\etocdepthtag.toc{appendix}
\numberwithin{equation}{section}
\setcounter{algorithm}{0}
\renewcommand{\thealgorithm}{S\arabic{algorithm}}
{\centering
\Large Supplementary Material\par
}
\vspace{1em}
\raggedbottom

\begingroup
\small
\hypersetup{linkcolor=black}
\setlength{\parskip}{0pt}
\etocsettagdepth{main}{none}
\etocsettagdepth{appendix}{subsection}
\etocsettocstyle{\section*{Appendix Contents}}{}
\tableofcontents
\endgroup
\clearpage

\noindent
\Cref{app:implementation,app:schedule,app:physical} detail the implementation, training updates, and physical degradation model of Dark-Zero. The evaluation protocol and experimental records appear in \cref{app:evaluation,app:controls,app:extendedcomparison}. Additional visual comparisons in \cref{app:qualitative} complement these records. All Dark-Zero training visualizations and qualitative results in this supplement use training seed~21.

\section{Implementation details}
\label{app:implementation}
The Challenger chooses a degradation operator and its parameters at each stage. We describe how the network makes these choices, then give the optimization and loss settings used in \cref{sec:implementation}.

\subsection{State encoding and policy heads}
\label{app:architecture}
At stage $t\in\{0,1,2,3\}$, the Challenger observes the current linear-RGB image $z_t$ and a four-bit vector $u_t$ indicating which operators have already been applied. We repeat each bit across a $64\times64$ grid to obtain the four channels of $U_t$ and fill a fifth channel $J_t$ with the normalized stage index $t/3$. The encoder input is
\begin{equation}
 b_t=\operatorname{concat}\bigl(\operatorname{resize}(z_t),U_t,J_t\bigr),
 \label{eq:appencoderinput}
\end{equation}
where $\operatorname{resize}$ denotes bilinear resizing to $64\times64$ and concatenation stacks the maps along the channel dimension. The resulting tensor $b_t\in\mathbb R^{8\times64\times64}$ is used to choose the next action. The degradation itself is applied at the original training patch resolution.

The encoder feeds the module, parameter, and value heads in \cref{tab:apparch}. Each head uses rectified linear unit (ReLU)~\citep{nair2010rectified} activations. The module head scores all four operators and masks those already applied. The parameter head predicts two means per operator and has a trainable log standard deviation for each coordinate. Exposure and gamma each use one coordinate, and noise and color cast each use two, for six controls per trajectory. The selected operator determines which coordinates contribute to its action likelihood and entropy.

\begin{table}[!htbp]
\caption{Challenger layout with four degradation operators. Outputs use channel-first dimensions. Convolutions use padding 1.}
\label{tab:apparch}
\centering\footnotesize
\setlength{\tabcolsep}{2.4pt}
\begin{tabular*}{1\linewidth}{@{\extracolsep{\fill}}>{\raggedright\arraybackslash}m{.25\linewidth}>{\raggedright\arraybackslash}m{.48\linewidth}>{\centering\arraybackslash}m{.20\linewidth}@{}}
\toprule
Component & Operation & Output\\
\midrule
State & Image, usage, stage & $8\times64\times64$\\
Encoder 1 & Convolution $4\times4$, stride 2, 32 channels & $32\times32\times32$\\
Encoder 2 & Convolution $4\times4$, stride 2, 64 channels & $64\times16\times16$\\
Encoder 3 & Convolution $4\times4$, stride 2, 128 channels & $128\times8\times8$\\
Pooling & Global average pooling & 128\\
Module head & Linear 128, ReLU, linear & 4 logits\\
Parameter head & Linear 128, ReLU, linear & $4\times2$ means\\
Value head & Linear 128, ReLU, linear & 1 scalar\\
\bottomrule
\end{tabular*}
\end{table}

The encoder in \cref{tab:apparch} reduces the image representation to 128 pooled features shared by all three heads. Each convolution uses leaky rectified linear unit (LeakyReLU)~\citep{maas2013rectifier} activations with negative slope 0.2. In the parameter head's final layer, the weights are initialized from a normal distribution with standard deviation 0.01 and the biases are set to zero. The Gaussian log standard deviations start at $-1$ and are bounded to $[-4,0]$, corresponding to standard deviations in $[\exp(-4),1]$.

For a selected module $m$, let $d_m\in\{1,2\}$ be its active dimension, $\mu_{\theta,m}(o_t)\in\mathbb R^{d_m}$ its predicted mean, and $\sigma_m\in(0,\infty)^{d_m}$ its trainable standard deviation vector, included in $\theta$. With $I_{d_m}$ the identity matrix and $\odot$ denoting elementwise multiplication, training samples the raw action by
\begin{equation}
 a_t=\mu_{\theta,m}(o_t)+\sigma_m\odot\nu_t,
 \qquad \nu_t\sim\mathcal N(0,I_{d_m}).
 \label{eq:apprawaction}
\end{equation}
The renderer maps the raw action to physical controls using \cref{eq:mapping}. The PPO buffer retains the raw Gaussian sample for likelihood evaluation. Write $\mu_{m,j}=\mu_{\theta,m,j}(o_t)$ and $\sigma_{m,j}$ for the mean and standard deviation of active coordinate $j$, respectively, and $\ell_t^{\rm par}=\log\pi^{\rm par}_\theta(a_t\mid o_t,m)$ for the action's log likelihood. Then
\begin{equation}
 \ell_t^{\rm par}=-\frac12\sum_{j=1}^{d_m}
 \bigl[\frac{(a_{t,j}-\mu_{m,j})^2}{\sigma_{m,j}^2}
 +\log(2\pi\sigma_{m,j}^2)\bigr].
 \label{eq:appgaussian}
\end{equation}
Clipping can map distinct raw actions to the same boundary parameter. Thus the log density in \cref{eq:appgaussian} is evaluated at the cached sample, before the action mapping.

\subsection{Optimizer and loss configuration}
\label{app:hyperparameters}
Each training run uses one NVIDIA H200 graphics processing unit (GPU). For the schedule with $M=4$ pairs per outer iteration and $K=2$ Solver updates per pair, \cref{tab:appoptim} gives the optimizer settings, including separate gradient norm limits for the Solver and Challenger. The three epsilon values stabilize Adam, advantage normalization, and the reward denominator, respectively.

\begin{table}[t]
\caption{Optimization configuration for the manuscript's training schedule. EMA denotes exponential moving average.}
\label{tab:appoptim}
\centering\footnotesize
\setlength{\tabcolsep}{2.4pt}
\begin{tabular*}{1\linewidth}{@{\extracolsep{\fill}}>{\raggedright\arraybackslash}m{.60\linewidth}>{\centering\arraybackslash}m{.37\linewidth}@{}}
\toprule
Setting & Value\\
\midrule
Solver / Challenger optimizer & Adam / Adam\\
Solver learning rate & $10^{-4}$\\
Challenger learning rate & $3\times10^{-5}$\\
Adam $(\beta_1,\beta_2)$ & $(0.9,0.999)$\\
Adam epsilon & $10^{-8}$\\
Weight decay / AMSGrad~\citep{reddi2019convergence} & 0 / disabled\\
Learning-rate schedule & Constant\\
Solver gradient norm limit & 1.0\\
Challenger gradient norm limit & 0.5\\
PPO trajectories / effective mini-batch size & 4 / 4\\
PPO optimizer steps per outer iteration & 2\\
PPO ratio clip range & 0.2\\
Value-loss coefficient & 0.5\\
Entropy coefficient & 0.01\\
Advantage normalization & Enabled\\
Normalization epsilon & $10^{-8}$\\
Reward denominator epsilon & 0.001\\
Solver EMA / progress rollback & Disabled / disabled\\
\bottomrule
\end{tabular*}
\end{table}

For the Solver loss in \cref{eq:solverloss}, predictions are clipped to $[0,1]$ in sRGB, and the L1 term averages absolute errors over pixels and channels. The SSIM training loss uses an $11\times11$ Gaussian window with standard deviation 1.5 and zero padding of 5. Its stability constants are $C_1=0.01^2$ and $C_2=0.03^2$, and scores are averaged over channels and spatial positions.

The perceptual loss maps sRGB inputs to $[-1,1]$ and uses pretrained AlexNet-based LPIPS in evaluation mode. Its parameters remain fixed while gradients with respect to the Solver output are retained. Benchmark metrics follow the separate evaluation protocol in \cref{app:evalpreprocess}.

The primary DarkIR-m configuration uses width 32, with encoder blocks $[1,2,3]$ and decoder blocks $[3,1,1]$. It also has two middle encoder blocks and two middle decoder blocks, with dilations $[1,4,9]$ and an additional depthwise convolution. CIDNet, Restormer~\citep{zamir2022restormer}, SNR-Aware~\citep{xu2022snr}, and LLFormer~\citep{wang2023ultra} use their original standard network architectures.

\section{Sequential update accounting}
\label{app:schedule}
Each outer iteration collects four degradation trajectories while updating the Solver twice per generated pair. \Cref{alg:training} makes this ordering explicit, and the state equations below specify how Solver weights and optimizer moments persist across pairs.

\begin{algorithm}[!t]
\caption{One outer iteration of Dark-Zero training}
\label{alg:training}
\begin{algorithmic}[1]
\Require Challenger $\pi_\theta$, value head parameters $\psi$, Solver $f_\phi$, persistent optimizers, $M=4$, $K=2$, $T_d=4$
\State Cache $\theta_{\rm old}\gets\theta$, $\psi_{\rm old}\gets\psi$
\State Initialize buffer $\mathcal B\gets\varnothing$
\For{$i=1,\ldots,M$}
  \State Sample a normal-light reference $x_i$ from $\mathcal D_N$
  \State Sample $T_d$-stage trajectory $\tau_i$ under $\pi_{\theta_{\rm old}}$ and render $y_i$
  \State Cache $V_{\theta_{\rm old},\psi_{\rm old}}(o_{i,0})$ and $\ell_{\theta_{\rm old}}(\tau_i)$
  \State Measure $P_{{\rm pre},i}$ on $(y_i,x_i)$
  \For{$k=1,\ldots,K$}
    \State Update $\phi$ on $(y_i,x_i)$ using \cref{eq:solverloss}
  \EndFor
  \State Measure $P_{{\rm post},i}$ with matched randomness
  \State Compute $R_i$ using \cref{eq:reward}
  \State Append $(\tau_i,R_i)$ and cached old values to $\mathcal B$
\EndFor
\For{$j=1,2$}
  \State Update Challenger and value head on $\mathcal B$ using \cref{eq:combined}
\EndFor
\State Carry the updated networks and both optimizer states into the next iteration
\end{algorithmic}
\end{algorithm}

Throughout collection in \cref{alg:training}, the behavior policy stays fixed and the value estimates are cached before any Challenger updates. The Solver changes after each pair, and the Challenger receives two joint PPO updates after all four rewards have been measured.

For the progress measurements in \cref{alg:training}, let $v\in\mathbb R^{3\times H\times W}$ be an sRGB prediction and $x$ its unit-range reference. The mean squared error (MSE) is $\operatorname{MSE}(v,x)=(3HW)^{-1}\|v-x\|_2^2$, and $\psnr(v,x)=-10\log_{10}\operatorname{MSE}(v,x)$ for positive, finite MSE.

\subsection{Training dynamics}
\label{app:trainingcurves}
\begin{figure}[!htbp]
\centering
\includegraphics[width=0.5\linewidth]{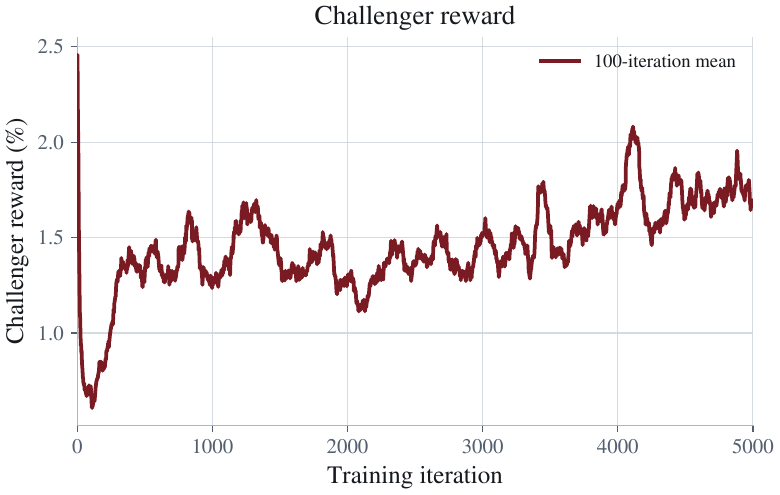}
\par{\small (a) Challenger reward\par}
\smallskip
\includegraphics[width=0.5\linewidth]{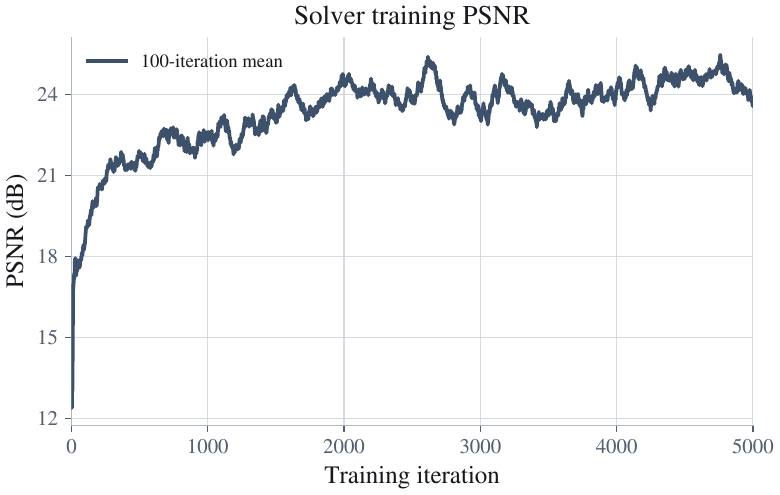}
\par{\small (b) Solver training PSNR\par}
\smallskip
\includegraphics[width=0.5\linewidth]{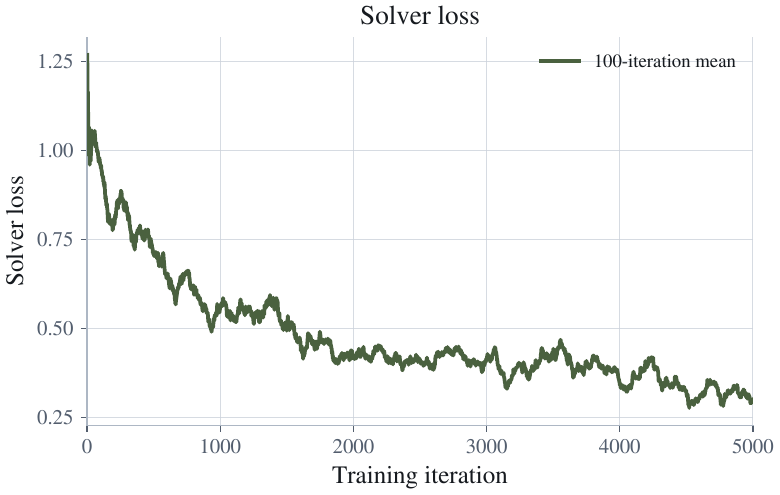}
\par{\small (c) Solver loss\par}
\caption{Training dynamics over 5,000 outer iterations. Panels show the Challenger learning-progress reward in percent, Solver training PSNR in dB, and Solver training loss. Curves use trailing 100-iteration means, with all available iterations used before iteration 100.}
\label{fig:apptrainingcurves}
\end{figure}

\Cref{fig:apptrainingcurves} records the Challenger reward, Solver PSNR, and Solver loss during co-adaptation, with PSNR measured on synthetic training pairs. For a fixed positive PSNR gain and a pre-update score above $\epsilon_R$, \cref{eq:reward} gives a larger reward at a lower pre-update score. This dependence offers one explanation for the early reward decline.

The reward curve in \cref{fig:apptrainingcurves} tracks a normalized within-pair improvement. Training PSNR and loss track reconstruction quality on the synthetic examples sampled at each Solver state.

\FloatBarrier
\subsection{Evolution of synthetic challenges}
\label{app:degradationcurves}
\begin{figure}[tbp]
\centering
\includegraphics[width=0.5\linewidth]{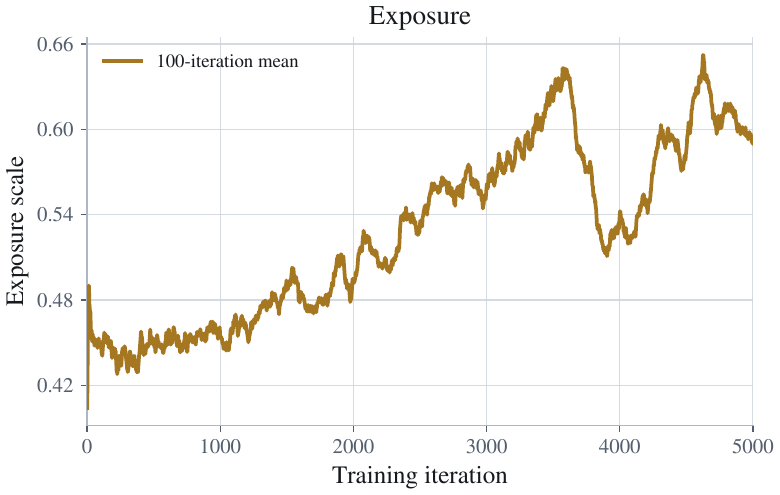}
\caption{Recorded exposure scale during Challenger training.}
\label{fig:appdegradationexposure}
\end{figure}

\begin{figure}[tbp]
\centering
\includegraphics[width=0.5\linewidth]{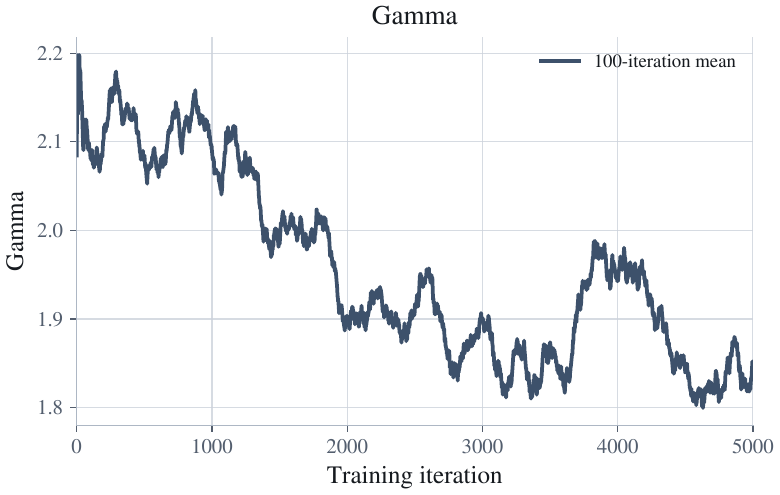}
\caption{Recorded gamma exponent during Challenger training.}
\label{fig:appdegradationgamma}
\end{figure}

\begin{figure}[tbp]
\centering
\includegraphics[width=0.5\linewidth]{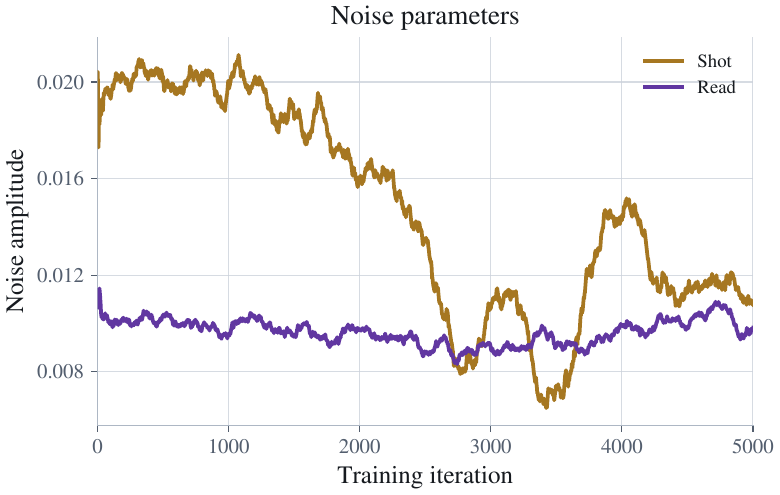}
\caption{Recorded shot and read noise amplitudes during Challenger training.}
\label{fig:appdegradationnoise}
\end{figure}

\begin{figure}[tbp]
\centering
\includegraphics[width=0.5\linewidth]{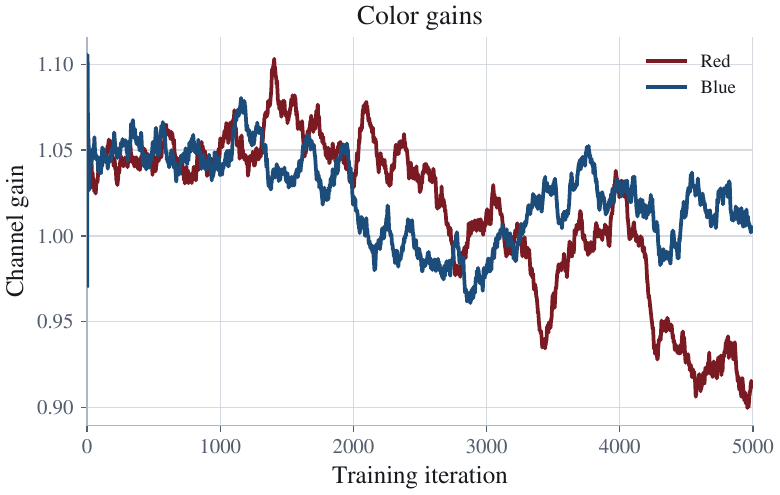}
\caption{Recorded red and blue channel gains during Challenger training.}
\label{fig:appdegradationcolor}
\end{figure}

\begin{figure}[!htbp]
\centering
\includegraphics[width=0.5\linewidth]{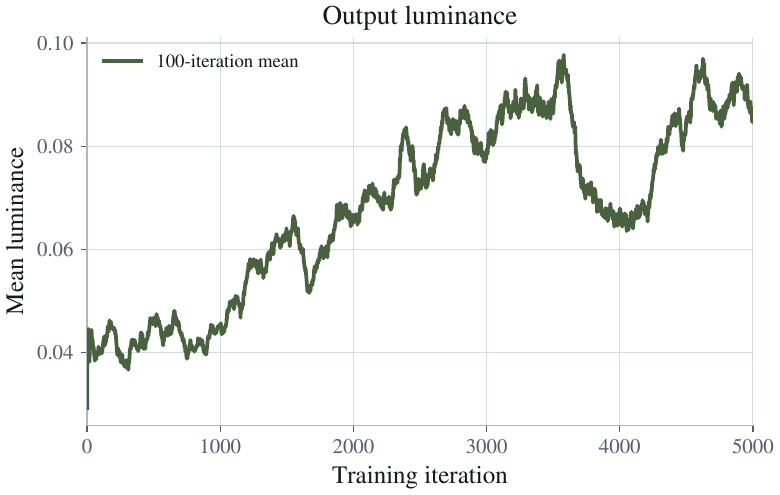}
\par{\small (a) Mean luminance\par}
\smallskip
\includegraphics[width=0.5\linewidth]{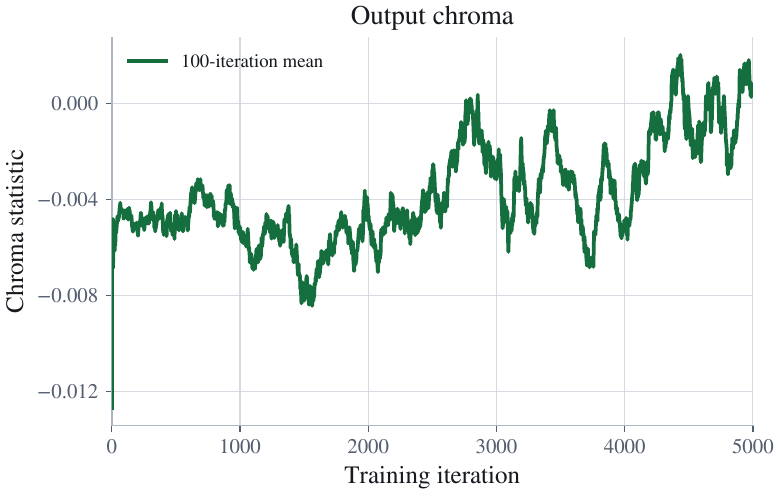}
\par{\small (b) Chroma\par}
\smallskip
\includegraphics[width=0.5\linewidth]{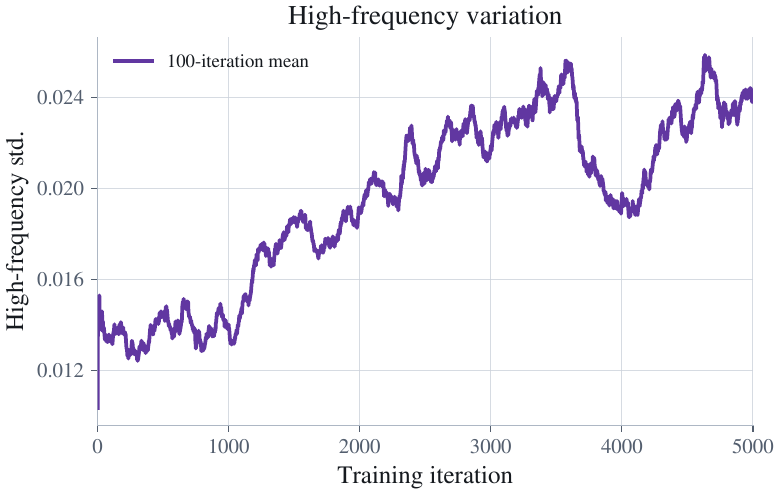}
\par{\small (c) High-frequency standard deviation\par}
\caption{Statistics of the synthesized training inputs over 5,000 outer iterations. Panels show mean luminance, the recorded chroma statistic, and high-frequency standard deviation. Curves use trailing 100-iteration means, with all available iterations used before iteration 100.}
\label{fig:appdegradationstatistics}
\end{figure}

\Cref{fig:appdegradationexposure,fig:appdegradationgamma,fig:appdegradationnoise,fig:appdegradationcolor} track exposure, gamma, noise amplitudes, and color gains during training. The accompanying synthesized-image statistics in \cref{fig:appdegradationstatistics} describe output luminance, chroma, and high-frequency standard deviation. As in \cref{fig:apptrainingcurves}, all curves use trailing 100-iteration means, with all available iterations used before iteration 100.

Between the first and last 100 iterations, the mean exposure scale rises from 0.4506 to 0.5903 in \cref{fig:appdegradationexposure}, and the mean gamma exponent falls from 2.0874 to 1.8504 in \cref{fig:appdegradationgamma}. For an input intensity in $(0,1)$ where the numerical floor and clipping are inactive, either change reduces attenuation in its corresponding operator in \cref{sec:operators}. Mean output luminance also rises from 0.0406 to 0.0858 in \cref{fig:appdegradationstatistics}(a), showing that the sampled training inputs become brighter overall.

The shot amplitude in \cref{fig:appdegradationnoise} decreases overall, while the read amplitude remains close to 0.01. Red and blue gains in \cref{fig:appdegradationcolor} follow distinct trajectories, with a larger reduction in red gain toward the end of training, while the mean chroma statistic in \cref{fig:appdegradationstatistics}(b) remains near zero. These changes are non-monotonic. Around iterations 3,500-4,000, for example, exposure and output luminance temporarily decrease as gamma increases.

The output high-frequency standard deviation in \cref{fig:appdegradationstatistics}(c) increases overall while the shot amplitude decreases. This output statistic depends on scene structure and the effects of the composed degradation operators, providing a diagnostic of the synthesized images alongside the individual operator parameters.

\begin{figure*}[!tp]
  \centering
  \includegraphics[width=\linewidth]{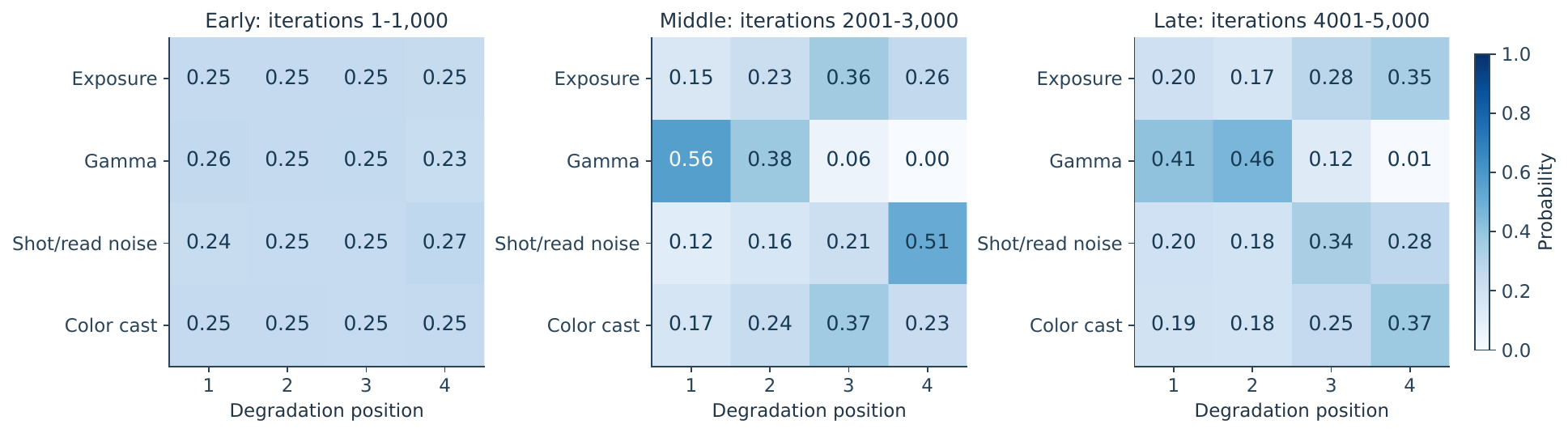}
  \caption{Operator positions during Challenger training. Each panel shows empirical frequencies over 4,000 sampled trajectories in a 1,000-iteration window of the degradation-logging run. Each trajectory uses every operator once.}
  \label{fig:operator_positions}
\end{figure*}

The three training windows in \cref{fig:operator_positions} supplement the parameter traces with the frequency of each operator at each stage.

The position distributions are nearly uniform in the early window. Gamma appears in the first two stages of approximately 51\% of early trajectories, approximately 94\% of middle trajectories, and approximately 87\% of late trajectories. The ordering continues to change after the middle window. For example, shot/read noise is applied last in approximately 51\% of middle trajectories and approximately 28\% of late trajectories.

\begin{figure*}[!tp]
  \centering
  \includegraphics[width=0.7\linewidth]{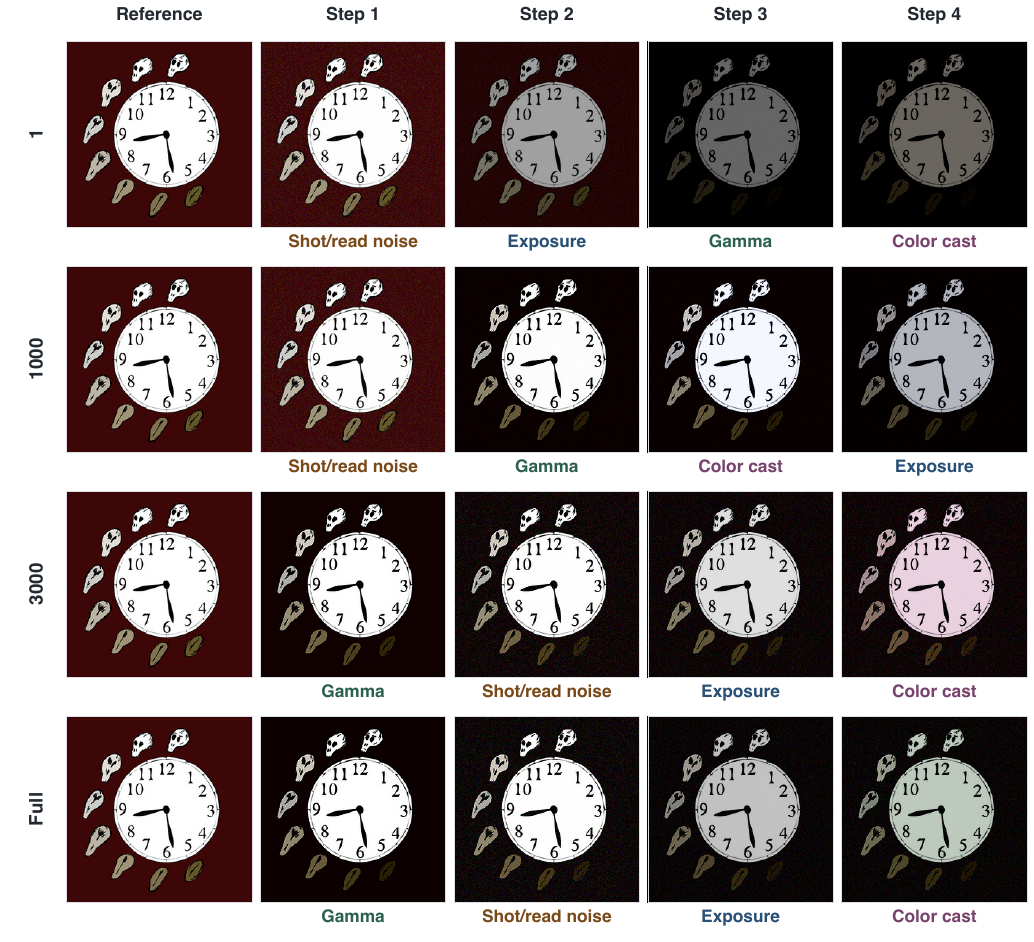}
  \caption{Four-stage degradation of the same normal-light COCO image, with one sampled trajectory per row. Rows 1, 1000, and 3000 use Challengers frozen after the indicated number of outer iterations, while Full uses a later checkpoint from the run with continued Challenger updates.}
  \label{fig:clock_trajectories}
\end{figure*}

To illustrate these shifts on an individual input, we fix the reference image and the trajectory sampling seed across the four configurations in \cref{fig:clock_trajectories}. The intermediate frames show how the image changes at each degradation stage.

Gamma moves from the third position in the first row to the first position in the third row, illustrating the shift toward earlier gamma application. The third and fourth rows use the same order. At the final stage, color cast gives the clock face a pink tint in row 3000 and a green tint in Full.

The shift toward brighter inputs and lower shot amplitudes may preserve visual information useful to the current Solver. Since the Challenger receives a reward for improvement on each synthetic pair, these trends are consistent with a curriculum guided by the training value of generated examples. The comparisons with random and frozen Challengers in \cref{sec:ablations} assess how continued adaptation contributes to real-image enhancement.

\FloatBarrier
\subsection{Persistent Solver states}
Let $n\in\{1,\ldots,5000\}$ index outer iterations and $i\in\{1,2,3,4\}$ index the sequential image pairs within an iteration. Write $w_{n,i}^{(k)}=(\phi_{n,i}^{(k)},\omega_{n,i}^{(k)})$ for the full Solver state after $k\in\{0,1,2\}$ updates on pair $i$, where $\omega$ contains Adam's persistent moment estimates and update counter. The state $w_{1,1}^{(0)}$ is the initialized Solver and Adam state. The transition on the fixed pair $(y_{n,i},x_{n,i})$ is
\begin{equation}
 w_{n,i}^{(k+1)}=\mathcal U_S(w_{n,i}^{(k)};y_{n,i},x_{n,i}),
 \qquad k=0,1.
 \label{eq:appsolverupdate}
\end{equation}
The final state of each pair initializes the next pair, including when training passes from one outer iteration to the next:
\begin{equation}
 w_{n,i+1}^{(0)}=w_{n,i}^{(2)},\quad i=1,2,3,\qquad
 w_{n+1,1}^{(0)}=w_{n,4}^{(2)},\quad n=1,\ldots,4999.
 \label{eq:appstatecarry}
\end{equation}
Thus the four pre-update measurements in one outer iteration use successive Solver states under a common Challenger behavior policy. PPO updates that policy after all four pairs have been processed.

Each generated pair is reused for its two supervised updates and both progress measurements. If progress evaluation introduces additional rollout input noise, the same seed is used for the pre-update and post-update measurements. \Cref{tab:applog} counts the resulting operator applications, optimizer steps, and metric evaluations.

\begin{table}[t]
\caption{Operation ledger for one outer iteration. A trajectory contains four degradation stages and yields one terminal reward. Counts describe the default single-rollout configuration.}
\label{tab:applog}
\centering\footnotesize
\setlength{\tabcolsep}{2.4pt}
\begin{tabular*}{1\linewidth}{@{\extracolsep{\fill}}>{\raggedright\arraybackslash}m{.78\linewidth}c@{}}
\toprule
Operation & Count\\
\midrule
Normal-light references processed & 4\\
Degradation trajectories & 4\\
Physical operator applications & 16\\
Pre-update restoration evaluations & 4\\
Supervised Solver optimizer steps & 8\\
Post-update restoration evaluations & 4\\
Individual learning-progress rewards & 4\\
PPO mini-batches / optimizer steps & 2 / 2\\
\bottomrule
\end{tabular*}
\end{table}

The eight progress evaluations in \cref{tab:applog} add eight forward passes to the eight supervised Solver updates. Each supervised update also requires backpropagation.

\subsection{Cached trajectories and joint PPO update}
\label{app:ppocache}
For every image, the PPO buffer stores the reference and terminal reward together with the initial value estimate. It also retains the total behavior log likelihood and the four stage states. Each stage records its usage mask and normalized index, as well as the selected module and raw continuous action. The updated policy can therefore evaluate the cached action sequence directly, using the same observations and measured reward.

For trajectory $\tau_i$, write $o_{i,t}$ for the cached observation at stage $t$, $m_{i,t}$ for its selected module, and $a_{i,t}$ for its raw continuous action. The policy log likelihood of this cached action sequence is
\begin{equation}
 \ell_\theta(\tau_i)=\sum_{t=0}^{T_d-1}\bigl[\log\pi^{\rm mod}_\theta(m_{i,t}\mid o_{i,t})
 +\log\pi^{\rm par}_\theta(a_{i,t}\mid o_{i,t},m_{i,t})\bigr].
 \label{eq:logprob}
\end{equation}
Only active continuous dimensions contribute to each Gaussian likelihood. \Cref{eq:logprob} sums the module-selection and parameter log probabilities over all stages to form the trajectory-level likelihood used in the PPO ratio.

For the raw advantages $A_i$ defined in \cref{sec:ppo}, let $\bar A$ be the batch mean, $v_A$ the batch variance with divisor $M=4$, and $\epsilon_A=10^{-8}$. The normalized advantages are
\begin{equation}
 \bar A=\frac14\sum_{i=1}^{4}A_i,\qquad
 v_A=\frac14\sum_{i=1}^{4}(A_i-\bar A)^2,\qquad
 \widehat A_i=\frac{A_i-\bar A}{\sqrt{v_A}+\epsilon_A}.
 \label{eq:appadvantagenorm}
\end{equation}
Advantage normalization centers and rescales policy feedback across the four trajectories, while value regression uses the signed raw reward $R_i$ as its target. If all raw advantages are equal, their normalized values are zero, although value regression and entropy can still contribute to the joint update. The denominator in \cref{eq:reward} instead scales each pair's PSNR gain using its stabilized pre-update score.

In \cref{eq:combined}, the entropy bonus $\mathcal H$ is formed from the categorical module entropy and the Gaussian entropy of the sampled module's active coordinates, averaged over trajectories and stages. We use clipping threshold $c=0.2$, value weight $\lambda_V=0.5$, and entropy weight $\lambda_H=0.01$, as listed in \cref{tab:appoptim}.

Each outer iteration takes two PPO optimizer steps with the objective in \cref{eq:combined}, using all four trajectories at each step. A single terminal advantage weights each complete trajectory as the steps update the shared encoder and the policy and value heads. The cached observations and behavior log likelihoods are treated as constants. The rewards, old value estimates, and normalized advantages are also detached during optimization. Degradation rendering and Solver optimization remain outside the Challenger's computation graph.

The four rewards are measured at successive Solver states, so the degradation decision problem evolves with the Solver. The policy observation $o_t$ contains the image, operator history, and stage index, while the Solver weights and optimizer state enter the reward-generating process.

\section{Physical degradation model}
\label{app:physical}
The physical operators impose bounded, interpretable changes on linear RGB. We derive the conditional moments of the shared-noise approximation and describe interactions among these operators within the Challenger's degradation pipeline.

\subsection{Parameterization and shared noise}
\Cref{tab:appranges} gives the bounds of the six active controls after action mapping. The green color gain is fixed to one before normalization by the mean of all three gains.

\begin{table}[t]
\caption{Active physical controls. Bounds refer to parameters after the action mapping in \cref{eq:mapping}. Color gains are normalized afterward.}
\label{tab:appranges}
\centering\footnotesize
\setlength{\tabcolsep}{2.4pt}
\begin{tabular*}{1\linewidth}{@{\extracolsep{\fill}}llc@{}}
\toprule
Operator & Control & Interval\\
\midrule
Exposure & Scale $e$ & $[0.03,0.85]$\\
\multirow[c]{2}{*}{Shot/read noise} & Shot amplitude $s$ & $[0,0.04]$\\
& Read amplitude $r$ & $[0,0.02]$\\
Gamma & Exponent $g$ & $[1.2,3.0]$\\
\multirow[c]{2}{*}{Color cast} & Red gain $q_R$ & $[0.6,1.5]$\\
& Blue gain $q_B$ & $[0.6,1.5]$\\
\bottomrule
\end{tabular*}
\end{table}

The intervals in \cref{tab:appranges} specify coordinatewise limits on the mapped physical controls. After color gains are normalized to have mean one, their effect on mean image brightness still depends on channel intensities and clipping.

To analyze the noise operator before clipping, fix amplitudes $s,r\geq0$. Given a nonnegative linear-RGB intensity $z$, draw $n\mid z\sim\mathcal N(0,1)$ and define
\begin{equation}
 \widetilde z=z+(s\sqrt z+r)n.
 \label{eq:appnoiseunclipped}
\end{equation}
The perturbation amplitude is fixed after conditioning on $z$, while $\mathbb E[n\mid z]=0$ and $\operatorname{Var}(n\mid z)=1$. Using linearity of expectation and the variance rule for a scaled random variable gives
\begin{align}
 \mathbb E[\widetilde z\mid z]
 &=z+(s\sqrt z+r)\mathbb E[n\mid z]=z,\label{eq:appnoisemean}\\
 \operatorname{Var}(\widetilde z\mid z)
 &=(s\sqrt z+r)^2\operatorname{Var}(n\mid z)
 =(s\sqrt z+r)^2=s^2z+2sr\sqrt z+r^2.\label{eq:appnoisevariance}
\end{align}
The cross term arises from the shared random draw for the shot and read contributions. With independent standard Gaussian draws instead, their conditional covariance would be zero, and their variances would add to $s^2z+r^2$. Clipping can alter both conditional mean and variance, so \cref{eq:appnoisemean,eq:appnoisevariance} apply to the unclipped signal-dependent Gaussian approximation.

\subsection{Composition of physical operators}
For an intensity where numerical floors and clipping are inactive, let exposure be $E_e(z)=ez$ and the gamma transformation be $G_g(z)=z^g$. Their two possible compositions satisfy
\begin{equation}
 G_g\bigl(E_e(z)\bigr)=(ez)^g=e^g z^g,\qquad
 E_e\bigl(G_g(z)\bigr)=e(z^g)=e z^g.
 \label{eq:appexposuregamma}
\end{equation}
Since $0<e<1$ and $g>1$, we have $e^{g-1}<1$ and hence $e^g<e$. For positive $z$ in this regime, exposure followed by gamma gives a darker value than the reverse order at identical parameters.

Exposure also changes the signal-dependent noise amplitude. Define the unclipped scalar noise operator $N(z)=z+(s\sqrt z+r)n$ as in \cref{eq:appnoiseunclipped}, drawing a standard normal after conditioning on its input. The compositions are $N(E_e(z))=ez+(s\sqrt{ez}+r)n$ and $E_e(N(z))=ez+e(s\sqrt z+r)n$. Applying the same conditional-variance rule as above yields
\begin{equation}
 \operatorname{Var}\Bigl(N\bigl(E_e(z)\bigr)\mid z\Bigr)=(s\sqrt{ez}+r)^2,\qquad
 \operatorname{Var}\Bigl(E_e\bigl(N(z)\bigr)\mid z\Bigr)=e^2(s\sqrt z+r)^2.
 \label{eq:appexposurenoise}
\end{equation}
The first expression adds noise after attenuation, whereas the second attenuates noise already added. These conditional variances generally differ at identical exposure, amplitudes, and reference intensity, so action order can change synthetic observations. Exposure and fixed channel gains commute when clipping is inactive because $e(\tilde q_c z)=\tilde q_c(ez)$ for each channel $c\in\{R,G,B\}$.

\section{Evaluation protocol and run records}
\label{app:evaluation}
The evaluation uses the benchmark splits and training/test separation in \cref{sec:datasets}. We first describe input preprocessing and the paired metrics, then specify how results are aggregated over unpaired images and independent training runs.

\subsection{Paired metrics and preprocessing}
\label{app:evalpreprocess}
Following MR.~Illuminate's evaluation protocol, we compute PSNR, SSIM, and LPIPS for each restored image and its paired reference before averaging within each dataset. Within a run, the LOL score combines the 15 LOL-v1 and 100 LOL-v2-real test images, giving their split means weights $15/115$ and $100/115$, respectively. Summaries across runs follow \cref{app:runsummary}.

We compute LPIPS with the same AlexNet-based configuration as MR.~Illuminate.

Before enhancement, we use MR.~Illuminate's mean-intensity scaling. The scale factor $\alpha(I)$ and the scaled input $I_{\rm in}$ are
\begin{equation}
 \begin{aligned}
 \alpha(I)&=\begin{cases}
 \tau_{\rm in}/\operatorname{mean}(I), & 0<\operatorname{mean}(I)<\tau_{\rm in},\\
 1, & \text{otherwise},
 \end{cases}\\
 I_{\rm in}&=\alpha(I)I,
 \end{aligned}
 \label{eq:appintensity}
\end{equation}
where $I\in[0,1]^{3\times H\times W}$ is the normalized sRGB input, $\operatorname{mean}(I)=(3HW)^{-1}\sum_{c,h,w}I_{c,h,w}$, and $\tau_{\rm in}=30/255$. Scaling can raise individual pixel values above one. MR.~Illuminate's public implementation additionally resizes inputs to dimensions divisible by eight and clips and quantizes scaled intensities to 8 bits. The reference image retains its original intensities, and evaluation uses zero additional input noise. Center cropping and random horizontal flipping are used during training, as specified in \cref{sec:datasets}.

We do not use ground-truth mean (GT-Mean) alignment in any of our evaluations.

\subsection{Unpaired collection}
\Cref{tab:appup} lists the constituent dataset counts for Unpaired, defined in \cref{sec:datasets}, following MR.~Illuminate's evaluation protocol. Collection-level metrics weight each image equally.

\begin{table}[t]
\caption{Image counts in the unpaired evaluation collection.}
\label{tab:appup}
\centering\footnotesize
\setlength{\tabcolsep}{2.4pt}
\begin{tabular*}{1\linewidth}{@{\extracolsep{\fill}}l*{5}{c}@{}}
\toprule
Dataset & DICM & LIME & MEF & NPE & VV\\
\midrule
Images & 44 & 10 & 79 & 75 & 24\\
\bottomrule
\end{tabular*}
\end{table}

For the 232 images in \cref{tab:appup}, let $\mathcal D$ be the set of five datasets, $n_d>0$ the size of dataset $d$, and $v_{d,j}$ the no-reference score for its $j$th image. The image-weighted and dataset-weighted summaries are
\begin{align}
 \bar v_{\rm image}&=\frac{\sum_{d\in\mathcal D}\sum_{j=1}^{n_d}v_{d,j}}{\sum_{d\in\mathcal D}n_d},
 \label{eq:appimageavg}\\
 \bar v_{\rm dataset}&=\frac15\sum_{d\in\mathcal D}\frac1{n_d}\sum_{j=1}^{n_d}v_{d,j}.
 \label{eq:appdatasetavg}
\end{align}
Dark-Zero uses the image-weighted summary in \cref{eq:appimageavg}, where each dataset contributes in proportion to its size. The alternative in \cref{eq:appdatasetavg} gives equal weight to each dataset.

\subsection{Run summaries and external values}
\label{app:runsummary}
Dark-Zero's quantitative experiments use seeds 0, 21, and 42. Within each run, we compute collection-level scores by weighting dataset means by their test-image counts. We then report the equal-weight mean and sample standard deviation (SD) across the three runs. The sample variance uses divisor $3-1=2$, and all statistics are computed before display rounding. PSNR is reported in dB.

Per-run collection scores appear in \cref{tab:appsolverruns}, with dataset-level records in \cref{tab:appsolverpairedruns,tab:appsolverunpairedruns}. The main table and the full-model reference in \cref{tab:ablations} summarize the same three DarkIR runs listed in \cref{tab:appextended}.

The inherited rows in \cref{tab:main,tab:appextended} use the measurements published by MR.~Illuminate. External rows retain their source training-data and pretrained-model assumptions, including the Retinexformer configuration trained on the MIT-Adobe FiveK (MIT5K) collection~\citep{bychkovsky2011learning}.

\section{Ablations and additional Solvers}
\label{app:controls}
This section details the seven ablation variants in \cref{sec:ablations} and the five-Solver comparison in \cref{sec:multisolver}. It reports individual runs and a PPO update-budget study, with all run summaries following \cref{app:runsummary}.

\subsection{Seven ablation variants}
All ablation variants are evaluated with seeds 0, 21, and 42, and \cref{tab:ablations} summarizes their LOL PSNR, SSIM, and LPIPS. For every variant, evaluation follows the mean-intensity preprocessing in \cref{eq:appintensity}, with the Solver in evaluation mode and zero additional input noise.

Uniform random Challenger randomly orders the four degradation operators for each trajectory and samples their physical parameters uniformly within the allowed ranges. The operators and parameter ranges are specified in \cref{sec:operators}.

L1-only Solver loss retains the mean L1 term in \cref{eq:solverloss} and sets the SSIM and LPIPS coefficients to zero.

Frozen Challenger (1), Frozen Challenger (1,000), and Frozen Challenger (3,000) stop Challenger updates after 1, 1,000, or 3,000 outer iterations, respectively. Solver training continues in the same run using the resulting policy, whose network weights remain fixed for the rest of the run. The policy continues sampling degradation actions after freezing.

Frozen Challenger (pretrained) takes the final Challenger from a completed 5,000-iteration training run with the same seed as the ablation run. A new Solver is then trained for 5,000 outer iterations with that Challenger held fixed from the start. The fixed policy continues sampling degradation actions throughout this separate run, providing a control for training with a fully trained degradation policy.

The Absolute PSNR gain variant uses the change in dB as its reward, $R_i^{\rm abs}=P_{{\rm post},i}-P_{{\rm pre},i}$. The relative reward in \cref{eq:reward} divides this gain by the stabilized pre-update PSNR. Thus two images with the same positive gain can receive different rewards if they start at different PSNRs. These rewards also serve as value targets. PPO normalizes advantages across the batch, so that step does not undo the image-specific scaling. All variants and the full model share the same inference architecture and cost.

\Cref{tab:appablationruns} provides the per-seed LOL scores for each variant under the shared 115-image evaluation protocol. Its full-model rows are the main-experiment runs reported in \cref{tab:appextended}.

\begin{table*}[tbp]
\caption{LOL ablation results with DarkIR for all three seeds, including the main configuration. Mean rows show mean $\pm$ SD, with PSNR in dB.}
\label{tab:appablationruns}
\centering
\footnotesize
\setlength{\tabcolsep}{1.6pt}
\renewcommand{\arraystretch}{1.08}
\begin{tabular*}{1\linewidth}{@{\extracolsep{\fill}}>{\raggedright\arraybackslash}m{.30\linewidth}>{\centering\arraybackslash}m{.10\linewidth}*{3}{>{\centering\arraybackslash}m{.19\linewidth}}@{}}
\toprule
Variant & Seed / mean & PSNR $\uparrow$ & SSIM $\uparrow$ & LPIPS $\downarrow$\\
\midrule
\multirow[c]{4}{=}{Full model} & 0 & 22.8103 & 0.8353 & 0.1576\\
 & 21 & 22.8907 & 0.8308 & 0.1574\\
 & 42 & 22.8165 & 0.8275 & 0.1657\\
 & Mean & \mbox{22.8392\hspace{0.5pt}{\fontsize{6}{7}\selectfont$\pm 0.0448$}} & \mbox{0.8312\hspace{0.5pt}{\fontsize{6}{7}\selectfont$\pm 0.0039$}} & \mbox{0.1602\hspace{0.5pt}{\fontsize{6}{7}\selectfont$\pm 0.0048$}}\\
\midrule
\multirow[c]{4}{=}{Absolute PSNR gain} & 0 & 22.0151 & 0.8207 & 0.1751\\
 & 21 & 22.1632 & 0.8033 & 0.1920\\
 & 42 & 22.1278 & 0.8058 & 0.2001\\
 & Mean & \mbox{22.1020\hspace{0.5pt}{\fontsize{6}{7}\selectfont$\pm 0.0773$}} & \mbox{0.8099\hspace{0.5pt}{\fontsize{6}{7}\selectfont$\pm 0.0094$}} & \mbox{0.1891\hspace{0.5pt}{\fontsize{6}{7}\selectfont$\pm 0.0128$}}\\
\midrule
\multirow[c]{4}{=}{L1-only Solver loss} & 0 & 21.4090 & 0.7251 & 0.2574\\
 & 21 & 21.5759 & 0.7297 & 0.2715\\
 & 42 & 21.2776 & 0.7230 & 0.2732\\
 & Mean & \mbox{21.4208\hspace{0.5pt}{\fontsize{6}{7}\selectfont$\pm 0.1495$}} & \mbox{0.7259\hspace{0.5pt}{\fontsize{6}{7}\selectfont$\pm 0.0034$}} & \mbox{0.2674\hspace{0.5pt}{\fontsize{6}{7}\selectfont$\pm 0.0087$}}\\
\midrule
\multirow[c]{4}{=}{Uniform random Challenger} & 0 & 21.7892 & 0.8243 & 0.1631\\
 & 21 & 21.7433 & 0.8079 & 0.1744\\
 & 42 & 21.7526 & 0.8155 & 0.1803\\
 & Mean & \mbox{21.7617\hspace{0.5pt}{\fontsize{6}{7}\selectfont$\pm 0.0243$}} & \mbox{0.8159\hspace{0.5pt}{\fontsize{6}{7}\selectfont$\pm 0.0082$}} & \mbox{0.1726\hspace{0.5pt}{\fontsize{6}{7}\selectfont$\pm 0.0087$}}\\
\midrule
\multirow[c]{4}{=}{Frozen Challenger (1)} & 0 & 21.6971 & 0.7978 & 0.1985\\
 & 21 & 21.4360 & 0.7216 & 0.2465\\
 & 42 & 21.3093 & 0.7014 & 0.2481\\
 & Mean & \mbox{21.4808\hspace{0.5pt}{\fontsize{6}{7}\selectfont$\pm 0.1977$}} & \mbox{0.7403\hspace{0.5pt}{\fontsize{6}{7}\selectfont$\pm 0.0508$}} & \mbox{0.2310\hspace{0.5pt}{\fontsize{6}{7}\selectfont$\pm 0.0282$}}\\
\midrule
\multirow[c]{4}{=}{Frozen Challenger (1,000)} & 0 & 21.7672 & 0.7980 & 0.1956\\
 & 21 & 21.0020 & 0.7904 & 0.2042\\
 & 42 & 21.0904 & 0.7984 & 0.1964\\
 & Mean & \mbox{21.2865\hspace{0.5pt}{\fontsize{6}{7}\selectfont$\pm 0.4186$}} & \mbox{0.7956\hspace{0.5pt}{\fontsize{6}{7}\selectfont$\pm 0.0045$}} & \mbox{0.1987\hspace{0.5pt}{\fontsize{6}{7}\selectfont$\pm 0.0048$}}\\
\midrule
\multirow[c]{4}{=}{Frozen Challenger (3,000)} & 0 & 22.3540 & 0.8284 & 0.1592\\
 & 21 & 22.2804 & 0.8265 & 0.1647\\
 & 42 & 21.9458 & 0.8248 & 0.1743\\
 & Mean & \mbox{22.1934\hspace{0.5pt}{\fontsize{6}{7}\selectfont$\pm 0.2176$}} & \mbox{0.8266\hspace{0.5pt}{\fontsize{6}{7}\selectfont$\pm 0.0018$}} & \mbox{0.1661\hspace{0.5pt}{\fontsize{6}{7}\selectfont$\pm 0.0076$}}\\
\midrule
\multirow[c]{4}{=}{Frozen Challenger (pretrained)} & 0 & 22.0748 & 0.8327 & 0.1630\\
 & 21 & 21.9925 & 0.8173 & 0.1789\\
 & 42 & 22.0711 & 0.8281 & 0.1392\\
 & Mean & \mbox{22.0461\hspace{0.5pt}{\fontsize{6}{7}\selectfont$\pm 0.0465$}} & \mbox{0.8260\hspace{0.5pt}{\fontsize{6}{7}\selectfont$\pm 0.0079$}} & \mbox{0.1604\hspace{0.5pt}{\fontsize{6}{7}\selectfont$\pm 0.0200$}}\\
\bottomrule
\end{tabular*}
\end{table*}

\begin{figure*}[!tp]
  \centering
  \includegraphics[width=\linewidth]{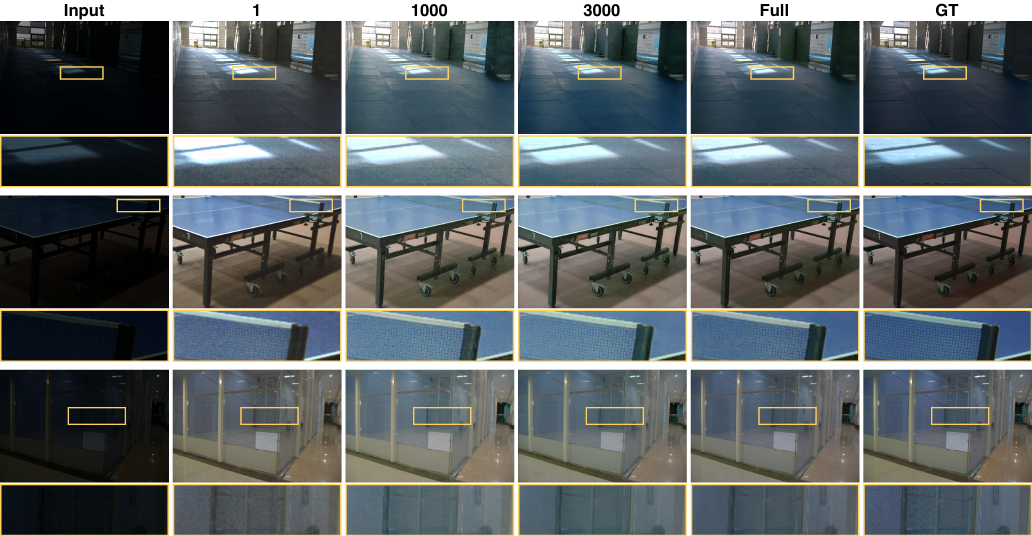}
  \caption{Qualitative comparison on LOL with Challengers frozen after 1, 1,000, or 3,000 outer iterations and with continued updates in Full. Full preserves the net mesh with less visible speckle than the earliest frozen variant. The yellow boxes mark crops of a floor reflection, the net, and a glass partition.}
  \label{fig:freeze_visuals}
\end{figure*}

On three LOL inputs, \cref{fig:freeze_visuals} compares continued Challenger updates with freezing at different stages of training.

The edge of the net post is also clearer in Full. In the glass crop, Full reduces the cyan cast visible in the variants frozen after 1,000 and 3,000 outer iterations, bringing the color closer to the gray-blue reference.

These observations complement the three-seed results in \cref{tab:appablationruns}. The variant frozen at 3,000 outer iterations has the highest mean LOL PSNR among the frozen variants, and Full exceeds that mean by 0.65\,dB.

\begin{figure*}[!tp]
  \centering
  \includegraphics[width=\linewidth]{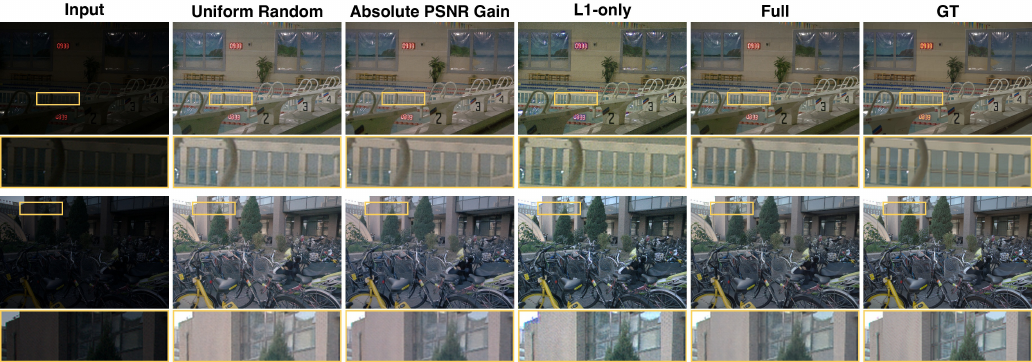}
  \caption{Qualitative ablation on LOL with a uniform random Challenger, an absolute PSNR-gain reward, and an L1-only Solver loss. Full renders the pool railing and surrounding surfaces closer to the reference colors, with less visible speckle than the L1-only variant. In the bicycle scene, the wall in Full is closer to the reference tone than the warmer wall in the absolute-gain variant.}
  \label{fig:component_ablation}
\end{figure*}

\Cref{fig:component_ablation} extends the visual comparison to the uniform random Challenger, absolute PSNR gain, and L1-only loss.

The L1-only variant also shows visible speckle in the bicycle crop.

\subsection{Dataset-level results and Solver comparison}
\label{app:datasetrecords}
\Cref{tab:appdatasetrecords} compares dataset-level means over the three seeds for all five Solvers with our evaluation of MR.~Illuminate's released checkpoint. The table separates the two LOL splits and the two LSRW subsets, then reports each constituent dataset of Unpaired. The checkpoint results are aggregated separately from the published values quoted in \cref{tab:appextended}.

\begin{table*}[!t]
\caption{Dataset-level comparison of our evaluation of MR.~Illuminate's released checkpoint with Dark-Zero means over three seeds. Dataset sizes are in parentheses, PSNR is in dB, and quality values are displayed to four decimal places.}
\label{tab:appdatasetrecords}
\centering
\footnotesize
\setlength{\tabcolsep}{1pt}
\renewcommand{\arraystretch}{1.08}
\begin{tabular*}{1\linewidth}{@{\extracolsep{\fill}}>{\raggedright\arraybackslash}m{.15\linewidth}>{\raggedright\arraybackslash}m{.095\linewidth}*{6}{>{\centering\arraybackslash}m{.12\linewidth}}@{}}
\toprule
\shortstack[l]{Dataset\\(images)} & Metric & \shortstack{MR.\\Illuminate} & DarkIR & CIDNet & Restormer & LLFormer & SNR-Aware\\
\midrule
\multicolumn{8}{l}{Paired datasets}\\
\midrule
\multirow[c]{3}{*}{LOL-v1 (15)} & PSNR $\uparrow$ & 17.5821 & 18.1222 & 18.9473 & 18.7447 & 19.0047 & 17.9017\\
 & SSIM $\uparrow$ & 0.7728 & 0.7814 & 0.7667 & 0.7883 & 0.7875 & 0.7611\\
 & LPIPS $\downarrow$ & 0.2067 & 0.1883 & 0.2284 & 0.1911 & 0.2003 & 0.2356\\
\midrule
\multirow[c]{3}{*}{LOL-v2-real (100)} & PSNR $\uparrow$ & 22.3645 & 23.5467 & 23.3293 & 23.2423 & 22.8690 & 22.9623\\
 & SSIM $\uparrow$ & 0.8247 & 0.8386 & 0.8104 & 0.8356 & 0.8318 & 0.8140\\
 & LPIPS $\downarrow$ & 0.1727 & 0.1560 & 0.1871 & 0.1512 & 0.1742 & 0.2046\\
\midrule
\multirow[c]{3}{*}{LSRW-Huawei (30)} & PSNR $\uparrow$ & 19.1090 & 19.5211 & 19.4113 & 18.7477 & 19.0727 & 19.3970\\
 & SSIM $\uparrow$ & 0.5602 & 0.5949 & 0.5788 & 0.5359 & 0.5899 & 0.5873\\
 & LPIPS $\downarrow$ & 0.3240 & 0.3446 & 0.3430 & 0.3623 & 0.3464 & 0.3868\\
\midrule
\multirow[c]{3}{*}{LSRW-Nikon (20)} & PSNR $\uparrow$ & 15.4819 & 15.6097 & 15.7753 & 14.8167 & 14.4707 & 16.5013\\
 & SSIM $\uparrow$ & 0.4675 & 0.4862 & 0.4832 & 0.4597 & 0.4711 & 0.4933\\
 & LPIPS $\downarrow$ & 0.2228 & 0.2514 & 0.2442 & 0.2473 & 0.2802 & 0.2928\\
\midrule
\multicolumn{8}{l}{Unpaired datasets}\\
\midrule
\multirow[c]{3}{*}{DICM (44)} & ILNIQE $\downarrow$ & 22.2748 & 21.5526 & 22.0519 & 22.2742 & 21.8530 & 22.2563\\
 & BRISQUE $\downarrow$ & 12.7317 & 13.9270 & 15.9380 & 11.7735 & 14.7059 & 20.1520\\
 & NL $\downarrow$ & 0.4462 & 0.5444 & 0.5419 & 0.6133 & 0.4835 & 0.5648\\
\midrule
\multirow[c]{3}{*}{LIME (10)} & ILNIQE $\downarrow$ & 27.0402 & 24.8427 & 27.2843 & 26.9388 & 25.8735 & 25.9680\\
 & BRISQUE $\downarrow$ & 17.0494 & 11.6394 & 20.0287 & 14.2607 & 16.5731 & 17.6870\\
 & NL $\downarrow$ & 0.3397 & 0.9363 & 0.6283 & 0.9076 & 0.7585 & 1.2190\\
\midrule
\multirow[c]{3}{*}{MEF (79)} & ILNIQE $\downarrow$ & 27.8251 & 25.0115 & 26.3573 & 24.9722 & 25.8483 & 26.0684\\
 & BRISQUE $\downarrow$ & 16.1510 & 16.0469 & 19.9044 & 15.1758 & 18.0847 & 19.9399\\
 & NL $\downarrow$ & 0.3834 & 0.9310 & 0.6598 & 0.9322 & 0.7423 & 1.2857\\
\midrule
\multirow[c]{3}{*}{NPE (75)} & ILNIQE $\downarrow$ & 23.8952 & 23.3234 & 23.6755 & 23.8563 & 23.8760 & 24.1696\\
 & BRISQUE $\downarrow$ & 18.7667 & 17.2648 & 21.6887 & 19.9814 & 18.3582 & 23.9126\\
 & NL $\downarrow$ & 0.3892 & 0.5529 & 0.4766 & 0.5630 & 0.4851 & 0.6096\\
\midrule
\multirow[c]{3}{*}{VV (24)} & ILNIQE $\downarrow$ & 23.0912 & 22.6363 & 22.3457 & 22.7696 & 22.2282 & 22.5830\\
 & BRISQUE $\downarrow$ & 15.7687 & 15.2314 & 24.2341 & 21.0043 & 17.9873 & 19.4823\\
 & NL $\downarrow$ & 0.3640 & 0.4161 & 0.4188 & 0.4507 & 0.3778 & 0.4431\\
\bottomrule
\end{tabular*}
\end{table*}


\subsection{Results for individual seeds and standard deviations}
\label{app:solverrunstats}
\Cref{tab:appsolverruns} reports the collection-level scores used to compute the Solver means in \cref{tab:solvers}. The corresponding dataset-level records in \cref{tab:appsolverpairedruns,tab:appsolverunpairedruns} underlie \cref{tab:appdatasetrecords} and show all three seeds together for every Solver.

\begin{table*}[!t]
\caption{Collection-level results for all five Solvers across three seeds, followed by their means and SDs. PSNR is in dB.}
\label{tab:appsolverruns}
\centering
\scriptsize
\setlength{\tabcolsep}{0.5pt}
\renewcommand{\arraystretch}{1.10}
\begin{tabular*}{1\linewidth}{@{\extracolsep{\fill}}>{\raggedright\arraybackslash}m{.10\linewidth}>{\centering\arraybackslash}m{.055\linewidth}*{9}{>{\centering\arraybackslash}m{.09\linewidth}}@{}}
\toprule
\tworowhead{Solver} & \tworowhead{Run} & \multicolumn{3}{c}{LOL} & \multicolumn{3}{c}{LSRW} & \multicolumn{3}{c}{Unpaired}\\
\cmidrule(lr){3-5}\cmidrule(lr){6-8}\cmidrule(lr){9-11}
 &  & PSNR $\uparrow$ & SSIM $\uparrow$ & LPIPS $\downarrow$ & PSNR $\uparrow$ & SSIM $\uparrow$ & LPIPS $\downarrow$ & ILNIQE $\downarrow$ & BRISQUE $\downarrow$ & NL $\downarrow$\\
\midrule
\multirow[c]{5}{=}{DarkIR} & 0 & 22.8103 & 0.8353 & 0.1576 & 17.8017 & 0.5524 & 0.3095 & 23.4979 & 15.6842 & 0.6651\\
 & 21 & 22.8907 & 0.8308 & 0.1574 & 17.8623 & 0.5479 & 0.3062 & 23.3726 & 16.1637 & 0.6717\\
 & 42 & 22.8165 & 0.8275 & 0.1657 & 18.2057 & 0.5541 & 0.3063 & 23.7998 & 15.4448 & 0.7104\\
 & Mean & 22.8392 & 0.8312 & 0.1602 & 17.9566 & 0.5515 & 0.3073 & 23.5568 & 15.7642 & 0.6824\\
 & SD & 0.0448 & 0.0039 & 0.0048 & 0.2179 & 0.0032 & 0.0018 & 0.2196 & 0.3661 & 0.0245\\
\midrule
\multirow[c]{5}{=}{CIDNet} & 0 & 22.6391 & 0.8011 & 0.2005 & 18.1336 & 0.5502 & 0.3120 & 24.0841 & 20.3016 & 0.5662\\
 & 21 & 22.9367 & 0.8135 & 0.1757 & 18.1096 & 0.5459 & 0.3061 & 24.4922 & 20.9717 & 0.5434\\
 & 42 & 22.6975 & 0.7994 & 0.2012 & 17.6276 & 0.5255 & 0.2923 & 24.3200 & 19.2733 & 0.5461\\
 & Mean & 22.7578 & 0.8047 & 0.1925 & 17.9569 & 0.5405 & 0.3035 & 24.2988 & 20.1822 & 0.5519\\
 & SD & 0.1577 & 0.0077 & 0.0145 & 0.2855 & 0.0132 & 0.0101 & 0.2049 & 0.8555 & 0.0125\\
\midrule
\multirow[c]{5}{=}{SNR-Aware} & 0 & 22.4345 & 0.8071 & 0.2028 & 18.1044 & 0.5464 & 0.3514 & 24.4243 & 20.5678 & 0.8175\\
 & 21 & 22.2816 & 0.8025 & 0.2153 & 18.0948 & 0.5460 & 0.3384 & 23.6121 & 17.7404 & 0.8677\\
 & 42 & 22.1907 & 0.8117 & 0.2080 & 18.5170 & 0.5567 & 0.3577 & 25.0636 & 25.0517 & 0.8360\\
 & Mean & 22.3022 & 0.8071 & 0.2087 & 18.2387 & 0.5497 & 0.3492 & 24.3667 & 21.1199 & 0.8404\\
 & SD & 0.1232 & 0.0046 & 0.0063 & 0.2410 & 0.0061 & 0.0099 & 0.7275 & 3.6868 & 0.0254\\
\midrule
\multirow[c]{5}{=}{LLFormer} & 0 & 22.3964 & 0.8191 & 0.1836 & 17.0852 & 0.5339 & 0.3020 & 23.9012 & 14.6013 & 0.5853\\
 & 21 & 22.2500 & 0.8226 & 0.1865 & 17.5856 & 0.5514 & 0.3253 & 23.7688 & 18.5712 & 0.6517\\
 & 42 & 22.4485 & 0.8364 & 0.1627 & 17.0248 & 0.5419 & 0.3325 & 24.5687 & 19.1987 & 0.4822\\
 & Mean & 22.3650 & 0.8260 & 0.1776 & 17.2319 & 0.5424 & 0.3199 & 24.0796 & 17.4571 & 0.5731\\
 & SD & 0.1029 & 0.0091 & 0.0130 & 0.3078 & 0.0088 & 0.0159 & 0.4287 & 2.4930 & 0.0854\\
\midrule
\multirow[c]{5}{=}{Restormer} & 0 & 22.6389 & 0.8355 & 0.1527 & 17.3608 & 0.5120 & 0.3148 & 23.9739 & 16.8712 & 0.7221\\
 & 21 & 22.5584 & 0.8182 & 0.1669 & 17.1076 & 0.5003 & 0.3124 & 24.0672 & 16.4023 & 0.6687\\
 & 42 & 22.7697 & 0.8346 & 0.1496 & 17.0574 & 0.5039 & 0.3216 & 23.8289 & 16.6693 & 0.7138\\
 & Mean & 22.6557 & 0.8294 & 0.1564 & 17.1753 & 0.5054 & 0.3163 & 23.9567 & 16.6476 & 0.7015\\
 & SD & 0.1067 & 0.0097 & 0.0092 & 0.1626 & 0.0060 & 0.0048 & 0.1201 & 0.2352 & 0.0287\\
\bottomrule
\end{tabular*}
\end{table*}

\begin{table*}[!t]
\caption{Paired-dataset results for each Solver across three seeds, followed by their means and SDs. PSNR is in dB, and run scores and means are displayed to four decimal places.}
\label{tab:appsolverpairedruns}
\centering
\scriptsize
\setlength{\tabcolsep}{0.5pt}
\renewcommand{\arraystretch}{1.10}
\begin{tabular*}{1\linewidth}{@{\extracolsep{\fill}}>{\raggedright\arraybackslash}m{.10\linewidth}>{\centering\arraybackslash}m{.055\linewidth}*{12}{>{\centering\arraybackslash}m{.066\linewidth}}@{}}
\toprule
\tworowhead{Solver} & \tworowhead{Run} & \multicolumn{3}{c}{LOL-v1} & \multicolumn{3}{c}{LOL-v2-real} & \multicolumn{3}{c}{LSRW-Huawei} & \multicolumn{3}{c}{LSRW-Nikon}\\
\cmidrule(lr){3-5}\cmidrule(lr){6-8}\cmidrule(lr){9-11}\cmidrule(lr){12-14}
 &  & PSNR $\uparrow$ & SSIM $\uparrow$ & LPIPS $\downarrow$ & PSNR $\uparrow$ & SSIM $\uparrow$ & LPIPS $\downarrow$ & PSNR $\uparrow$ & SSIM $\uparrow$ & LPIPS $\downarrow$ & PSNR $\uparrow$ & SSIM $\uparrow$ & LPIPS $\downarrow$\\
\midrule
\multirow[c]{5}{=}{DarkIR} & 0 & 17.9877 & 0.7798 & 0.1829 & 23.5337 & 0.8436 & 0.1538 & 19.4415 & 0.5970 & 0.3485 & 15.3420 & 0.4854 & 0.2509\\
 & 21 & 18.1003 & 0.7827 & 0.1880 & 23.6093 & 0.8380 & 0.1528 & 19.4527 & 0.5919 & 0.3452 & 15.4766 & 0.4820 & 0.2477\\
 & 42 & 18.2785 & 0.7818 & 0.1940 & 23.4972 & 0.8343 & 0.1615 & 19.6691 & 0.5959 & 0.3402 & 16.0106 & 0.4913 & 0.2555\\
 & Mean & 18.1222 & 0.7814 & 0.1883 & 23.5467 & 0.8386 & 0.1560 & 19.5211 & 0.5949 & 0.3446 & 15.6097 & 0.4862 & 0.2514\\
 & SD & 0.1466 & 0.0015 & 0.0056 & 0.0572 & 0.0047 & 0.0048 & 0.1283 & 0.0027 & 0.0042 & 0.3536 & 0.0047 & 0.0039\\
\midrule
\multirow[c]{5}{=}{CIDNet} & 0 & 18.1800 & 0.7564 & 0.2422 & 23.3080 & 0.8078 & 0.1942 & 19.5740 & 0.5907 & 0.3512 & 15.9730 & 0.4895 & 0.2532\\
 & 21 & 19.3880 & 0.7787 & 0.2072 & 23.4690 & 0.8187 & 0.1710 & 19.4320 & 0.5846 & 0.3506 & 16.1260 & 0.4879 & 0.2394\\
 & 42 & 19.2740 & 0.7651 & 0.2358 & 23.2110 & 0.8046 & 0.1960 & 19.2280 & 0.5611 & 0.3272 & 15.2270 & 0.4721 & 0.2400\\
 & Mean & 18.9473 & 0.7667 & 0.2284 & 23.3293 & 0.8104 & 0.1871 & 19.4113 & 0.5788 & 0.3430 & 15.7753 & 0.4832 & 0.2442\\
 & SD & 0.6670 & 0.0112 & 0.0186 & 0.1303 & 0.0074 & 0.0139 & 0.1739 & 0.0156 & 0.0137 & 0.4810 & 0.0096 & 0.0078\\
\midrule
\multirow[c]{5}{=}{SNR-Aware} & 0 & 18.0110 & 0.7597 & 0.2299 & 23.0980 & 0.8142 & 0.1987 & 19.2860 & 0.5833 & 0.3848 & 16.3320 & 0.4910 & 0.3012\\
 & 21 & 17.9390 & 0.7588 & 0.2451 & 22.9330 & 0.8091 & 0.2108 & 19.1940 & 0.5833 & 0.3792 & 16.4460 & 0.4901 & 0.2772\\
 & 42 & 17.7550 & 0.7647 & 0.2317 & 22.8560 & 0.8188 & 0.2044 & 19.7110 & 0.5952 & 0.3963 & 16.7260 & 0.4989 & 0.2999\\
 & Mean & 17.9017 & 0.7611 & 0.2356 & 22.9623 & 0.8140 & 0.2046 & 19.3970 & 0.5873 & 0.3868 & 16.5013 & 0.4933 & 0.2928\\
 & SD & 0.1320 & 0.0032 & 0.0083 & 0.1236 & 0.0049 & 0.0061 & 0.2758 & 0.0069 & 0.0087 & 0.2027 & 0.0048 & 0.0135\\
\midrule
\multirow[c]{5}{=}{LLFormer} & 0 & 18.3260 & 0.7737 & 0.2155 & 23.0070 & 0.8259 & 0.1788 & 18.9740 & 0.5819 & 0.3279 & 14.2520 & 0.4618 & 0.2632\\
 & 21 & 18.2630 & 0.7772 & 0.2203 & 22.8480 & 0.8294 & 0.1814 & 19.3140 & 0.5955 & 0.3556 & 14.9930 & 0.4853 & 0.2798\\
 & 42 & 20.4250 & 0.8117 & 0.1651 & 22.7520 & 0.8401 & 0.1623 & 18.9300 & 0.5924 & 0.3557 & 14.1670 & 0.4661 & 0.2976\\
 & Mean & 19.0047 & 0.7875 & 0.2003 & 22.8690 & 0.8318 & 0.1742 & 19.0727 & 0.5899 & 0.3464 & 14.4707 & 0.4711 & 0.2802\\
 & SD & 1.2304 & 0.0210 & 0.0306 & 0.1288 & 0.0074 & 0.0104 & 0.2102 & 0.0071 & 0.0160 & 0.4543 & 0.0125 & 0.0172\\
\midrule
\multirow[c]{5}{=}{Restormer} & 0 & 19.0780 & 0.7960 & 0.1857 & 23.1730 & 0.8414 & 0.1478 & 18.8420 & 0.5421 & 0.3574 & 15.1390 & 0.4669 & 0.2509\\
 & 21 & 18.7680 & 0.7821 & 0.1964 & 23.1270 & 0.8236 & 0.1625 & 18.8520 & 0.5322 & 0.3556 & 14.4910 & 0.4524 & 0.2477\\
 & 42 & 18.3880 & 0.7868 & 0.1913 & 23.4270 & 0.8418 & 0.1433 & 18.5490 & 0.5333 & 0.3738 & 14.8200 & 0.4599 & 0.2434\\
 & Mean & 18.7447 & 0.7883 & 0.1911 & 23.2423 & 0.8356 & 0.1512 & 18.7477 & 0.5359 & 0.3623 & 14.8167 & 0.4597 & 0.2473\\
 & SD & 0.3456 & 0.0071 & 0.0054 & 0.1616 & 0.0104 & 0.0100 & 0.1721 & 0.0054 & 0.0100 & 0.3240 & 0.0073 & 0.0038\\
\bottomrule
\end{tabular*}
\end{table*}

\begin{table*}[!t]
\caption{Unpaired subset results for each Solver across three seeds, followed by their means and SDs. Run scores and means are displayed to four decimal places.}
\label{tab:appsolverunpairedruns}
\centering
\fontsize{7.5}{9}\selectfont
\setlength{\tabcolsep}{0.5pt}
\renewcommand{\arraystretch}{1.10}
\begin{tabular*}{1\linewidth}{@{\extracolsep{\fill}}>{\raggedright\arraybackslash}m{.10\linewidth}>{\centering\arraybackslash}m{.055\linewidth}*{15}{>{\centering\arraybackslash}m{.053\linewidth}}@{}}
\toprule
\tworowhead{Solver} & \tworowhead{Run} & \multicolumn{3}{c}{DICM} & \multicolumn{3}{c}{LIME} & \multicolumn{3}{c}{MEF} & \multicolumn{3}{c}{NPE} & \multicolumn{3}{c}{VV}\\
\cmidrule(lr){3-5}\cmidrule(lr){6-8}\cmidrule(lr){9-11}\cmidrule(lr){12-14}\cmidrule(lr){15-17}
 &  & \shortstack{{\fontsize{6}{7}\selectfont ILNIQE}\\$\downarrow$} & \shortstack{{\fontsize{6}{7}\selectfont BRISQUE}\\$\downarrow$} & \shortstack{{\fontsize{6}{7}\selectfont NL}\\$\downarrow$} & \shortstack{{\fontsize{6}{7}\selectfont ILNIQE}\\$\downarrow$} & \shortstack{{\fontsize{6}{7}\selectfont BRISQUE}\\$\downarrow$} & \shortstack{{\fontsize{6}{7}\selectfont NL}\\$\downarrow$} & \shortstack{{\fontsize{6}{7}\selectfont ILNIQE}\\$\downarrow$} & \shortstack{{\fontsize{6}{7}\selectfont BRISQUE}\\$\downarrow$} & \shortstack{{\fontsize{6}{7}\selectfont NL}\\$\downarrow$} & \shortstack{{\fontsize{6}{7}\selectfont ILNIQE}\\$\downarrow$} & \shortstack{{\fontsize{6}{7}\selectfont BRISQUE}\\$\downarrow$} & \shortstack{{\fontsize{6}{7}\selectfont NL}\\$\downarrow$} & \shortstack{{\fontsize{6}{7}\selectfont ILNIQE}\\$\downarrow$} & \shortstack{{\fontsize{6}{7}\selectfont BRISQUE}\\$\downarrow$} & \shortstack{{\fontsize{6}{7}\selectfont NL}\\$\downarrow$}\\
\midrule
\multirow[c]{5}{=}{DarkIR} & 0 & 21.4787 & 13.3828 & 0.5543 & 25.0134 & 11.4127 & 0.8820 & 24.8749 & 16.4345 & 0.8831 & 23.3678 & 17.5354 & 0.5497 & 22.4427 & 13.4285 & 0.4210\\
 & 21 & 21.6225 & 13.7839 & 0.5305 & 24.7481 & 12.6533 & 0.9274 & 24.5742 & 16.3222 & 0.9145 & 23.1538 & 17.2403 & 0.5484 & 22.7369 & 18.1034 & 0.4102\\
 & 42 & 21.5567 & 14.6144 & 0.5485 & 24.7666 & 10.8522 & 0.9994 & 25.5854 & 15.3839 & 0.9953 & 23.4485 & 17.0188 & 0.5607 & 22.7293 & 14.1624 & 0.4172\\
 & Mean & 21.5526 & 13.9270 & 0.5444 & 24.8427 & 11.6394 & 0.9363 & 25.0115 & 16.0469 & 0.9310 & 23.3234 & 17.2648 & 0.5529 & 22.6363 & 15.2314 & 0.4161\\
 & SD & 0.0720 & 0.6282 & 0.0124 & 0.1481 & 0.9217 & 0.0592 & 0.5193 & 0.5769 & 0.0579 & 0.1523 & 0.2592 & 0.0068 & 0.1677 & 2.5141 & 0.0055\\
\midrule
\multirow[c]{5}{=}{CIDNet} & 0 & 21.8491 & 16.1532 & 0.5406 & 26.8875 & 20.8568 & 0.6724 & 26.0952 & 19.4561 & 0.7083 & 23.5802 & 21.5650 & 0.4647 & 21.9680 & 26.5105 & 0.4188\\
 & 21 & 22.0032 & 15.3317 & 0.5238 & 28.1493 & 20.4837 & 0.5987 & 26.5587 & 21.7656 & 0.6461 & 23.9258 & 22.4714 & 0.4800 & 22.4992 & 24.2154 & 0.4164\\
 & 42 & 22.3035 & 16.3290 & 0.5614 & 26.8160 & 18.7456 & 0.6137 & 26.4180 & 18.4915 & 0.6251 & 23.5205 & 21.0296 & 0.4850 & 22.5699 & 21.9765 & 0.4211\\
 & Mean & 22.0519 & 15.9380 & 0.5419 & 27.2843 & 20.0287 & 0.6283 & 26.3573 & 19.9044 & 0.6598 & 23.6755 & 21.6887 & 0.4766 & 22.3457 & 24.2341 & 0.4188\\
 & SD & 0.2311 & 0.5323 & 0.0188 & 0.7500 & 1.1267 & 0.0389 & 0.2376 & 1.6825 & 0.0433 & 0.2188 & 0.7288 & 0.0106 & 0.3290 & 2.2671 & 0.0024\\
\midrule
\multirow[c]{5}{=}{SNR-Aware} & 0 & 22.4223 & 20.1079 & 0.5577 & 25.7466 & 15.5462 & 1.1902 & 26.2421 & 19.4589 & 1.2343 & 24.0677 & 23.6791 & 0.6026 & 22.6744 & 17.4302 & 0.4384\\
 & 21 & 21.8187 & 15.9488 & 0.5445 & 25.7087 & 15.8294 & 1.2454 & 24.9862 & 16.7359 & 1.3915 & 23.4591 & 20.4775 & 0.5914 & 21.9816 & 16.5743 & 0.4419\\
 & 42 & 22.5279 & 24.3992 & 0.5923 & 26.4487 & 21.6855 & 1.2215 & 26.9769 & 23.6249 & 1.2314 & 24.9819 & 27.5812 & 0.6348 & 23.0930 & 24.4423 & 0.4489\\
 & Mean & 22.2563 & 20.1520 & 0.5648 & 25.9680 & 17.6870 & 1.2190 & 26.0684 & 19.9399 & 1.2857 & 24.1696 & 23.9126 & 0.6096 & 22.5830 & 19.4823 & 0.4431\\
 & SD & 0.3826 & 4.2254 & 0.0247 & 0.4167 & 3.4657 & 0.0277 & 1.0067 & 3.4696 & 0.0916 & 0.7665 & 3.5576 & 0.0225 & 0.5613 & 4.3168 & 0.0053\\
\midrule
\multirow[c]{5}{=}{LLFormer} & 0 & 22.0848 & 11.6491 & 0.5195 & 25.5361 & 13.5913 & 0.7549 & 25.3477 & 14.8326 & 0.7310 & 23.7622 & 16.4728 & 0.5112 & 22.2228 & 13.8249 & 0.3873\\
 & 21 & 21.4437 & 15.8106 & 0.4649 & 25.5419 & 17.6970 & 0.9559 & 25.3326 & 18.9344 & 0.9553 & 23.8021 & 19.3184 & 0.4902 & 22.0410 & 20.4658 & 0.3732\\
 & 42 & 22.0306 & 16.6580 & 0.4662 & 26.5426 & 18.4311 & 0.5648 & 26.8645 & 20.4870 & 0.5407 & 24.0636 & 19.2835 & 0.4538 & 22.4208 & 19.6712 & 0.3730\\
 & Mean & 21.8530 & 14.7059 & 0.4835 & 25.8735 & 16.5731 & 0.7585 & 25.8483 & 18.0847 & 0.7423 & 23.8760 & 18.3582 & 0.4851 & 22.2282 & 17.9873 & 0.3778\\
 & SD & 0.3555 & 2.6810 & 0.0312 & 0.5794 & 2.6083 & 0.1956 & 0.8801 & 2.9214 & 0.2075 & 0.1637 & 1.6329 & 0.0290 & 0.1900 & 3.6266 & 0.0082\\
\midrule
\multirow[c]{5}{=}{Restormer} & 0 & 22.2085 & 11.7792 & 0.6264 & 26.8386 & 14.4880 & 0.9682 & 25.1718 & 15.5383 & 0.9674 & 23.7195 & 20.0482 & 0.5707 & 22.8688 & 21.6584 & 0.4608\\
 & 21 & 22.3515 & 11.2998 & 0.6009 & 27.0329 & 13.7909 & 0.8508 & 25.2089 & 14.6115 & 0.8691 & 23.8579 & 20.2828 & 0.5471 & 22.8732 & 20.6133 & 0.4373\\
 & 42 & 22.2627 & 12.2415 & 0.6126 & 26.9448 & 14.5033 & 0.9038 & 24.5359 & 15.3777 & 0.9602 & 23.9914 & 19.6132 & 0.5713 & 22.5667 & 20.7412 & 0.4541\\
 & Mean & 22.2742 & 11.7735 & 0.6133 & 26.9388 & 14.2607 & 0.9076 & 24.9722 & 15.1758 & 0.9322 & 23.8563 & 19.9814 & 0.5630 & 22.7696 & 21.0043 & 0.4507\\
 & SD & 0.0722 & 0.4709 & 0.0128 & 0.0973 & 0.4070 & 0.0588 & 0.3783 & 0.4953 & 0.0548 & 0.1360 & 0.3398 & 0.0138 & 0.1757 & 0.5701 & 0.0121\\
\bottomrule
\end{tabular*}
\end{table*}

In \cref{tab:appsolverruns}, DarkIR has the smallest LOL PSNR standard deviation among the five Solvers, while Restormer has the smallest Unpaired ILNIQE and BRISQUE standard deviations. The dataset-level records in \cref{tab:appsolverpairedruns,tab:appsolverunpairedruns} show how this variation differs across subsets.

\subsection{Additional-Solver evaluation and inference cost}
The five Solvers in \cref{tab:solvers} share the evaluation protocol in \cref{app:evalpreprocess}. Each runs in evaluation mode at native image resolution, with mean-intensity input preprocessing and zero additional input noise. The DarkIR row uses the main-experiment mean from \cref{tab:main}, with individual run scores and sample standard deviations in \cref{tab:appsolverruns}.

\paragraph{Inference FLOPs sources and counting.}
\label{app:flops}
Since the Challenger and LPIPS loss network are used only during training, enhancement requires one forward pass of the chosen Solver. DarkIR-m~\citep{Feijoo_2025_CVPR} is reported to use $O_{\rm src}=7.25\times10^9$ multiply-accumulate (MAC) operations at $H_{\rm src}=W_{\rm src}=256$. We estimate FLOPs at MR.~Illuminate's resolution, $H_{\rm tgt}=400$ and $W_{\rm tgt}=600$, by converting at $\kappa=2$ FLOPs per MAC and scaling by image area. The estimate is
\begin{equation}
 \widehat F_{\rm ours}=\kappa O_{\rm src}
 \frac{H_{\rm tgt}W_{\rm tgt}}{H_{\rm src}W_{\rm src}}
 =2(7.25\times10^9)\frac{400\cdot600}{256\cdot256}.
 \label{eq:flopsestimate}
\end{equation}
\Cref{eq:flopsestimate} estimates the cost of the existing DarkIR architecture at approximately $5.31\times10^{10}$ FLOPs. The table rounds the coefficient to one decimal place.

For CIDNet and SNR-Aware in \cref{tab:solvers}, we retain the $400\times600$ FLOPs reported in Table~1 of MR.~Illuminate, namely $6.0\times10^{10}$ and $1.9\times10^{11}$. These entries share the source and input size of the corresponding values in \cref{tab:appextended}.

For Restormer and LLFormer, we count convolution and matrix-multiplication operations in the official implementations using \texttt{FlopCounterMode} in PyTorch~2.10, with two FLOPs per MAC. The counts include attention matrix multiplications and omit normalization, activation, and other elementwise operations such as softmax. Shape-only forward passes use batch size one and evaluation mode, requiring the network structure alone.

Restormer uses width 48 and blocks $[4,6,6,8]$, followed by four refinement blocks. Following the official architecture, we set the attention heads to $[1,2,4,8]$ and the expansion factor to 2.66. Its $400\times600$ forward pass contains $1{,}032{,}645{,}240{,}000$ convolution FLOPs and $101{,}744{,}640{,}000$ matrix-multiplication FLOPs, totaling $1.13438988\times10^{12}$.

LLFormer uses width 16 and blocks $[2,4,8,16]$, followed by two refinement blocks. Its heads are $[1,2,4,8]$, and its expansion factor is 2.66, matching the official test configuration. The same test procedure reflect-pads a $400\times600$ image to $400\times608$, runs the network once, and crops its output back to the input size. This forward pass contains $163{,}540{,}812{,}800$ convolution FLOPs and $248{,}367{,}513{,}600$ matrix-multiplication FLOPs, totaling $4.119083264\times10^{11}$. \Cref{tab:solvers} distinguishes these operator counts ($\ddagger$), the DarkIR area estimate ($\dagger$), and the quoted literature values through their markers and sources.

\subsection{PPO update budget}
\label{app:ppoupdates}
\Cref{tab:appppoupdates} compares two, three, and four PPO optimizer steps per outer iteration with DarkIR on the 115-image LOL test collection. Each setting reports PSNR, SSIM, and LPIPS for all three seeds, with the two-step setting using the main-experiment runs.

\begin{table}[tbp]
\caption{LOL results for different numbers of PPO optimizer steps per outer iteration, with two steps corresponding to the main experiments. Mean rows show mean $\pm$ SD, with PSNR in dB. Boldface marks the best score for each metric within each seed and in the mean rows.}
\label{tab:appppoupdates}
\centering
\footnotesize
\setlength{\tabcolsep}{1pt}
\renewcommand{\arraystretch}{1.10}
\begin{tabular*}{1\linewidth}{@{\extracolsep{\fill}}>{\raggedright\arraybackslash}m{.115\linewidth}>{\centering\arraybackslash}m{.095\linewidth}*{3}{>{\centering\arraybackslash}m{.25\linewidth}}@{}}
\toprule
PPO steps & Run & PSNR $\uparrow$ & SSIM $\uparrow$ & LPIPS $\downarrow$\\
\midrule
\multirow[c]{4}{=}{2 (main)} & 0 & \textbf{22.8103} & \textbf{0.8353} & \textbf{0.1576}\\
 & 21 & \textbf{22.8907} & \textbf{0.8308} & \textbf{0.1574}\\
 & 42 & \textbf{22.8165} & \textbf{0.8275} & \textbf{0.1657}\\
 & Mean & \mbox{\textbf{22.8392}\hspace{0.5pt}{\fontsize{6}{7}\selectfont$\pm 0.0448$}} & \mbox{\textbf{0.8312}\hspace{0.5pt}{\fontsize{6}{7}\selectfont$\pm 0.0039$}} & \mbox{\textbf{0.1602}\hspace{0.5pt}{\fontsize{6}{7}\selectfont$\pm 0.0048$}}\\
\midrule
\multirow[c]{4}{=}{3} & 0 & 22.3198 & 0.8185 & 0.1700\\
 & 21 & 22.2098 & 0.8053 & 0.1924\\
 & 42 & 22.5679 & 0.7977 & 0.1982\\
 & Mean & \mbox{22.3658\hspace{0.5pt}{\fontsize{6}{7}\selectfont$\pm 0.1834$}} & \mbox{0.8072\hspace{0.5pt}{\fontsize{6}{7}\selectfont$\pm 0.0105$}} & \mbox{0.1869\hspace{0.5pt}{\fontsize{6}{7}\selectfont$\pm 0.0149$}}\\
\midrule
\multirow[c]{4}{=}{4} & 0 & 22.5315 & 0.8158 & 0.1710\\
 & 21 & 22.1591 & 0.7489 & 0.2405\\
 & 42 & 22.6118 & 0.7922 & 0.2026\\
 & Mean & \mbox{22.4341\hspace{0.5pt}{\fontsize{6}{7}\selectfont$\pm 0.2415$}} & \mbox{0.7856\hspace{0.5pt}{\fontsize{6}{7}\selectfont$\pm 0.0339$}} & \mbox{0.2047\hspace{0.5pt}{\fontsize{6}{7}\selectfont$\pm 0.0348$}}\\
\bottomrule
\end{tabular*}
\end{table}

Two PPO steps achieve the highest PSNR and SSIM and the lowest LPIPS within every seed, yielding mean scores of 22.8392\,dB, 0.8312, and 0.1602, respectively. This configuration also has the smallest sample standard deviations on all three metrics. Thus, among the tested settings, two PPO steps combine the strongest restoration scores with the lowest policy update count.

\section{Extended comparison with published results}
\label{app:extendedcomparison}
\Cref{tab:appextended} places all three DarkIR runs alongside the 16 external rows of \cref{tab:main}. The table also includes GDP and additional supervised results from MR.~Illuminate.

The lower block contains results from additional publications under their original evaluation settings. Its LOL column covers LOL-v1 only. The corresponding Dark-Zero scores are in \cref{tab:appdatasetrecords}. Superscripts identify LSRW and Unpaired scores reconstructed from published subset means.

\begin{table*}[!t]
\caption{Extended comparison with all three Dark-Zero seeds and additional source reports whose LOL columns cover LOL-v1 only. Both \methodtype{Z} and \methodtype{ZN} denote zero-shot settings, with \textbf{(P)} indicating \textbf{pretrained diffusion models} and \textbf{(I)} indicating \textbf{iterative optimization of model parameters for every input image}. Superscripts mark GT-Mean alignment in quoted results ($\mathrm{GT}$), reconstructed metrics ($\mathrm w$, $\ast$), and estimated FLOPs ($\dagger$), while a dash denotes an unavailable or insufficiently verified entry.}
\label{tab:appextended}
\centering
\scriptsize
\setlength{\tabcolsep}{1.65pt}
\resizebox{\linewidth}{!}{%
\begin{tabular}{@{}cllcccccccccc@{}}
\toprule
\tworowhead{Type} & \tworowhead{Method} & \tworowhead{Train data} & \tworowhead{FLOPs $\downarrow$} & \multicolumn{3}{c}{LOL} & \multicolumn{3}{c}{LSRW} & \multicolumn{3}{c}{Unpaired}\\
\cmidrule(lr){5-7}\cmidrule(lr){8-10}\cmidrule(lr){11-13}
& & & & PSNR $\uparrow$ & SSIM $\uparrow$ & LPIPS $\downarrow$ & PSNR $\uparrow$ & SSIM $\uparrow$ & LPIPS $\downarrow$ & ILNIQE $\downarrow$ & BRISQUE $\downarrow$ & NL $\downarrow$\\
\midrule
\methodtype{S} & Retinexformer~\citep{cai2023retinexformer} & MIT5K & $1.2\!\times\!10^{11}$ & 13.0263 & 0.4256 & 0.3649 & 11.4123 & 0.2689 & 0.5060 & 36.3183 & 38.2946 & 0.3994\\
\methodtype{S} & CIDNet~\citep{11093284} & MIT5K & $6.0\!\times\!10^{10}$ & 12.0433 & 0.3766 & 0.3933 & 10.8772 & 0.2395 & 0.4565 & 34.7661 & 21.8344 & 0.6823\\
\methodtype{S} & Diff-Plugin~\citep{liu2024diff} & LOL-v1 & $2.9\!\times\!10^{13}$ & 18.8273 & 0.7041 & 0.1826 & 17.9620 & 0.5224 & 0.3032 & 22.6731 & 18.1443 & 0.8530\\
\methodtype{S} & GSAD~\citep{hou2023global} & LOL-v1 & $1.1\!\times\!10^{13}$ & 20.6046 & 0.8470 & 0.1118 & 17.5835 & 0.5509 & 0.3269 & 26.3613 & 29.0197 & 0.7501\\
\methodtype{S} & SNR-Aware~\citep{xu2022snr} & LOL-v1 & $1.9\!\times\!10^{11}$ & 21.8877 & 0.8480 & 0.1561 & 17.8400 & 0.5639 & 0.4774 & 28.4187 & 35.2478 & 0.4037\\
\methodtype{S} & GT-Mean Loss${}^{\mathrm{GT}}$~\citep{liao2025gt} & LOL-v1 & $1.2\!\times\!10^{11}$ & 27.7090 & 0.8746 & 0.1227 & 17.5336 & 0.5493 & 0.3913 & 26.6812 & 29.9124 & 0.6085\\
\methodtype{S} & CIDNet${}^{\mathrm{GT}}$~\citep{11093284} & LOL-v1 & $6.0\!\times\!10^{10}$ & 28.9198 & 0.8890 & 0.0671 & 17.4020 & 0.5354 & 0.2855 & 24.9865 & 24.4371 & 0.6823\\
\midrule
\methodtype{U} & CLIP-LIT~\citep{liang2023iterative} & Own data & $1.3\!\times\!10^{11}$ & 14.8179 & 0.5243 & 0.3706 & 13.4835 & 0.4051 & 0.3533 & 28.1403 & 26.0421 & 1.8638\\
\methodtype{U} & Zero-DCE~\citep{guo2020zero} & Own data & $3.8\!\times\!10^{10}$ & 17.6417 & 0.5717 & 0.3154 & 15.8680 & 0.4506 & 0.3152 & 26.8335 & 23.1635 & 1.5336\\
\methodtype{U} & Zero-DCE++~\citep{9369102} & Own data & $4.5\!\times\!10^{7}$ & 17.0175 & 0.4439 & 0.3145 & 16.2453 & 0.4568 & 0.3273 & 25.9503 & 17.1177 & 1.2573\\
\methodtype{U} & NeRCo~\citep{yang2023implicit} & LOL-v1 & $1.7\!\times\!10^{12}$ & 21.0728 & 0.7248 & 0.2591 & 17.3814 & 0.5303 & 0.5182 & 28.1309 & 30.0167 & 0.8358\\
\methodtype{U} & ZeroIG~\citep{shi2024zero} & LOL-v1 & $7.7\!\times\!10^{10}$ & 18.6589 & 0.7496 & 0.2415 & 16.4431 & 0.5087 & 0.3764 & 27.5778 & 26.2227 & 1.4861\\
\methodtype{U} & GEFU~\citep{11446150} & LOL-v1+ & $4.2\!\times\!10^{12}$ & 20.8308 & 0.7898 & 0.2078 & 18.7498 & 0.5421 & 0.2806 & 24.1183 & 17.8064 & 0.6349\\
\methodtype{U} & EnlightenGAN~\citep{jiang2021enlightengan} & LOL-v1+ & $2.4\!\times\!10^{11}$ & 18.4888 & 0.6722 & 0.3105 & 17.0811 & 0.4705 & 0.3273 & 24.7138 & 18.9759 & 0.9667\\
\methodtype{U} & PairLIE~\citep{fu2023learning} & LOL-v1+ & $1.6\!\times\!10^{11}$ & 19.6999 & 0.7737 & 0.2353 & 17.6104 & 0.5190 & 0.3309 & 26.7961 & 31.0421 & 1.6407\\
\methodtype{U} & ZeroIG~\citep{shi2024zero} & LSRW & $7.7\!\times\!10^{10}$ & 17.5677 & 0.4778 & 0.3799 & 16.7516 & 0.5010 & 0.4000 & 34.2056 & 34.8248 & 2.4482\\
\methodtype{U} & QuadPrior~\citep{wang2024zero} & COCO & $4.2\!\times\!10^{13}$ & 20.3016 & 0.8096 & 0.2032 & 16.9469 & 0.5601 & 0.3824 & 24.4373 & 18.1736 & 0.4215\\
\midrule
\methodtype{Z} & RRDNet~\citep{zhu2020zero} & n/a (I) & $6.1\!\times\!10^{13}$ & 13.5719 & 0.4791 & 0.3238 & 13.4272 & 0.3918 & 0.3358 & 26.6567 & 17.9583 & 1.1570\\
\methodtype{Z} & CoLIE~\citep{chobola2024fast} & n/a (I) & $5.2\!\times\!10^{12}$ & 14.8999 & 0.4985 & 0.3268 & 14.0013 & 0.4053 & 0.3424 & 26.8185 & 18.9680 & 0.9642\\
\methodtype{Z} & ExCNet~\citep{zhang2019zero} & n/a (I) & $1.7\!\times\!10^{7}$ & 16.2972 & 0.4589 & 0.3745 & 15.7021 & 0.4098 & 0.3375 & 27.3933 & 19.2416 & 1.8234\\
\methodtype{Z} & TAO~\citep{gou2023test} & n/a (P) & $4.7\!\times\!10^{15}$ & 19.1807 & 0.6065 & 0.3897 & 15.6891 & 0.4306 & 0.7023 & 42.0846 & 42.1383 & 0.3840\\
\methodtype{Z} & FourierDiff~\citep{lv2024fourier} & n/a (P) & $8.5\!\times\!10^{14}$ & 16.9525 & 0.6039 & 0.2934 & 15.6251 & 0.4610 & 0.3207 & 25.9272 & 26.5694 & 1.2206\\
\methodtype{Z} & MR.~Illuminate~\citep{cho2026mr} & n/a (P) & $5.8\!\times\!10^{13}$ & 21.7393 & 0.8152 & 0.1771 & 17.6634 & 0.5185 & 0.2829 & 25.4181 & 16.1843 & 0.3794\\
\methodtype{Z} & GDP~\citep{fei2023generative} & n/a (P) & - & 14.6630 & 0.5037 & 0.3559 & 13.0678 & 0.3918 & 0.4476 & 28.6436 & 27.0142 & 0.5280\\
\midrule
\methodtype{ZN} & Dark-Zero / Seed 42 & COCO & $5.3\!\times\!10^{10}\,{}^{\dagger}$ & 22.8165 & 0.8275 & 0.1657 & 18.2057 & 0.5541 & 0.3063 & 23.7998 & 15.4448 & 0.7104\\
\methodtype{ZN} & Dark-Zero / Seed 21 & COCO & $5.3\!\times\!10^{10}\,{}^{\dagger}$ & 22.8907 & 0.8308 & 0.1574 & 17.8623 & 0.5479 & 0.3062 & 23.3726 & 16.1637 & 0.6717\\
\methodtype{ZN} & Dark-Zero / Seed 0 & COCO & $5.3\!\times\!10^{10}\,{}^{\dagger}$ & 22.8103 & 0.8353 & 0.1576 & 17.8017 & 0.5524 & 0.3095 & 23.4979 & 15.6842 & 0.6651\\
\toprule
\multicolumn{13}{l}{\textit{Additional source reports (LOL-v1 only, $\mathrm{w}$ reweighted LSRW, $\ast$ reweighted Unpaired BRISQUE)}}\\
\methodtype{S} & Multinex~\citep{brateanu2026multinex} & LOL-v1 & - & 23.19 & 0.843 & 0.129 & - & - & - & - & - & -\\
\methodtype{S} & Multinex-Nano~\citep{brateanu2026multinex} & LOL-v1 & - & 19.42 & 0.742 & 0.276 & - & - & - & - & - & -\\
\midrule
\methodtype{U} & LightenDiff.~\citep{jiang2024lightendiffusion} & Own data & - & 20.453 & 0.803 & 0.192 & 18.555 & 0.539 & 0.311 & - & - & -\\
\methodtype{U} & CLODE~\citep{ICLR2025_5c89b403} & LOL-v1 & - & 19.61 & 0.718 & 0.263 & 17.28 & 0.533 & 0.331 & - & 15.2528${}^{\ast}$ & -\\
\methodtype{U} & Li \etal~\citeyearpar{ICLR2025_6264f71a} & LOL-v1 & - & 19.80 & 0.750 & 0.253 & - & - & - & - & - & -\\
\midrule
\methodtype{Z} & AGLLDiff~\citep{Lin_Ye_Chen_Fu_Wang_Chai_Xing_Li_Zhu_Ding_2025} & n/a (P) & - & 21.81 & 0.84 & 0.15 & - & - & - & - & 19.4331${}^{\ast}$ & -\\
\midrule
\methodtype{ZN} & HVR-SSLE~\citep{11396663} & COCO & - & 17.44 & 0.7286 & - & 17.1200${}^{\mathrm w}$ & 0.5143${}^{\mathrm w}$ & - & - & - & -\\
\bottomrule
\end{tabular}}
\end{table*}

\subsection{Variation across main-experiment runs}
\Cref{tab:appmainvariance} summarizes the nine quality metrics in the main experiment across three seeds. These summaries cover LOL, LSRW, and Unpaired under the protocol in \cref{app:runsummary}.

The three-run means reproduce the Dark-Zero row in \cref{tab:main}. The LOL summaries also match the full-model entries in \cref{tab:appablationruns}.

\subsection{Sources and scope of the added rows}
Rows inherited from MR.~Illuminate retain their source labels, and its measurement of GDP exceeded the computational limit for reporting FLOPs. For the added rows, we use the categories defined in \cref{sec:introduction}. AGLLDiff falls under \methodtype{Z} since it uses a pretrained diffusion prior without task-specific LLIE training. 

AGLLDiff values come from Tables~1 and 2 of the published AAAI paper. Table~1 reports paired metrics on LOL-v1, SICE~\citep{cai2018learning}, and LOL-v2-synthetic. We retain the LOL-v1 triplet in \cref{tab:appextended}, while keeping LOL-v2-synthetic separate from the real split and leaving LSRW unavailable. The source's unpaired table supplies BRISQUE, natural image quality evaluator (NIQE)~\citep{6353522}, and perceptual index (PI)~\citep{10.1007/978-3-030-11021-5_21}. Since NIQE and PI differ from ILNIQE and NL, respectively, only BRISQUE can be reconstructed for our Unpaired columns.

For CLODE, we report the default PSNR and SSIM from the Normal columns of Table~1. Supplementary Tables~12 and~13 supply LPIPS and unpaired BRISQUE, respectively. Table~12 reports a single LSRW LPIPS score without specifying its exposure setting.

LightenDiffusion, abbreviated LightenDiff. in \cref{tab:appextended}, uses the paired scores in Table~1 of arXiv:2407.08939v1. Its first training stage uses low-light observations from SICE, followed by diffusion training with approximately 180,000 unpaired low-light and normal-light image pairs, represented by ``Own data'' in our table. The official repository provides pretrained-model links and an evaluation script. The published no-reference scores on DICM, NPE, and VV use NIQE and PI.

Li \etal provide an interpretable method for joint denoising and enhancement with zero-reference training on low-light observations. We use the paired metrics in Table~1 of their ICLR 2025 paper for LOL-v1 and LOL-v2-real. The official repository lists separate checkpoints for these datasets and testing instructions.

Multinex and Multinex-Nano add CVPR 2026 lightweight supervised baselines. We use Table~1 of arXiv:2604.10359v3, and the repository provides checkpoints and evaluation instructions. Since their LOL-v1 and LOL-v2-real scores use dataset-specific training, we preserve the two splits separately in \cref{tab:appextended,tab:apprecentv2}. The source's complexity counts use a different input resolution and are therefore omitted from our FLOPs column. For BRISQUE, \cref{tab:apprecentbr} retains the four available dataset means from source Table~2, with VV unavailable. These results use the standard models, while the source's dagger-marked retrained variants are treated as separate models.

HVR-SSLE trains on synthetic pairs from COCO. Its fixed network runs recurrently during inference. The authors provide \texttt{HVR.safetensors}, its configuration, and an inference script. Its Table~2 reports PSNR and SSIM of 17.44\,dB and 0.7286 on LOL-v1, and 19.99\,dB and 0.7710 on LOL-v2-real. On LSRW, the Huawei subset has PSNR of 17.52\,dB and SSIM of 0.5465, while the Nikon subset has 16.52\,dB and 0.4659, respectively. Weighting the 30 Huawei and 20 Nikon images using \cref{eq:appweightedpublished} gives PSNR of approximately 17.1200\,dB and SSIM of 0.5143, marked $\mathrm w$ in \cref{tab:appextended}.
HVR-SSLE's four dataset-level BRISQUE scores in \cref{tab:apprecentbr} come from Table~3 of the source paper.
\Cref{tab:apprecentv2} collects the published LOL-v2-real scores under each source's original evaluation procedure.
For Unpaired, \cref{tab:apprecentbr} reports the four available subset scores for the standard Multinex models and HVR-SSLE. VV scores would be needed to reconstruct a mean over all five subsets.


\subsection{Image-count-weighted reconstruction}
For a metric reported as an image-wise mean $\bar m_d$ on each constituent dataset $d$ in a collection $\mathcal D$, let $n_d$ be its positive image count. Writing its image-level values as $m_{d,j}$, the identity $\bar m_d=n_d^{-1}\sum_{j=1}^{n_d}m_{d,j}$ gives the pooled mean
\begin{equation}
 \bar m_{\mathcal D}
 =\frac{\sum_{d\in\mathcal D}\sum_{j=1}^{n_d}m_{d,j}}{\sum_{d\in\mathcal D}n_d}
 =\frac{\sum_{d\in\mathcal D}n_d\bar m_d}{\sum_{d\in\mathcal D}n_d}.
 \label{eq:appweightedpublished}
\end{equation}
We reweight the published BRISQUE means in \cref{tab:appbrinputs} using the image counts in \cref{tab:appup} and mark the resulting scores with $\ast$ in \cref{tab:appextended}. The scores inherit rounding from the published means and represent a pooled mean over our test set only if the sources use the same test images and metric definition. For PSNR, \cref{eq:appweightedpublished} averages image-level scores in dB.

\medskip
\noindent\begin{minipage}{\linewidth}
\captionsetup{type=table,skip=4pt}
\caption{Main-experiment results for Dark-Zero with DarkIR as the Solver, reported as mean $\pm$ SD across three seeds. PSNR is in dB.}
\label{tab:appmainvariance}
\centering
\footnotesize
\setlength{\tabcolsep}{1.6pt}
\renewcommand{\arraystretch}{1}
\begin{tabular*}{1\linewidth}{@{\extracolsep{\fill}}>{\raggedright\arraybackslash}m{.21\linewidth}>{\raggedright\arraybackslash}m{.18\linewidth}>{\centering\arraybackslash}m{.52\linewidth}@{}}
\toprule
Collection & Metric & Mean $\pm$ SD\\
\midrule
\multirow[c]{3}{=}{LOL} & PSNR $\uparrow$ & \mbox{22.8392\hspace{0.5pt}{\fontsize{6}{7}\selectfont$\pm 0.0448$}}\\
 & SSIM $\uparrow$ & \mbox{0.8312\hspace{0.5pt}{\fontsize{6}{7}\selectfont$\pm 0.0039$}}\\
 & LPIPS $\downarrow$ & \mbox{0.1602\hspace{0.5pt}{\fontsize{6}{7}\selectfont$\pm 0.0048$}}\\
\midrule
\multirow[c]{3}{=}{LSRW} & PSNR $\uparrow$ & \mbox{17.9566\hspace{0.5pt}{\fontsize{6}{7}\selectfont$\pm 0.2179$}}\\
 & SSIM $\uparrow$ & \mbox{0.5515\hspace{0.5pt}{\fontsize{6}{7}\selectfont$\pm 0.0032$}}\\
 & LPIPS $\downarrow$ & \mbox{0.3073\hspace{0.5pt}{\fontsize{6}{7}\selectfont$\pm 0.0018$}}\\
\midrule
\multirow[c]{3}{=}{Unpaired} & ILNIQE $\downarrow$ & \mbox{23.5568\hspace{0.5pt}{\fontsize{6}{7}\selectfont$\pm 0.2196$}}\\
 & BRISQUE $\downarrow$ & \mbox{15.7642\hspace{0.5pt}{\fontsize{6}{7}\selectfont$\pm 0.3661$}}\\
 & NL $\downarrow$ & \mbox{0.6824\hspace{0.5pt}{\fontsize{6}{7}\selectfont$\pm 0.0245$}}\\
\bottomrule
\end{tabular*}
\end{minipage}
\par\medskip
\noindent\begin{minipage}{\linewidth}
\captionsetup{type=table,skip=4pt}
\caption{Published LOL-v2-real results. Li \etal and Multinex use models trained on the corresponding dataset, while HVR-SSLE uses its COCO-trained model.}
\label{tab:apprecentv2}
\centering\footnotesize
\setlength{\tabcolsep}{2.4pt}
\begin{tabular*}{1\linewidth}{@{\extracolsep{\fill}}lccc@{}}
\toprule
Method & PSNR $\uparrow$ & SSIM $\uparrow$ & LPIPS $\downarrow$\\
\midrule
Li \etal & 20.22 & 0.793 & 0.266\\
Multinex & 23.04 & 0.860 & 0.178\\
Multinex-Nano & 19.66 & 0.784 & 0.266\\
HVR-SSLE & 19.99 & 0.7710 & -\\
\bottomrule
\end{tabular*}
\end{minipage}

\par\medskip
\noindent\begin{minipage}{\linewidth}
\captionsetup{type=table,skip=4pt}
\caption{BRISQUE reports for the standard Multinex models from Table~2 of arXiv:2604.10359v3 and HVR-SSLE from its source Table~3. Lower is better.}
\label{tab:apprecentbr}
\centering\footnotesize
\setlength{\tabcolsep}{2.4pt}
\begin{tabular*}{1\linewidth}{@{\extracolsep{\fill}}lccc@{}}
\toprule
Dataset & Multinex & Multinex-Nano & HVR-SSLE\\
\midrule
DICM & 16.39 & 19.97 & 25.30\\
MEF & 13.82 & 15.06 & 11.83\\
LIME & 14.60 & 15.17 & 13.11\\
NPE & 12.52 & 15.42 & 18.67\\
VV & - & - & -\\
\bottomrule
\end{tabular*}
\end{minipage}
\par\medskip
\noindent\begin{minipage}{\linewidth}
\captionsetup{type=table,skip=4pt}
\caption{Published BRISQUE means from AGLLDiff Table~2 and CLODE supplementary Table~13, with target collection counts. Lower is better.}
\label{tab:appbrinputs}
\centering\footnotesize
\setlength{\tabcolsep}{2.4pt}
\begin{tabular*}{1\linewidth}{@{\extracolsep{\fill}}lccc@{}}
\toprule
Dataset & Images & AGLLDiff & CLODE\\
\midrule
DICM & 44 & 19.13 & 20.30\\
LIME & 10 & 19.13 & 16.26\\
MEF & 79 & 23.21 & 12.78\\
NPE & 75 & 15.13 & 13.72\\
VV & 24 & 21.13 & 18.51\\
\midrule
Weighted & 232 & 19.4331 & 15.2528\\
\bottomrule
\end{tabular*}
\end{minipage}

\begingroup

\section{Additional qualitative comparisons}
\label{app:qualitative}
Restorations from the DarkIR Solver and the baselines complement the scores in \cref{tab:main,tab:appextended}. Yellow boxes mark the regions enlarged below each image.

Green denotes supervised methods \methodtype{S}, and red denotes unsupervised methods \methodtype{U}. Blue and gold denote the zero-shot settings \methodtype{Z} and \methodtype{ZN}, respectively, with normal-light-only training in \methodtype{ZN}. QuadPrior retains MR.~Illuminate's \methodtype{U} label from \cref{tab:appextended}. RetinexNet~\citep{wei2018deep} is marked \methodtype{S} since it trains on paired images.

The preprocessing in \cref{app:evalpreprocess} applies to these examples. CIDNet uses MIT-Adobe FiveK weights throughout. For LSRW and Unpaired, Retinexformer uses FiveK weights, ZeroIG uses LSRW-Huawei weights, and GEFU uses its SCUF checkpoint. On Unpaired, GDP uses 1,000 sampling steps on \(256\times256\) inputs.

\subsection{Paired benchmarks}
\label{app:qualitativepaired}
The LOL crops in \crefrange{fig:supp-real-lolv2-00714}{fig:supp-real-lolv2-00770} show differences in local contrast, detail, and color relative to the normal-light references. The LSRW comparisons in \crefrange{fig:supp-real-lsrw-2041}{fig:supp-real-lsrw-2044} focus on local color casts on floors and walls.

\begin{figure}[!htbp]
  \centering
  \includegraphics[width=\linewidth]{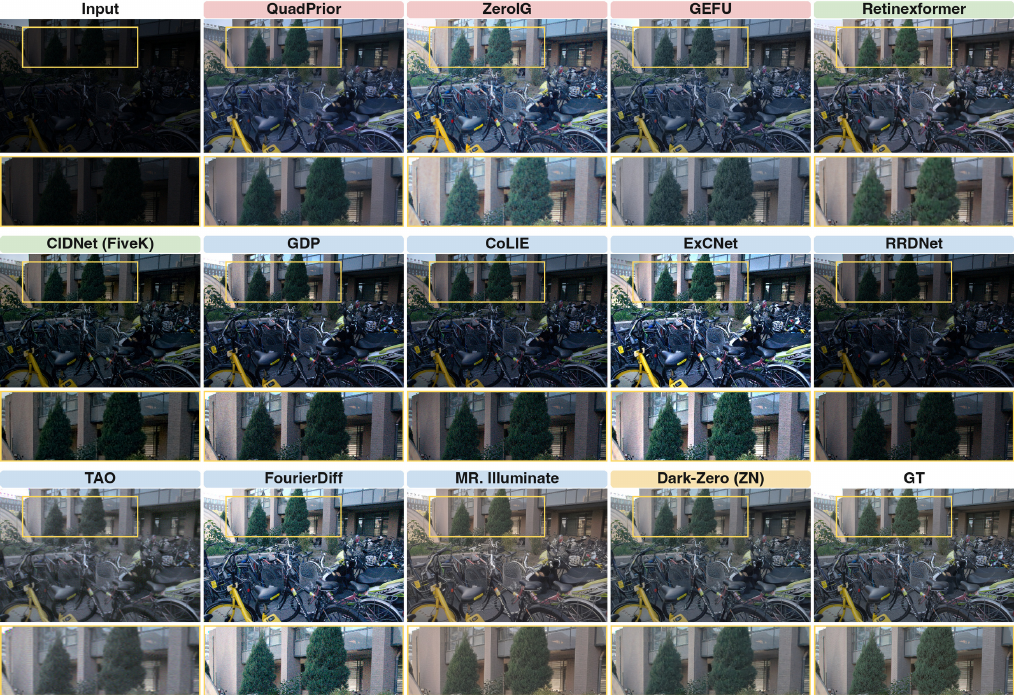}
  \caption{LOL. Dark-Zero retains the muted brick tones and green foliage of the reference in the facade crop. MR.~Illuminate gives the facade a warmer tint, and QuadPrior and FourierDiff shift it toward blue.}
  \label{fig:supp-real-lolv2-00714}
\end{figure}

\begin{figure*}[!tp]
  \centering
  \includegraphics[width=\linewidth]{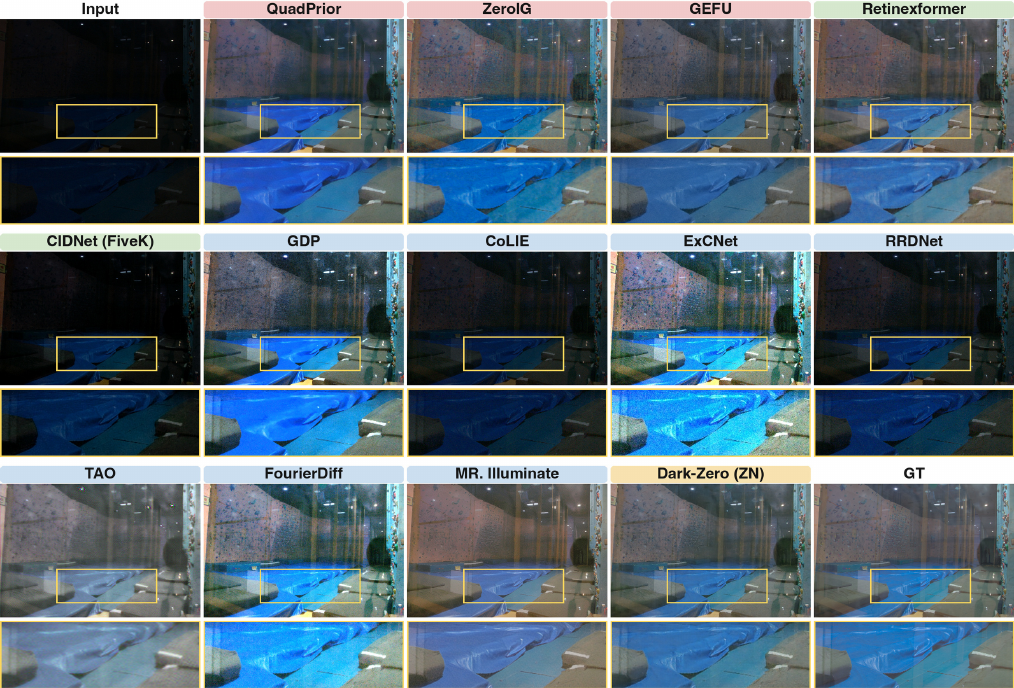}
  \caption{LOL. The folds in the blue padding behind the glass are visible in Dark-Zero, and the padding remains distinct from the floor. MR.~Illuminate renders it in a muted violet-blue, and QuadPrior and FourierDiff produce more saturated blues.}
  \label{fig:supp-real-lolv2-00758}
\end{figure*}

\begin{figure*}[!tp]
  \centering
  \includegraphics[width=\linewidth]{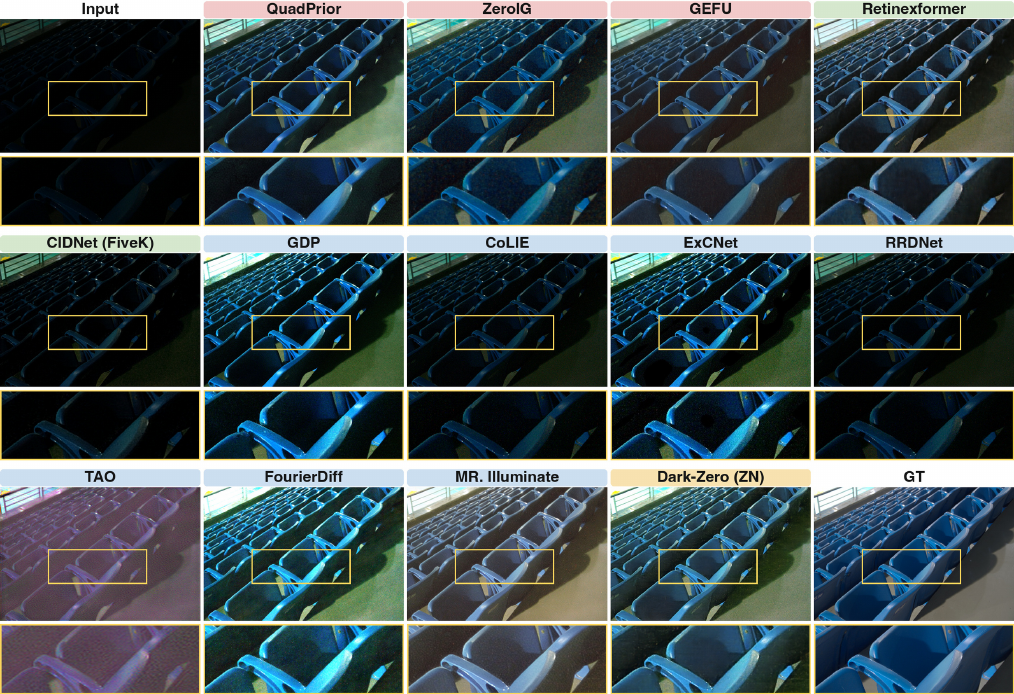}
  \caption{LOL. Dark-Zero retains the blue seats, lighter rims, and shaded interiors. MR.~Illuminate produces a more muted gray-blue, and FourierDiff shifts the rims toward cyan.}
  \label{fig:supp-real-lolv2-00767}
\end{figure*}

\begin{figure*}[!tp]
  \centering
  \includegraphics[width=\linewidth]{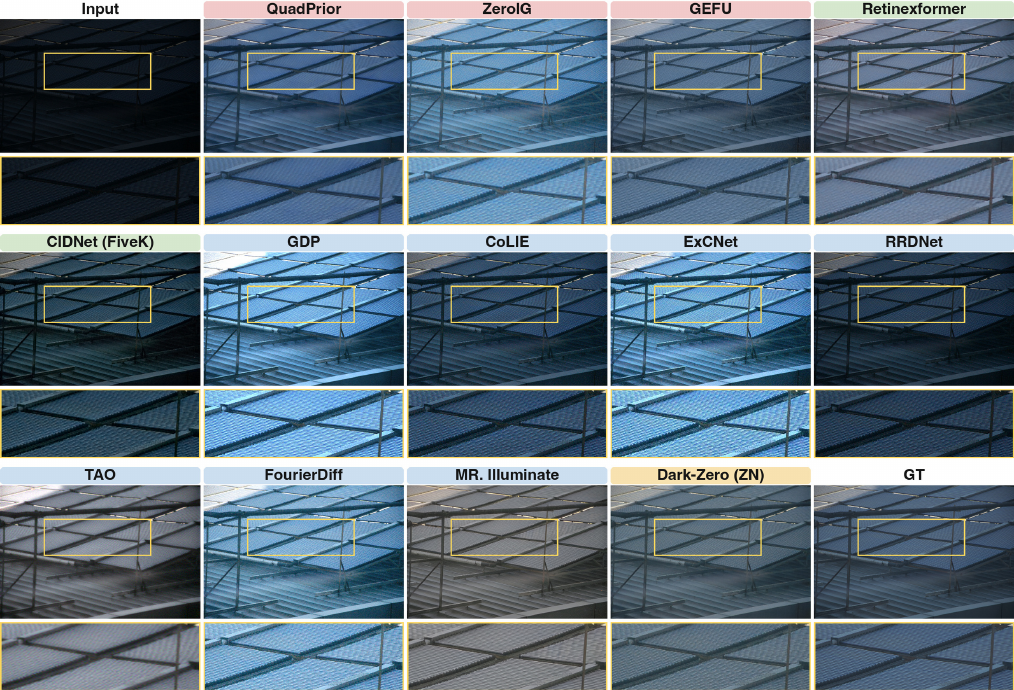}
  \caption{LOL. Dark-Zero retains the cool tone of the roof and the corrugation between the diagonal supports. MR.~Illuminate renders the roof closer to neutral gray, and GDP and ExCNet brighten the panels relative to the reference.}
  \label{fig:supp-real-lolv2-00771}
\end{figure*}

\begin{figure*}[!tp]
  \centering
  \includegraphics[width=\linewidth]{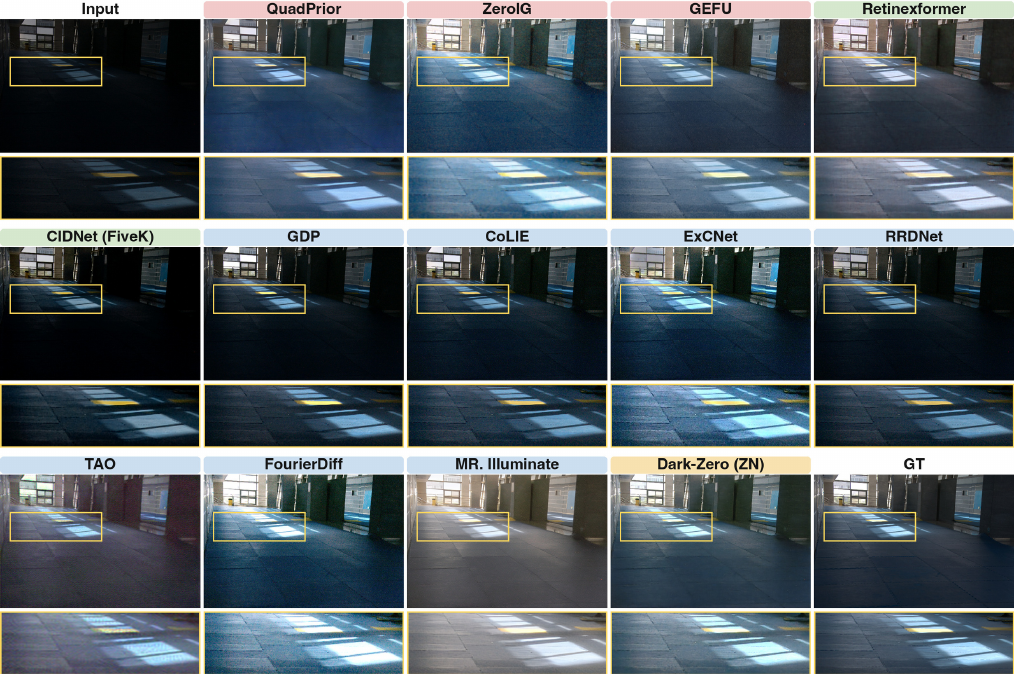}
  \caption{LOL. Dark-Zero keeps the tile boundaries visible between the bright window reflections and the shaded floor. MR.~Illuminate brightens the surrounding tiles and shifts their tone from cool blue-gray toward gray-brown.}
  \label{fig:supp-real-lolv2-00770}
\end{figure*}

\begin{figure*}[!tp]
  \centering
  \includegraphics[width=\linewidth]{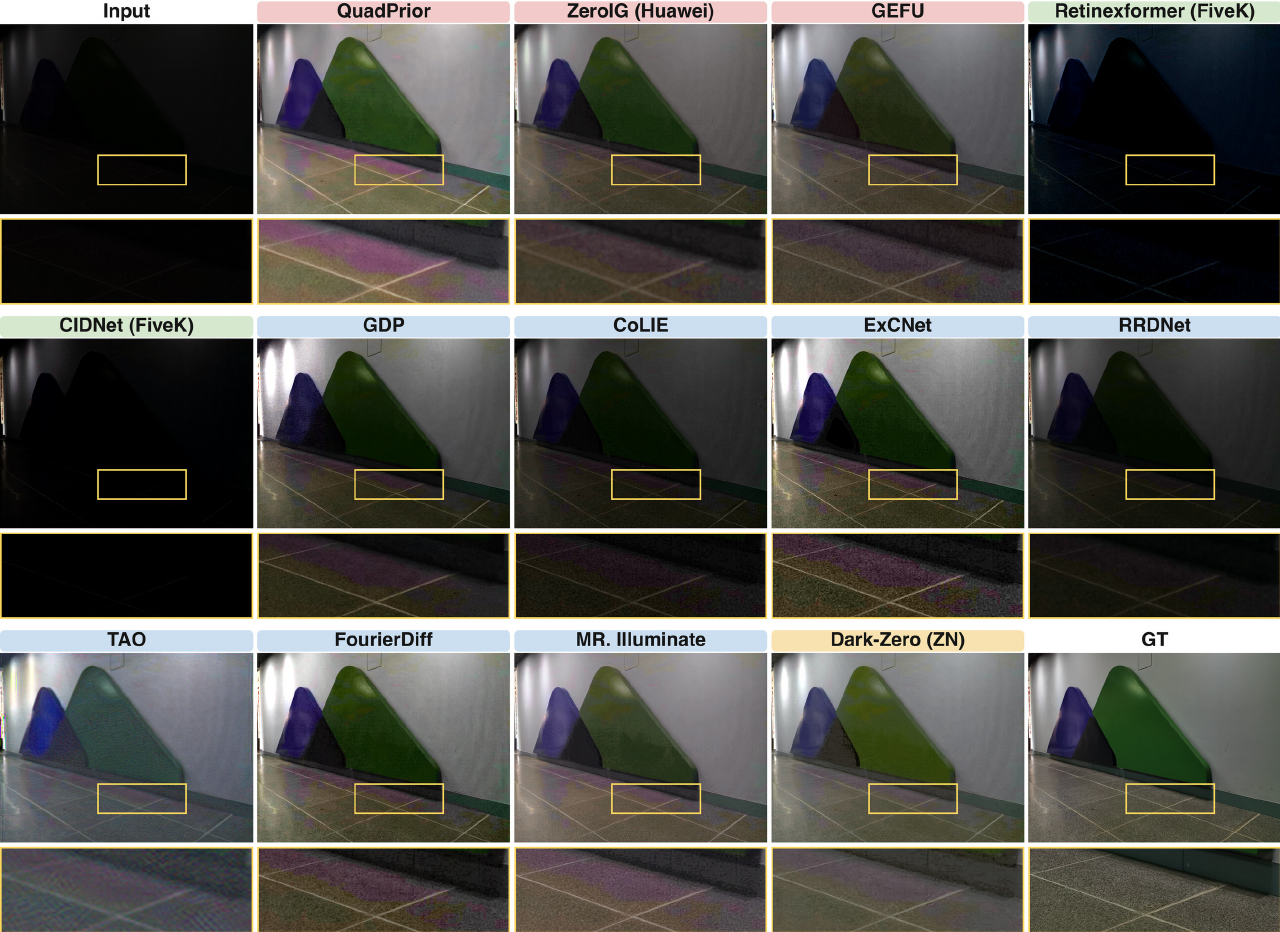}
  \caption{LSRW. Near the baseboard, Dark-Zero brings the floor color closer to the reference. MR.~Illuminate and QuadPrior show a stronger magenta cast in that region.}
  \label{fig:supp-real-lsrw-2041}
\end{figure*}

\begin{figure*}[!tp]
  \centering
  \includegraphics[width=\linewidth]{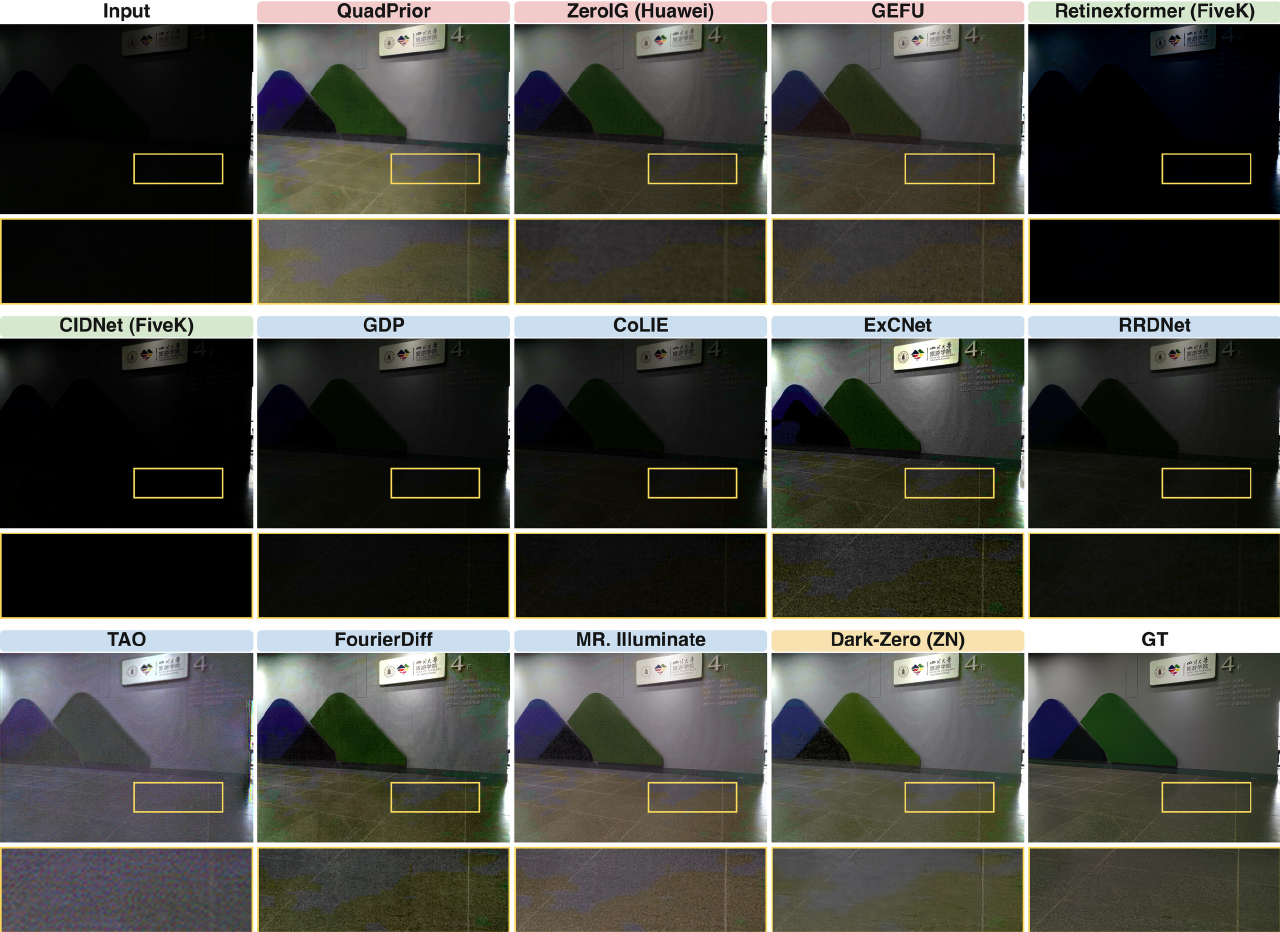}
  \caption{LSRW. In the enlarged floor region, Dark-Zero reduces the purple cast visible in MR.~Illuminate and renders the tiles closer to the reference tone.}
  \label{fig:supp-real-lsrw-2042}
\end{figure*}

\begin{figure*}[!tp]
  \centering
  \includegraphics[width=\linewidth]{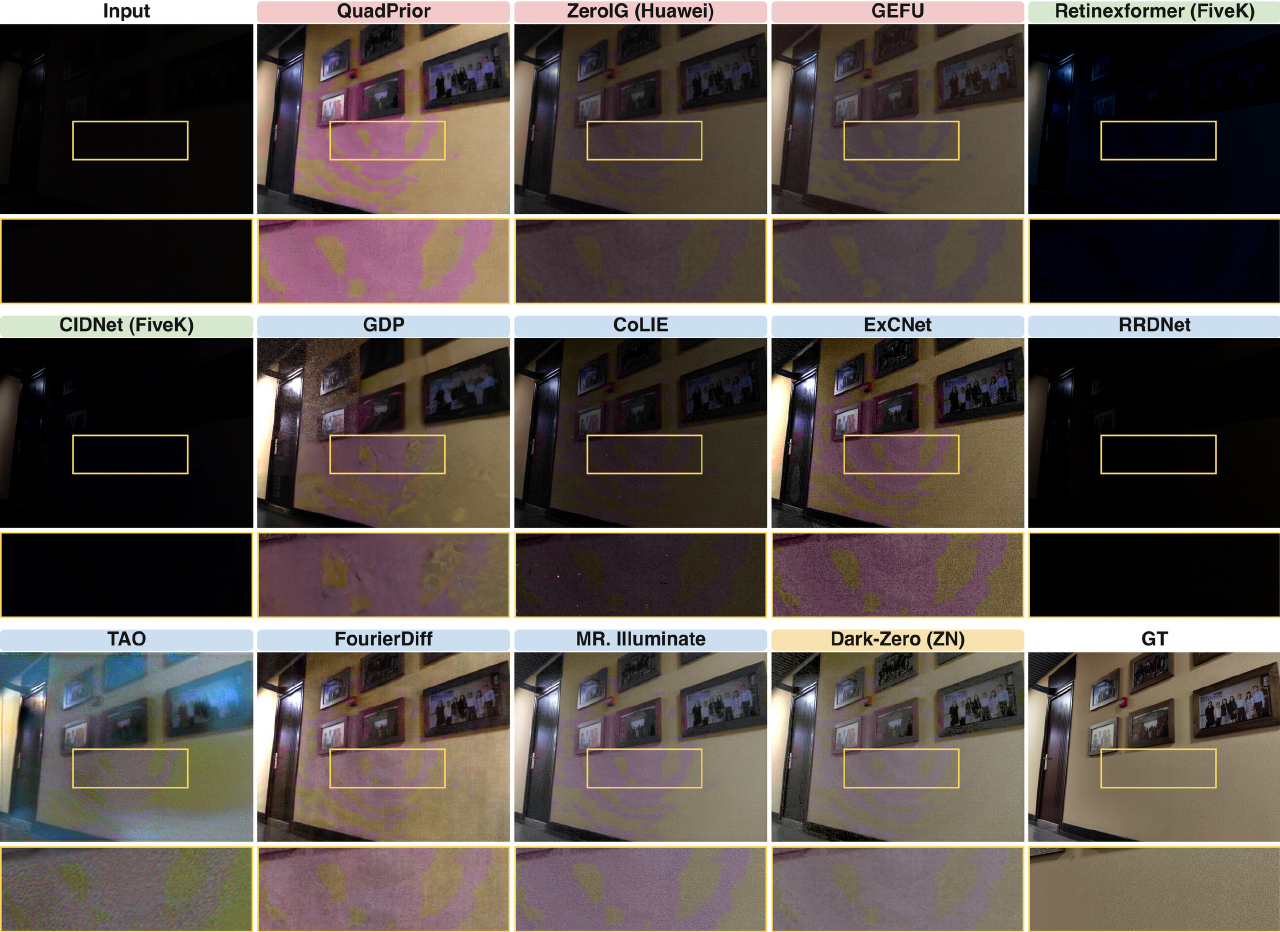}
  \caption{LSRW. Dark-Zero shows a weaker purple cast on the beige wall than MR.~Illuminate and QuadPrior. The wall color is closer to the reference in the enlarged region.}
  \label{fig:supp-real-lsrw-2044}
\end{figure*}

\begin{figure*}[!tp]
  \subsection{Unpaired benchmarks}
  \label{app:qualitativeunpaired}
  The unpaired examples in \crefrange{fig:supp-real-unpaired-dicm-41}{fig:supp-real-unpaired-vv-p1040134} cover an outdoor scene in DICM, a candlelit scene in MEF, and scenes with strong backlighting in NPE and VV. In the MEF sequence in \crefrange{fig:supp-real-unpaired-mef-6069}{fig:supp-real-unpaired-mef-6079}, the four exposures show how tabletop details and color balance change as the input darkens.
  \par\medskip
  \centering
  \includegraphics[width=\linewidth]{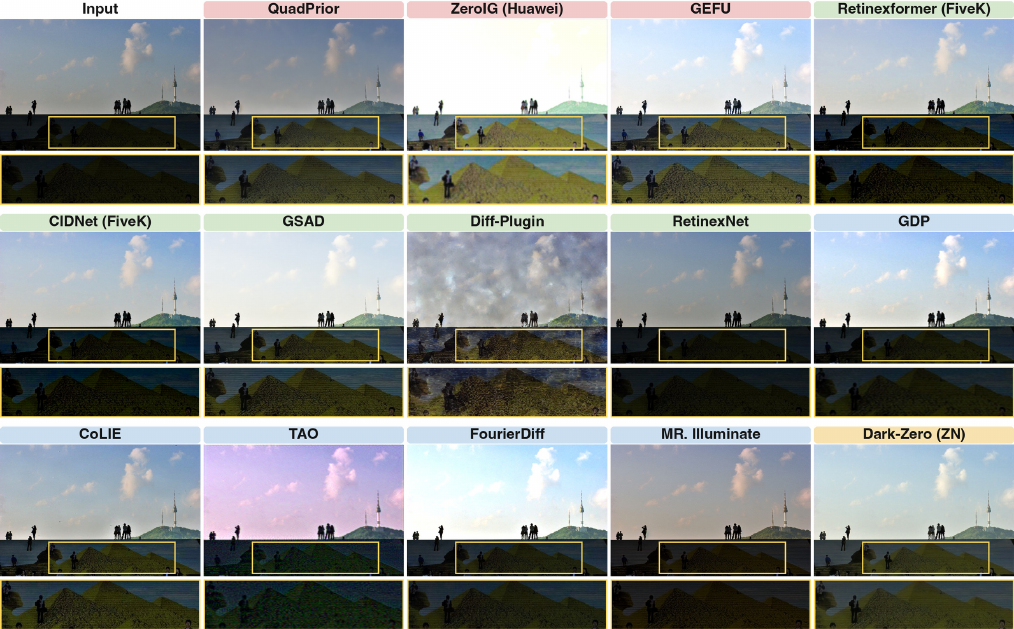}
  \caption{Unpaired. The foreground figure and grassy slope are clearer in Dark-Zero than in MR.~Illuminate, and cloud structure remains visible. The sky in ZeroIG is brighter and has less cloud contrast.}
  \label{fig:supp-real-unpaired-dicm-41}
\end{figure*}

\begin{figure*}[!tp]
  \centering
  \includegraphics[width=\linewidth]{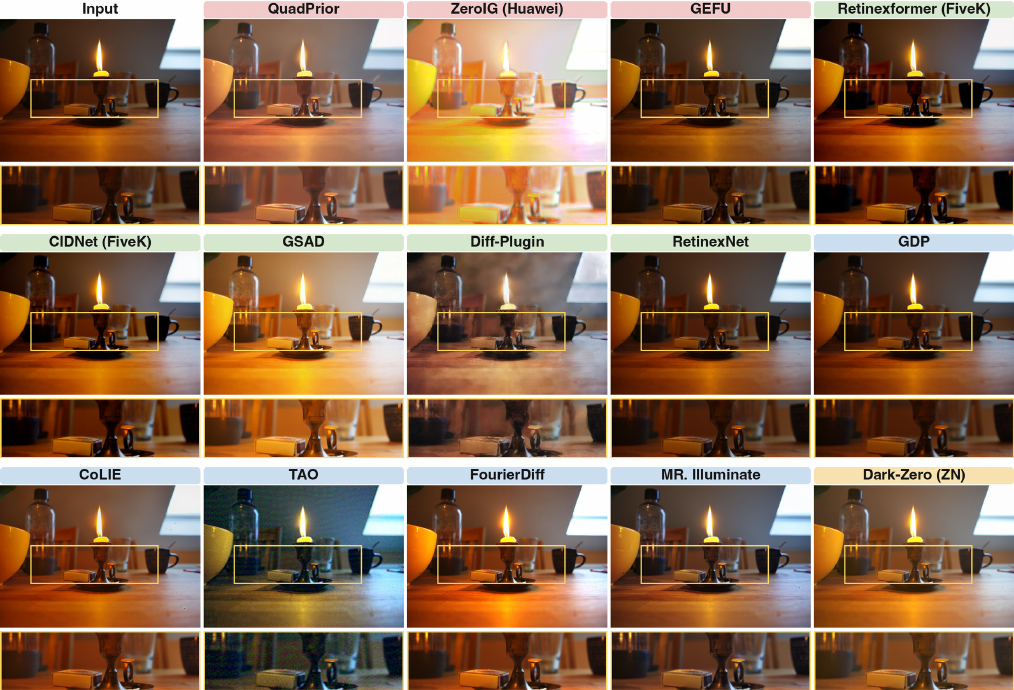}
  \caption{Unpaired. Dark-Zero retains the contrast between warm candlelight on the table and cooler light near the window. MR.~Illuminate gives the right side of the table a stronger blue tint, and ZeroIG brightens the window-side region enough to obscure some detail.}
  \label{fig:supp-real-unpaired-mef-6069}
\end{figure*}

\begin{figure*}[!tp]
  \centering
  \includegraphics[width=\linewidth]{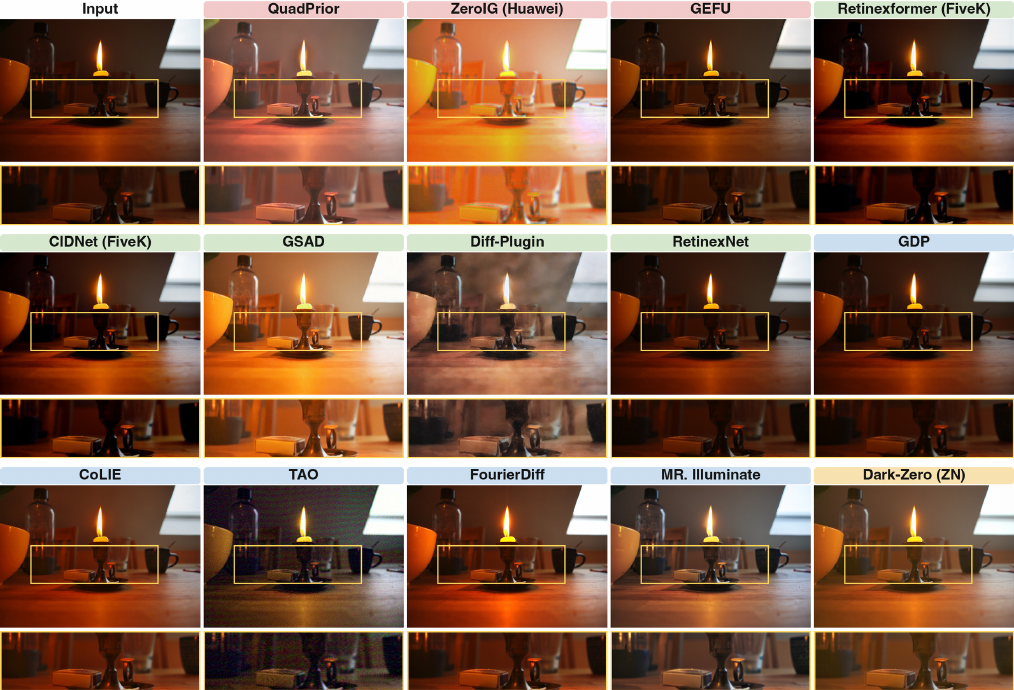}
  \caption{Unpaired. Dark-Zero reveals the edge of the matchbox and the candleholder stem while retaining the warm reflection below the flame. In Retinexformer and CIDNet, these objects remain in deeper shadow. MR.~Illuminate renders the tabletop with a cooler tone.}
  \label{fig:supp-real-unpaired-mef-6076}
\end{figure*}

\begin{figure*}[!tp]
  \centering
  \includegraphics[width=\linewidth]{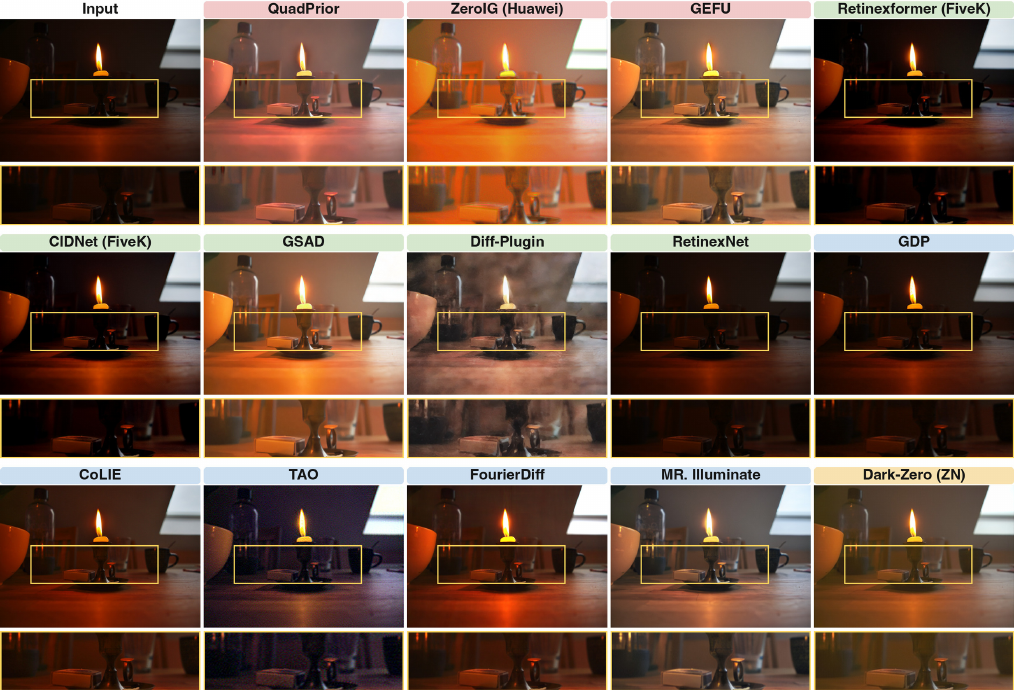}
  \caption{Unpaired. Dark-Zero and MR.~Illuminate reveal the matchbox and candleholder in shadow. Dark-Zero keeps a stronger amber reflection beneath the flame, and FourierDiff renders the central tabletop in a more saturated orange.}
  \label{fig:supp-real-unpaired-mef-6077}
\end{figure*}

\begin{figure*}[!tp]
  \centering
  \includegraphics[width=\linewidth]{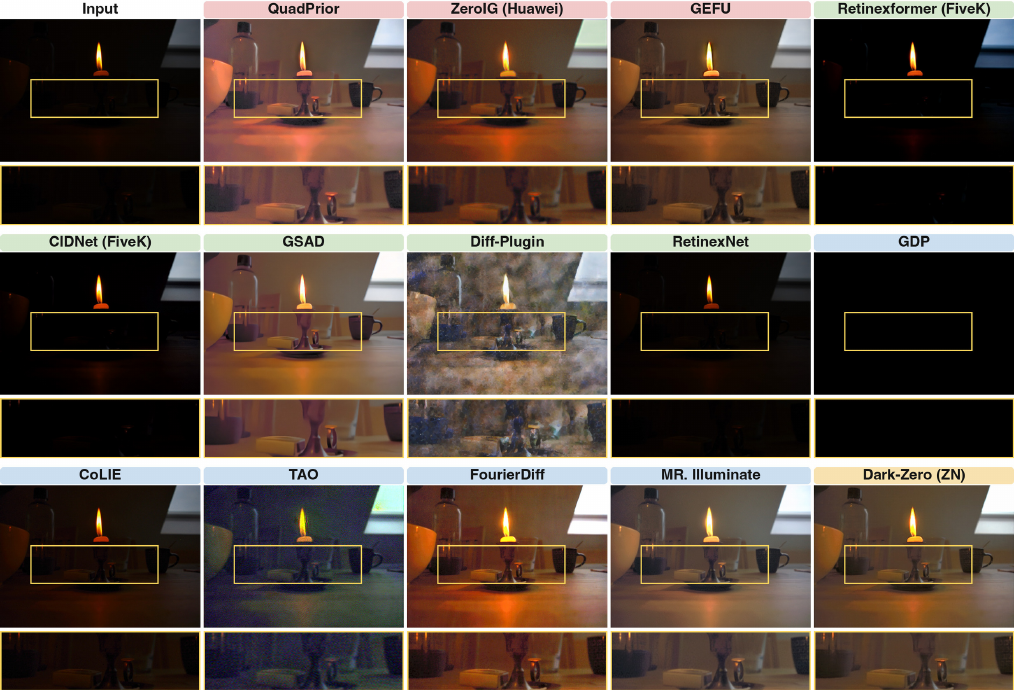}
  \caption{Unpaired, using the darkest input in the candlelit sequence. Dark-Zero and MR.~Illuminate reveal the outlines of the objects on the table, and Dark-Zero retains a warmer reflection below the flame. Most objects remain in deep shadow in Retinexformer and CIDNet.}
  \label{fig:supp-real-unpaired-mef-6079}
\end{figure*}

\begin{figure*}[!tp]
  \centering
  \includegraphics[width=\linewidth]{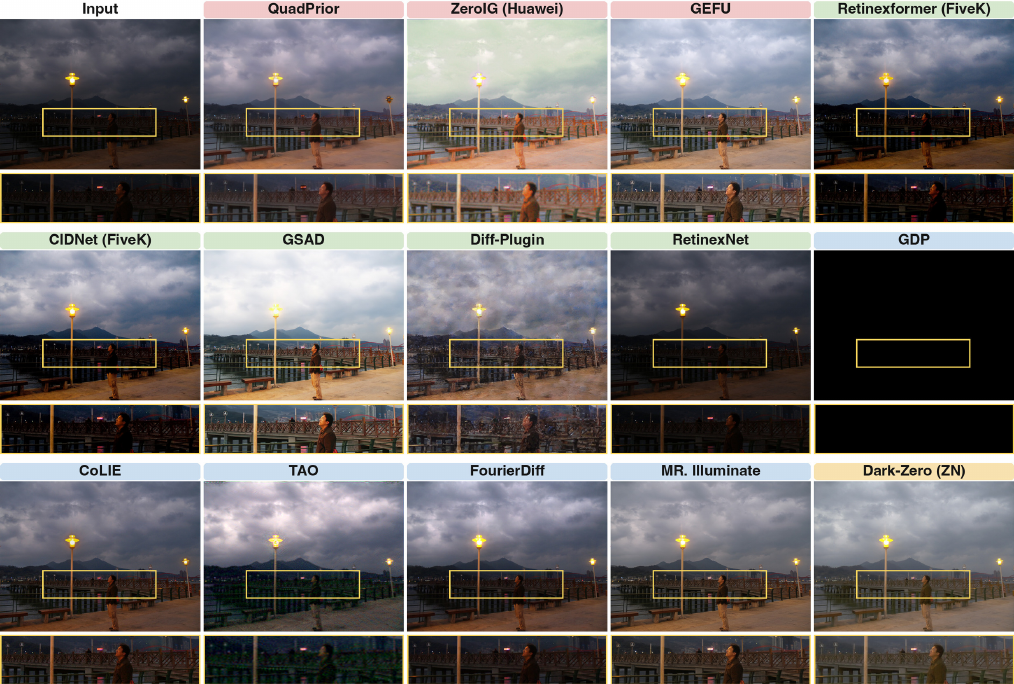}
  \caption{Unpaired. The person and foreground railing are clearer in Dark-Zero than in MR.~Illuminate. Dark-Zero also retains visible cloud structure and warm lamplight, and ZeroIG gives the sky a pale green tint. 
  }
  \label{fig:supp-real-unpaired-npe-cloudy-6}
\end{figure*}

\begin{figure*}[!tp]
  \centering
  \includegraphics[width=\linewidth]{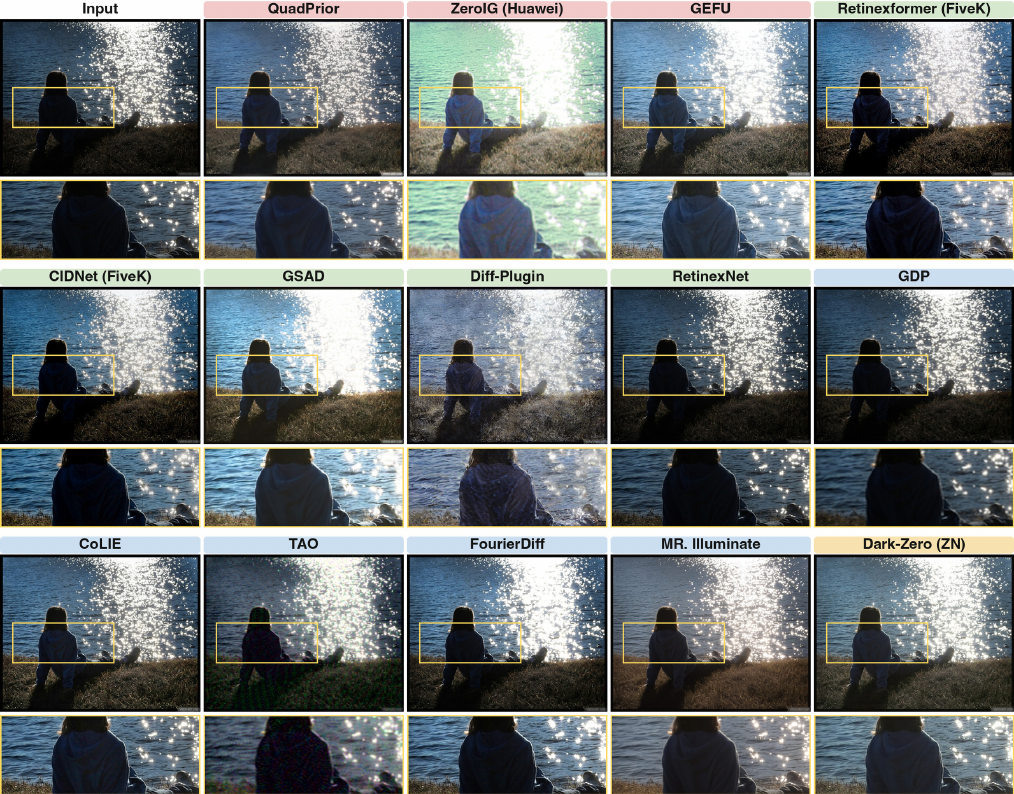}
  \caption{Unpaired. Dark-Zero reveals folds in the jacket and retains green tones in the foreground and cool blue tones in the water. MR.~Illuminate renders the grass with a warmer brown tint. ZeroIG shifts the water toward green and reduces detail in some of its brightest reflections.}
  \label{fig:supp-real-unpaired-npe-night-50}
\end{figure*}

\begin{figure*}[!tp]
  \centering
  \includegraphics[width=\linewidth]{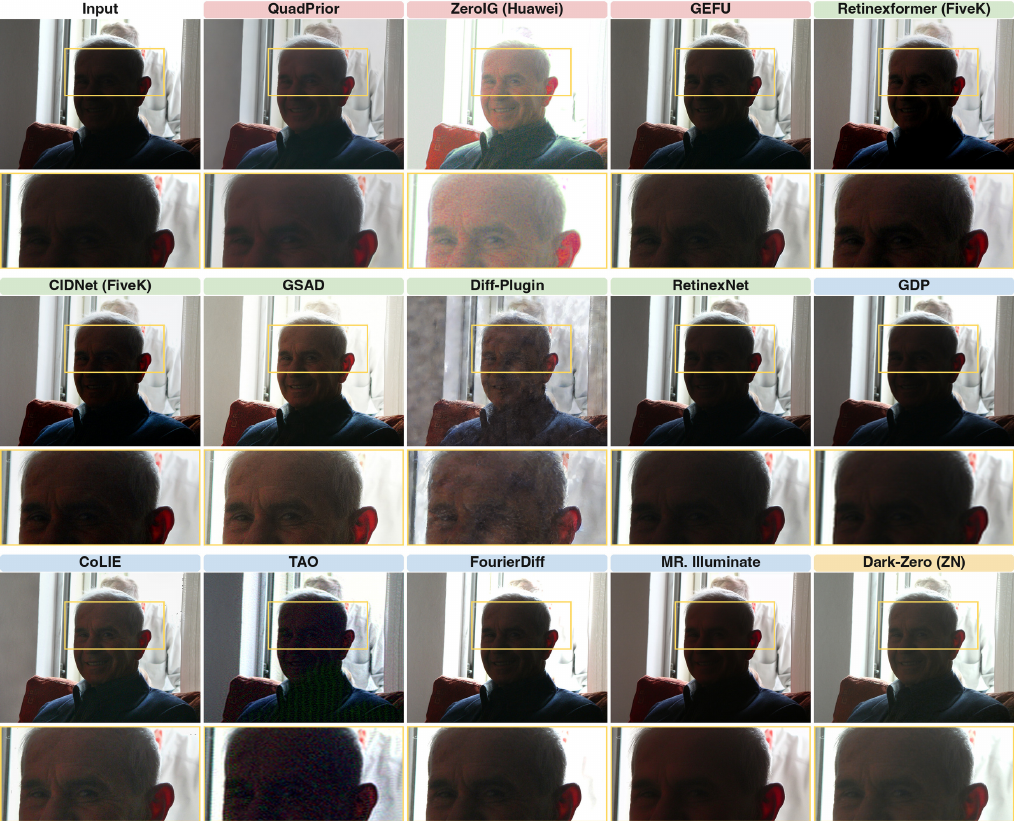}
  \caption{Unpaired. Under strong backlighting, Dark-Zero reveals more detail around the eyes and forehead than MR.~Illuminate. In MR.~Illuminate, these features remain in deeper shadow and have a warmer tone.}
  \label{fig:supp-real-unpaired-vv-p1020150}
\end{figure*}

\begin{figure*}[!tp]
  \centering
  \includegraphics[width=\linewidth]{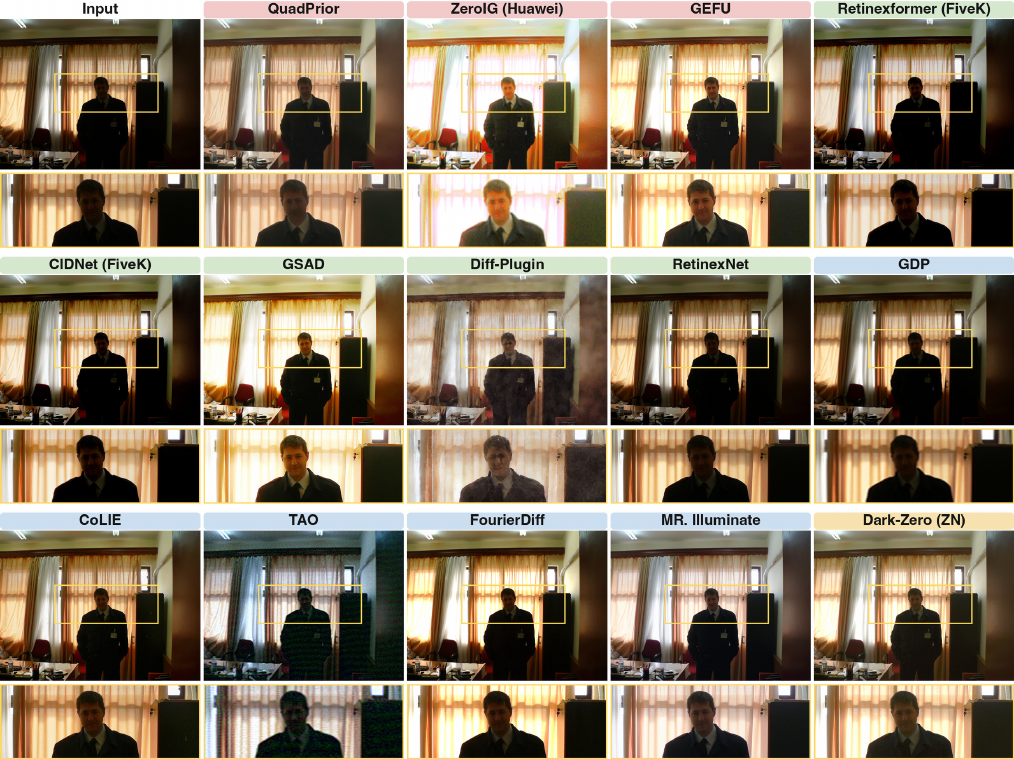}
  \caption{Unpaired. Dark-Zero reveals the face and jacket collar and keeps the warm illumination and folds of the curtains visible. MR.~Illuminate gives the curtains a cooler, slightly pink tint, and ZeroIG renders them more yellow-orange.}
  \label{fig:supp-real-unpaired-vv-p1040134}
\end{figure*}

\endgroup
\FloatBarrier

\end{document}